\documentclass{article}

\usepackage{microtype}
\usepackage{graphicx}
\usepackage{subfigure}
\usepackage{booktabs} % for professional tables

\usepackage{hyperref}

\usepackage[accepted]{icml2023} 

\usepackage{amsmath}
\usepackage{amssymb}
\usepackage{mathtools}
\usepackage{amsthm}

\usepackage[capitalize,noabbrev]{cleveref}

\theoremstyle{plain}
\newtheorem{theorem}{Theorem}[section]

\newtheorem{lemma}[theorem]{Lemma}

\theoremstyle{definition}

\newtheorem{assumption}[theorem]{Assumption}
\theoremstyle{remark}

\usepackage[textsize=tiny]{todonotes}

\usepackage{enumitem} 

\usepackage{multirow} 
\usepackage{soul}
\usepackage{microtype}
\usepackage{graphicx}
\usepackage{subfigure}
\usepackage{booktabs} % for professional tables
\usepackage{makecell}
\usepackage{fancybox}
\usepackage{soul} 
\usepackage{graphicx} 
\usepackage{amsmath}
\usepackage{booktabs}
\usepackage{algorithm}
\usepackage{algorithmic}
\usepackage{todonotes}

\usepackage{amsfonts} 
\usepackage{verbatim} 
\usepackage{amssymb}
\usepackage{verbatim}
\usepackage{enumitem}
\usepackage{sidecap,color} 
\usepackage{verbatim,colortbl}

\def \teta {\tilde{\eta}}

\def \x {\mathbf{x}}
\def \u {\mathbf{u}}

\def \w {\mathbf{w}}
\def \wb {\bar{\mathbf{w}}} 
\def \z {\mathbf{z}}

\def \E {\mathbb{E}}
\def \I {\mathbb{I}}
\def \S{\mathcal{S}}
\def \H{\mathcal{H}}
\def \B{\mathcal{B}}

\def \bG {\bar{G}} 
\def \R {\mathbb{R}}

\def \tL {\tilde{L}}

\icmltitlerunning{FeDXL: Provable Federated Learning for Deep X-Risk Optimization} 

\begin{document} 

\twocolumn[
\icmltitle{FeDXL: Provable Federated Learning for Deep X-Risk Optimization}

% It is OKAY to include author information, even for blind
% submissions: the style file will automatically remove it for you
% unless you've provided the [accepted] option to the icml2023
% package.

% List of affiliations: The first argument should be a (short)
% identifier you will use later to specify author affiliations
% Academic affiliations should list Department, University, City, Region, Country
% Industry affiliations should list Company, City, Region, Country

% You can specify symbols, otherwise they are numbered in order.
% Ideally, you should not use this facility. Affiliations will be numbered
% in order of appearance and this is the preferred way.
\icmlsetsymbol{equal}{*}

\begin{icmlauthorlist}
\icmlauthor{Zhishuai Guo}{yyy}
\icmlauthor{Rong Jin}{comp}
\icmlauthor{Jiebo Luo}{sch} 
\icmlauthor{Tianbao Yang}{yyy} 
%\icmlauthor{}{sch}
%\icmlauthor{}{sch}
\end{icmlauthorlist}

\icmlaffiliation{yyy}{Department of Computer Science and Engineering, Texas A\&M University}
\icmlaffiliation{comp}{Alibaba}
\icmlaffiliation{sch}{Department of Computer Science, University of Rochester} 

\icmlcorrespondingauthor{Zhishuai Guo}{zhishguo@tamu.edu}
\icmlcorrespondingauthor{Tianbao Yang}{tianbao-yang@tamu.edu} 
% You may provide any keywords that you
% find helpful for describing your paper; these are used to populate
% the "keywords" metadata in the PDF but will not be shown in the document
%\icmlkeywords{Machine Learning, ICML}
\vskip 0.3in
]

% this must go after the closing bracket ] following \twocolumn[ ...

% This command actually creates the footnote in the first column
% listing the affiliations and the copyright notice.
% The command takes one argument, which is text to display at the start of the footnote.
% The \icmlEqualContribution command is standard text for equal contribution.
% Remove it (just {}) if you do not need this facility.

\printAffiliationsAndNotice{}  % leave blank if no need to mention equal contribution
% \printAffiliationsAndNotice{\icmlEqualContribution} % otherwise use the standard text.

\begin{abstract}
In this paper, we tackle a novel federated learning (FL) problem for optimizing a family of X-risks, to which no existing FL algorithms are applicable. 
In particular, the objective has the form of $\mathbb{E}_{\mathbf{z}\sim \mathcal S_1} f(\mathbb{E}_{\mathbf{z}'\sim\mathcal S_2} \ell(\mathbf{w}; \mathbf{z}, \mathbf{z}'))$, where two sets of data $\mathcal S_1, \mathcal S_2$ are distributed over multiple machines, $\ell(\cdot; \cdot,\cdot)$ is a pairwise loss that only depends on the prediction outputs of the input data pairs $(\mathbf{z}, \mathbf{z}')$.
This problem has important applications in machine learning, e.g., AUROC maximization with a pairwise loss, and partial AUROC maximization with a compositional loss. 
The challenges for designing an FL algorithm for X-risks lie in the non-decomposability of the objective over multiple machines and the interdependency between different machines. 
To this end, we propose an {\bf active-passive decomposition} framework that decouples the gradient's components with two types, namely active parts and passive parts, where the {\it active} parts depend on local data that are computed with the local model and the {\it passive} parts depend on other machines that are communicated/computed based on historical models and samples. 
Under this framework, we design two FL algorithms (FeDXL) for handling linear and nonlinear $f$, respectively, based on {\bf federated averaging and merging} and develop a novel theoretical analysis to combat the latency of the passive parts and the interdependency between the local model parameters and the involved data for computing local gradient estimators. 
We establish both iteration and communication complexities and show that using the historical samples and models for computing the passive parts do not degrade the complexities. 
We conduct empirical studies of FeDXL for deep AUROC and  partial AUROC maximization, and demonstrate their performance compared with several baselines. 
\end{abstract}

{\setlength\abovedisplayskip{4pt}
\setlength\belowdisplayskip{4pt}
\section{Introduction} 
This work is motivated by solving the following optimization problem arising in many ML applications in a {\bf federated learning (FL)} setting: 
\begin{align}\label{eqn:DXO}
    \min_{\w\in\R^d}\frac{1}{|\S_1|}\sum_{\z\in\S_1}f\bigg(\underbrace{\frac{1}{|\S_2|}\sum_{\z'\in\S_2}\ell(\w,\z, \z')}\limits_{g(\w,\z, \S_2)}\bigg),
    \end{align}
where $\S_1$ and $\S_2$ denote two sets of data points that are distributed over many machines, $\w$ denotes the model of a prediction function $h(\w,\cdot)\in\R^{d_o}$, $f(\cdot)$ is a deterministic function that could be linear or  non-linear (possibly non-convex),  and $\ell(\w,\z, \z') = \ell(h(\w, \z), h(\w,\z'))$ denotes a pairwise loss that only depends on the prediction outputs of the input data $\z, \z'$. %We refer to the above objective as compositional pairwise risk (DXO) to facilitate the comparison with the traditional empirical risk.  
The above problem belongs to a broader family of machine learning problems called deep X-risk optimization (DXO)~\citep{DBLP:journals/corr/abs-2206-00439}.  We provide details of some X-risk minimization applications in Appendix \ref{app:applicaitons}.

When $f$ is a linear function, the above problem is the classic pairwise loss minimization problem, which has applications in AUROC (AUC) maximization~\citep{DBLP:conf/icml/GaoJZZ13,DBLP:conf/icml/ZhaoHJY11,DBLP:conf/ijcai/GaoZ15,DBLP:conf/pkdd/CaldersJ07,DBLP:conf/icml/CharoenphakdeeL19,yang2021simple,10.1145/3554729}, bipartite ranking~\citep{cohen1997learning,clemenccon2008ranking,DBLP:conf/icml/KotlowskiDH11,dembczynski2012consistent}, and distance metric learning~\citep{radenovic2016cnn,wu2017sampling}.  When $f$ is a non-linear function, the above problem is a special case of finite-sum coupled compositional optimization problem~\citep{wang2022finite}, which has found applications in various performance measure optimization such as partial AUC maximization~\citep{DBLP:journals/corr/abs-2203-00176},  average precision  maximization~\citep{DBLP:conf/nips/QiLXJY21,DBLP:conf/aistats/0006YZY22}, NDCG maximization~\citep{DBLP:conf/icml/QiuHZZY22}, p-norm push optimization~\citep{DBLP:journals/jmlr/Rudin09,wang2022finite} and contrastive loss optimization \citep{goldberger2004neighbourhood,DBLP:conf/icml/YuanWQDZZY22}.

This is in sharp contrast with most existing studies on FL algorithms~\citep{yang2013trading,konevcny2016federated,mcmahan2017communication,kairouz2021advances,smith2018cocoa,stich2018local,yu2019linear,yu2019parallel,khaled2020tighter,woodworth2020minibatch,woodworth2020local,karimireddy2020scaffold,haddadpour2019local}, which focus on the following empirical risk minimization (ERM) problem with the data set $\S$ distributed over different machines:
\begin{align}\label{eqn:erm}
    \min_{\w\in\R^d}\frac{1}{|\S|}\sum_{\z\in\S}\ell(\w,\z). 
\end{align}
The major differences between DXO and ERM are (i) the ERM's objective is decomposable over training data, while the DXO is not; and (ii) the data-dependent losses in ERM are decoupled between different data points; in contrast the data-dependent loss in DXO couples different training data points. These differences pose a big challenge for  DXO in the FL setting where the training data are distributed on different machines and are prohibited to be moved to a central server. In particular, the gradient of X-risk cannot be written as the sum of local gradients at individual machines that only depend on the local data in those machines. Instead,  the gradient of DXO at each machine not only depends on local data but also on data in other machines. As a result, the design of communication-efficient FL algorithms for DXO is much more complicated than that for ERM.  In addition, the presence of non-linear function $f$ makes the algorithm design and analysis even more challenging than that with linear $f$. There are two levels of coupling in DXO with nonlinear $f$ with one level at the pairwise loss $\ell(h(\w,\z), h(\w,\z'))$ and another level at the non-linear risk of $f(g(\w,\z, \S_2))$, which makes estimation of stochastic gradient more tricky. 

Although DXO can be solved by existing algorithms in a centralized learning setting~\citep{DBLP:conf/nips/HuZCH20,wang2022finite},  extension of the existing algorithms to the FL setting is {\bf non-trivial}. This is different from the extension of centralized algorithms for ERM problems to the FL setting. In the design and analysis of FL algorithms for ERM, the individual machines compute local gradients and update local models and communicate periodically to average models. The rationale of local FL algorithms for ERM is that as long as the gap error between local models and the averaged model is on par with the noise in the stochastic gradients by controlling the communication frequency, the convergence of local FL algorithms will not be sacrificed and is able to enjoy the parallel speed-up of using multiple machines. However, this rationale is not sufficient for developing FL algorithms for DXO optimization due to the challenges mentioned above. 

To address these challenges, we propose two novel FL algorithms named {\bf FeDXL1 and FeDXL2} for  DXO with linear and non-linear $f$, respectively. The main innovation in the algorithm design lies at an active-passive decomposition framework that decouples the gradient of the objective into two types, active parts and passive parts. The active parts depend on data in local machines and the passive parts depend on data in other machines. We estimate the active parts using the local data and the local model and estimate the passive parts using the information with delayed communications from other machines that are computed at historical models in the previous round. In terms of analysis, the challenge is that the model used in the computation of stochastic gradient estimator depends on the (historical) samples  for computing the passive parts at the current iteration, which is only exacerbated in the presence of non-linear function $f$. 
We develop a novel analysis that allows us to transfer the error of the gradient estimator into the latency error of the passive parts and the gap error between local models and the global model. Hence, the rationale is that as long as the latency error of the passive parts and the gap error between local models and the global model is on par with the noise in the stochastic gradient estimator we are able to achieve convergence and linear speed-up.

\begin{figure*}[t]
    \centering
    \includegraphics[width=0.65\textwidth]{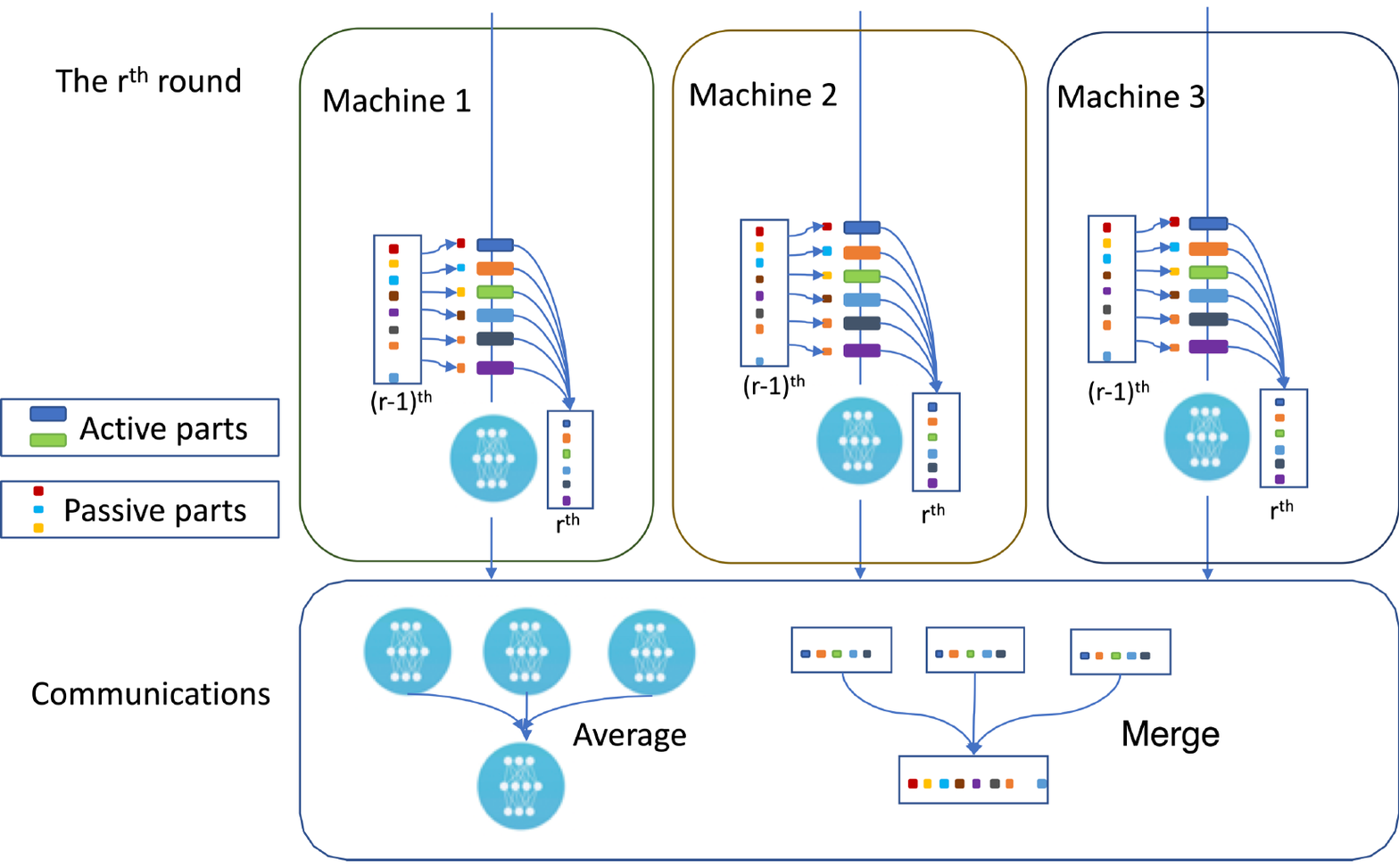}
    \vspace*{-0.2in}
    \caption{Illustration of the proposed Active-Passive Decomposition Framework of FeDXL, which is enabled by Federated Averaging and Merging, where the merged prediction outputs from previous rounds are used for computing the passive parts in stochastic gradient estimator, and its active parts  are computed by using local  model and local data. }
    \label{fig:non}
\end{figure*}
The main contributions of this work are as follows:
\begin{itemize}[leftmargin=*]
    \item We propose two novel communication-efficient algorithms, FeDXL1 and FeDXL2, for DXO with linear and nonlinear $f$, respectively, based on federated averaging and merging. Besides communicating local models for federated averaging, the proposed algorithms need to communicate local prediction outputs only periodically for federated merging to enable the computing of passive parts. The diagram of the proposed FeDXL algorithms is shown in Figure~\ref{fig:non}. 
    \item We perform novel technical analysis to prove the convergence of both algorithms. We show that both algorithms enjoy parallel  speed-up  in terms of the iteration complexity, and a lower-order communication complexity. 
    \item We conduct empirical studies on two tasks for federated deep partial AUC optimization with a compositional loss and federated deep AUC optimization with a pairwise loss, and demonstrate the advantages of the proposed algorithms over several baselines. 
\end{itemize}

\section{Related Work}
\noindent\textbf{FL for ERM.} The challenge of FL is how to utilize the distributed data to learn a ML model  with light communication cost without harming the data privacy~\citep{konevcny2016federated,mcmahan2017communication}. To reduce the communication cost, many algorithms have been proposed to skip communications \citep{stich2018local,yu2019linear,yu2019parallel,yang2013trading,karimireddy2020scaffold} or compress the communicated statistics \citep{stich2018sparsified,basu2019qsparse,jiang2018linear,wangni2018gradient,bernstein2018signsgd}. Tight analysis has been performed in various studies~\citep{kairouz2021advances,yu2019linear,yu2019parallel,khaled2020tighter,woodworth2020minibatch,woodworth2020local,karimireddy2020scaffold,haddadpour2019local}. However,  most of these works target at ERM. 

\noindent\textbf{FL for Non-ERM Problems.} In \citep{guo2020communication,yuan2021federated,deng2021local,deng2020distributionally,liu2020decentralized,sharma2022federated}, federated minimax optimization algorithms have been studied, which are not applicable to our problem when  $f$ is non-convex.   
\citet{gao2022convergence} considered a much simpler federated compositional optimization in the form of $\sum_k \E_{\zeta\sim\mathcal D_f^k}f_k(\E_{\xi\sim\mathcal D_g^k}g_k(\w; \xi); \zeta)$, where $k$ denotes the machine index. Compared with the X-risk, their objective does not involve interdependence between different machines.   \citet{li2022communication,huang2022federated} analyzed FL algorithms for bi-level problems where only the low-level objective involves distribution over many machines. 
\citet{pmlr-v162-tarzanagh22a} considered another federated bilevel problem, where both upper and lower level objective are distributed over many machines, but the lower level objective is not coupled with the data in the upper objective. \citet{9832778} studied a federated bilevel optimization in a server-clients setting, where the central server solves an objective that depends on optimal solutions of local clients. Our problem cannot be mapped into these federated bilevel optimization problems. There are works that optimize non-ERM problems using local data or data from other machines, which are mostly adhoc and lack of theoretical guarantees~\citep{https://doi.org/10.48550/arxiv.2207.09158,DBLP:journals/corr/abs-2010-08982,wu2022federated,li2022fedgrec}. 

\begin{table*}[t] 
	\caption{Comparison for sample complexity on each machine for solving the DXO problem to find an $\epsilon$-stationary point, i.e., $\E[\|F(\w)\|^2] \leq \epsilon^2$. $n$ is the number of finite-sum components in outer finite-sum setting, which is the number of data on the outer function. $n_{\text{in}}$ denotes the number of finite-sum components for the inner function $g$ when it is of finite-sum structure.  In federated learning setting,  $n_i$ denotes the number components in the outer function of machine $i$. }\label{tab:app_survey} 
	\centering 
	\label{tab:0} 
	\scalebox{0.8}{\begin{tabular}{l|c|c|l}  
			\toprule 
% \multicolumn{2}{c|}{\textbf{Finite-Sum}} 
&Method & Sample Complexity & Setting\\
\hline 
 &  BSGD~\citep{DBLP:conf/nips/HuZCH20} & $O(1/\epsilon^6)$ & Inner Expectation  + Outer Expectation\\  
&  BSpiderBoost~\citep{DBLP:conf/nips/HuZCH20}  & $O(1/\epsilon^5)$ & Inner Expectation  + Outer Expectation \\  
\multirow{2}{*}{Centralized}& SOX~\citep{wang2022finite}  & $O(n/\epsilon^4)$& Inner Expectation + Outer Finite-sum \\ 
&  MSVR~\citep{jiang2022multi} & $O(\max(1/\epsilon^4, n/\epsilon^3))$ & Inner Expectation + Outer  Finite-sum \\
& MSVR~\citep{jiang2022multi}&  $O(n\sqrt{n_{\text{in}}}/\epsilon^2)$ & Inner Finite-sum + Outer Finite-sum \\
\hline
 \multirow{1}{*}{Federated} &This Work &  $O(\max_i n_i/\epsilon^4)$ 
& Inner Expectation +  Outer Finite-sum \\ 
\bottomrule
%\hline 
%\hline 
%\multicolumn{2}{c|}{\textbf{Non-Finite-Sum}} & Upper Bound & Lower Bound\\
%\hline 
%\multirow{2}{*}{ERM} & Centralized & & $O(1/\epsilon^4)$ \cite{arjevani2022lower} \\
%& Federated  & $O({1}/{N\epsilon^4})$ \citep{karimireddy2020scaffold} & -- \\
%\hline 
%\multirow{2}{*}{DXO} & Centralized & $O(1/\epsilon^6)$ \citep{DBLP:conf/nips/HuZCH20} & -- \\ 
%& Federated & -- & -- \\ 
%		\bottomrule
	\end{tabular}} 
	%\vspace*{-0.15in} 
\end{table*} 

\noindent\textbf{Centralized Algorithms for DXO.} In the centralized setting  DXO has been considered in recent works~\citep{DBLP:conf/nips/QiLXJY21,DBLP:conf/aistats/0006YZY22,wang2022finite,DBLP:conf/icml/QiuHZZY22}.  
In particular, \citet{wang2022finite} have proposed a stochastic algorithm named SOX for solving~(\ref{eqn:DXO}) and  achieved state-of-the-art sample complexity of $O(|\mathcal S_1|/\epsilon^4)$  to ensure the expected convergence to an $\epsilon$-stationary point. Nevertheless, it is non-trivial to extend the centralized algorithms to the FL setting due to the challenges mentioned earlier. Recently, \cite{jiang2022multi} further proposed an advanced variance-reduction technique named MSVR to improve the sample complexity of solving finite-sum coupled compositional optimization problems.  {We provide a summary of state-of-the-art sample complexities for solving DXO in both centralized and FL setting in Table~\ref{tab:0}.}

\section{FeDXL for DXO}
We assume $\S_1, \S_2$ are split into $N$ non-overlapping subsets that are distributed over $N$ clients~\footnote{We use clients and machines interchangeably.}, i.e., $\S_1=\S_1^1\cup \S_1^2\ldots\cup\S_1^N$ and $\S_2=\S_2^1\cup \S_2^2\ldots\cup\S_2^N$.  We denote by $\E_{\z\sim\S} = \frac{1}{|\S|}\sum_{\z\in\S}$. 
Denote by $\nabla_1\ell(\cdot, \cdot)$ and $\nabla_2\ell(\cdot, \cdot)$ the partial gradients in terms of the first argument and the second argument, respectively. 
Without loss of generality, we assume the dimensionality of $h(\w, \z)$ is 1 (i.e., $d_o=1$) in the following presentation. Notations used in the algorithms are summarized in Appendix \ref{app:notations}.

\subsection{FeDXL1 for DXO with linear $f$}
%With linear $f$, we rewrite the DXO risk into an equivalent form that is tailored to the FL setting: 
We consider the following FL objective for DXO: 
%\vspace{-0.1in} 
\begin{small}
\begin{align}\label{eqn:DXO1}
    \min_{\w\in\R^d}F(\w)=\frac{1}{N}\sum_{i=1}^N\E_{\z\in\S_1^i}\frac{1}{N}\sum_{j=1}^N\E_{\z'\in\S_2^j}\ell(h(\w, \z), h(\w, \z')). 
\end{align}
\end{small}
To highlight the challenge and motivate FeDXL, we decompose the gradient of the objective function into:
\begin{small}
\begin{align*}
&\nabla F(\w)=\\
&\frac{1}{N}\sum_{i=1}^N\underbrace{\E_{\z\in\S_1^i}\frac{1}{N}\sum_{j=1}^N\E_{\z'\in\S_2^j}\nabla_1\ell(h(\w, \z), h(\w, \z'))\nabla h(\w, \z)}\limits_{\Delta_{i1}}\\
&\!+\!\frac{1}{N}\!\sum_{i=1}^N\underbrace{\E_{\z'\in\S_2^i}\frac{1}{N}\sum_{j=1}^N\E_{\z\in\S_1^j}\nabla_2\ell(h(\w,\z), h(\w, \z'))\nabla h(\w,\z')}\limits_{\Delta_{i2}}.
\end{align*}
\end{small} 
Let $\nabla F_i(\w):=\!\Delta_{i,1}\!+\!\Delta_{i,2}$. %Define  {\color{red} Do we need to introduce $\nabla F_i(\w)$ here? I suggest to remove it. otherwise you just need to define $\Delta_{i,1/2}$ once.}
%\begin{align}\label{eqn:app_linear_F_i}
%  \nabla F_i(\w)&:=\underbrace{\E_{\z\in\S_1^i}\frac{1}{N}\sum_{j=1}^N\E_{\z'\in\S_2^j}\nabla_1\ell(h(\w, \z), h(\w, \z'))\nabla h(\w,\z)}\limits_{\Delta_{i1}}\\
%&+\underbrace{\E_{\z'\in\S_2^i}\frac{1}{N}\sum_{j=1}^N\E_{\z\in\S_1^j}\nabla_2\ell(h(\w,\z), h(\w, \z'))\nabla h(\w,\z')}\limits_{\Delta_{i2}}.
%\end{align} 
Then $\nabla F(\w) = \frac{1}{N} \sum\limits_{i=1}^{N} \nabla F_i(\w)$.  

With the above decomposition, we can see that the main task at the local client $i$ is to estimate the gradient terms $\Delta_{i1}$ and $\Delta_{i2}$. Due to the symmetry between $\Delta_{i1}$ and $\Delta_{i2}$, below, we only use $\Delta_{i1}$ as an illustration for explaining the proposed algorithm. The difficulty in computing $\Delta_{i1}$ lies at it relies on data in other machines due to the presence of $\E_{\z'\in\S_{2}^j}$ for all $j$. To overcome this difficulty, we decouple the data-dependent factors in $\Delta_{i1}$ into two types marked by green and blue shown below: 
\begin{small}
\begin{align}\label{eq:G1}
 %\Delta_{i1}=
 \underbrace{\colorbox{green!30}{$\E_{\z\in\S_1^i}$}}\limits_{\text{local1}}\underbrace{\colorbox{blue!30}{$\frac{1}{N}\sum\limits_{j=1}^N\E_{\z'\in\S_2^j}$}}\limits_{\text{global1}}\nabla_1\ell(\underbrace{\colorbox{green!30}{$h(\w, \z)$}}\limits_{\text{local2}}, \underbrace{\colorbox{blue!30}{$h(\w,\z')$}}\limits_{\text{global2}})\underbrace{\colorbox{green!30}{$\nabla h(\w,\z)$}}\limits_{\text{local3}}.
\end{align}
\end{small} 
It is notable that the three green terms can be estimated or computed based the local data. In particular, local1 can be estimated by sampling data from $\S_{1}^i$ and local2 and local3 can be computed based on the sampled data $\z$ and the local model parameter.  The difficulty springs from estimating and computing the two blue terms that depend on data on all machines. %It is not communication efficient that at each iteration all machines or a subset of machines sample data $\z'$ and compute $h_\w(\z')$ and communicate them to all other machines in order to estimate/compute the blue terms above. 
{{\it We would like to avoid communicating $h(\w; \z')$ at every iteration for estimating the blue terms as each communication would incur additional communication overhead. }} To tackle this, we propose to leverage the historical information computed in the previous round~\footnote{A round is defined as a sequence of local updates between two consecutive communications.}. To put this into context of optimization, we consider the update at the $k$-th iteration during the $r$-th round, where $k=0, \ldots, K-1$. Let $\w^r_{i,k}$ denote the local model in $i$-th client at the $k$-th iteration within $r$-th round. Let $\z^r_{i, k,1}\in\S_{1}^i, \z^r_{i,k,2}\in\S_2^i$ denote the data sampled at the $k$-th iteration from $\S_{1}^i$ and $\S_2^i$, respectively.  Each local machine will compute $h(\w^r_{i,k}, \z^r_{i,k,1})$ and $h({\w^r_{i,k}},\z^r_{i,k,2})$, which will be used for computing the active parts. Across all iterations $k=0, \ldots, K-1$, we will accumulate the computed prediction outputs over sampled data and stored in two sets $\mathcal H^r_{i,1}=\{h(\w^r_{i,k}, \z^r_{i,k,1}), k=0, \ldots, K-1\}$ and $\mathcal H^r_{i,2}=\{h(\w^r_{i,k},\z^r_{i,k,2}), k=0, \ldots, K-1\}$. 
At the end of round $r$, we will communicate $\w^r_{i,K}$ and $\mathcal H^r_{i,1}$ and $\mathcal H^r_{i,2}$ to the central server, which will average the local models to get a global model $\w_r$ and also merge $\H^r_1=\H^r_{1,1}\cup \H^r_{2,1}\ldots\cup\H^r_{N,1}$ and $\H^r_2=\H^r_{1,2}\cup \H^r_{2,2}\ldots\cup\H^r_{N,2}$. These merged information will be broadcast to each individual client. Then, at the $k$-th iteration in the $r$-th round, we estimate the blue term by sampling $h^{r-1}_{2, \xi}\in\H^{r-1}_2$ without replacement and compute an estimator of $\Delta_{i1}$ by 
\begin{align}\label{eqn:G1}
    G^r_{i,k,1} =  \nabla_1 \ell(\underbrace{\colorbox{green!30}{$h(\w^r_{i,k}, \z^r_{i,k,1})$}}\limits_{\text{active}},  \underbrace{\colorbox{blue!30}{$h^{r-1}_{2,\xi}$}}\limits_{\text{passive}}) \underbrace{\colorbox{green!30}{$\nabla h(\w^r_{i,k}, \z^r_{i,k,1})$}}\limits_{\text{active}},
\end{align}
where $\xi=(j, t, \z^{r-1}_{j, t, 2})$ represents a random variable that captures the randomness in the sampled client $j\in\{1, \ldots, N\}$, iteration index $k\in\{0, \ldots, K-1\}$ and data sample $\z^{r-1}_{j,t,2}\in\S_{2}^j$, which is used for estimating the global1 in~(\ref{eq:G1}). We refer to the green factors in $G_{i,k,1}$ as the active parts and the blue factor in $G_{i,k,1}$ as the passive part. Similarly, we can estimate $\Delta_{i2}$ by $G_{i,k,2}$ 
\begin{align}\label{eqn:G2}
    G^r_{i,k,2} =  \nabla_2 \ell(\underbrace{\colorbox{blue!30}{$h^{r-1}_{1,\zeta}$}}\limits_{\text{passive}}, \underbrace{\colorbox{green!30}{$h(\w^r_{i,k}, \z^r_{i,k,2})$}}\limits_{\text{active}}) \underbrace{\colorbox{green!30}{$\nabla h(\w^r_{i,k}, \z^r_{i,k,2})$}}\limits_{\text{active}},
\end{align}
where $h^{r-1}_{1,\zeta}\in\H^{r-1}_1$ is a randomly sampled prediction output in the previous round with $\zeta=(j',t', \z^{r-1}_{j', t', 1})$ representing a random variable including a client sample $j'$ and iteration sample $t'$ and the data sample $\z^{r-1}_{j', t', 1}$. Then we will update the local model parameter $\w^r_{i,k}$ by using a gradient estimator $
G^r_{i,k,1}+G^r_{i,k,2}$. 

We present the detailed steps of the proposed algorithm FeDXL1 in Algorithm~\ref{alg:FeDXL1}.  Several remarks are following: (i) at every round, the algorithm needs to communicate both the model parameters $\w^r_{i,K}$ and the historical prediction outputs $\H^{r-1}_{i,1}$ and $\H^{r-1}_{i,2}$, where  $\H^{r-1}_{i,*}$ is constructed by collecting all or sub-sampled computed predictions in the $(r-1)$-th round. The bottom line for constructing $\H^{r-1}_{i,*}$ is to ensure that $\H^{r-1}_{*}$ contains at least $K$ independently sampled predictions that are from the previous round on all machines such that the corresponding data samples involved in $\H^{r-1}_{*}$ can be used to approximate $\frac{1}{N}\sum_{i=1}^N\E_{\z\in\S^i_*}$ $K$ times. Hence, to keep the communication costs minimal, each client at least needs to sample $O(\lceil K/N\rceil )$ sampled predictions from all iterations $k=0, 1, \ldots, K-1$ and send them to the server for constructing $\H^{r-1}_{*}$, which is then broadcast to all clients for computing the passive parts in the round $r$. As a result,  the minimal communication costs per-round per-client is $O(d + Kd_o/N)$. Nevertheless, for simplicity in Algorithm~\ref{alg:FeDXL1} we simply put all historical predictions into $\H^{r-1}_{i,*}$. 

Similar to all other FL algorithms, FeDXL1 does not require communicating the raw input data, hence protects the privacy of the data. However, compared with most FL algorithms for ERM, FeDXL1 for DXO has an additional communication overhead at least $O(d_oK/N)$ which depends on the dimensionality of prediction output $d_o$. For learning a high-dimensional model (e.g. deep neural network with $d\gg 1$) with score-based pairwise losses ($d_o=1$), the additional communication cost $O(K/N)$ could be  marginal.  For updating the buffer $\B_{i,1}$ and $\B_{i,2}$, we can simply flush the history and add the newly received $\mathcal R^{r-1}_{i,1}$ and $\mathcal R^{r-1}_{i,2}$ with random shuffling to $\B_{i,1}$ and $\B_{i,2}$, respectively. 

For analysis, we make the following assumptions regarding the DXO with linear $f$ problem, i.e., problem (\ref{eqn:DXO1}). 
\begin{assumption}~
\begin{itemize}
    \item $\ell(\cdot)$ is differentiable, $L_{\ell}$-smooth and $C_{\ell}$-Lipschitz.
    \item $h(\cdot, \z)$ is differentiable, $L_{h}$-smooth and $C_{h}$-Lipschitz on $\w$ for any $\z \in \mathcal{S}_1 \cup \mathcal{S}_2$.
    \item $\E_{\z\in\mathcal{S}_1^i} \E_{j\in[1:N]} \E_{\z' \in \mathcal{S}_2^j} \|\nabla_1 \ell(h(\w,\z), h(\w,\z'))\nabla h(\w,\z) \\
    \!+\! \nabla_2 \ell(h(\w,\z), h(\w,\z'))\nabla h(\w,\z') \!-\! \nabla F_{i}(\w)  \|^2  \!\leq\! \sigma^2$. 
    \item  $\exists D$ such that $\|\nabla F_{i}(\w) - \nabla F(\w)\|^2 \leq D^2, \forall i$. 
\end{itemize} 
\label{ass:linear}
\end{assumption} 
\begin{algorithm}[H]  %[htbp]   
\caption {FeDXL1: FL for DXO with linear $f$} \label{alg:FeDXL1}
\begin{algorithmic}[1] 
\STATE{On Client $i$: {\bf Require} parameters $\eta, K$} 
\STATE{Initialize model $\w_{i,K}^0$ and  initialize Buffer $\B_{i,1}, \B_{i,2}=\emptyset$} 
\STATE{Sample $K$ points from $\S_1^i$, compute their predictions using model $\w_{i,K}^{0}$ denoted by $\H^{0}_{i,1}$} 
\STATE{Sample $K$ points from $\S_2^i$, compute their predictions using model $\w_{i,K}^{0}$ denoted by $\H^{0}_{i,2}$} 
\FOR{$r=1,..., R$}
\STATE Sends $\w^{r-1}_{i,K}$ to the server
\STATE Receives $\wb^{r}$ from the server and set $\w^{r}_{i,0} = \wb^{r}$ 
\STATE{Send $\H^{r-1}_{i,1}, \H^{r-1}_{i,2}$ to the server}
\STATE Receive $\mathcal R^{r-1}_{i,1}, \mathcal R^{r-1}_{i,2}$ from the server 
\STATE{Update buffer $\B_{i,1}, \B_{i,2}$ using $\mathcal R^{r-1}_{i,1}, \mathcal R^{r-1}_{i,2}$ with shuffling}  \hfill $\diamond$ see text for updating the buffer
\STATE{Set $\H^r_{i,1}=\emptyset$, $\H^r_{i,2}=\emptyset$}
\FOR{$k=0, .., K-1$}
\STATE{Sample $\z^r_{i, k, 1}$ from $\S^i_{1}$, sample $\z^r_{i,k,2}$ from $\S^i_2$} \hfill $\diamond$ or sample two mini-batches of data 
\STATE{Take next  $h^{r-1}_\xi$ and $h^{r-1}_{\zeta}$ from  $\B_{i,1}$ and $\B_{i,2}$, resp.}
\STATE{Compute $h(\w^r_{i,k}, \z^r_{i,k,1})$ and $h(\w^r_{i,k}, \z^r_{i,k,2})$}
\STATE{Add $h(\w^r_{i,k}, \z^r_{i,k,1})$ into $\H^r_{i,1}$ and add $h(\w^r_{i,k}, \z^r_{i,k,2})$ into $\H^r_{i,2}$}
\STATE {Compute $G^r_{i,k,1}$ and $G^r_{i, k, 2}$ according to~(\ref{eqn:G1}) and~(\ref{eqn:G2})}
\STATE{$\w^r_{i, k+1} = \w^r_{i, k} - \eta (G^r_{i,k,1} + G^r_{i,k,2})$} 
\ENDFOR  
\ENDFOR 
\vspace*{0.1in} 
\hrule
\vspace*{0.05in}
\STATE{On Server }
\FOR{$r=1,..., R$} 
\STATE Receive $\w^{r-1}_{i,K}$, from clients $i\in [N]$, compute $\wb^{r} = \frac{1}{N}\sum_{i=1}^N\w^{r}_{i, K}$ and broadcast it to all clients. 
\STATE Collects $\H^{r-1}_{1}=\H^{r-1}_{1,1}\cup\H^{r-1}_{2,1}\ldots\cup\H^{r-1}_{N,1}$ and $\H^{r-1}_{2}=\H^{r-1}_{1,2}\cup\H^{r-1}_{2,2}\ldots\cup\H^{r-1}_{N,2}$
\STATE Set $\mathcal R^{r-1}_{i,1}=\H^{r-1}_1, \mathcal R^{r-1}_{i,2}=\H^{r-1}_2$
\STATE Send $\mathcal R^{r-1}_{i,1}, \mathcal R^{r-1}_{i,2}$ to client $i$ for all $i\in[N]$ 
\ENDFOR  
\end{algorithmic}  
\label{alg:1}
\end{algorithm}

The first three assumptions are standard in the optimization of DXO problems \citep{wang2022finite}. The last assumption embodies the data heterogeneity that is also common in federated learning \citep{yu2019linear,karimireddy2020scaffold}.
Next, we present the theoretical results on the convergence of FeDXL1.

\begin{theorem}
\label{thm:linear_thm1_formal}
Under Assumption \ref{ass:linear}, by setting $\eta = O(\frac{N}{R^{2/3}})$ and $K=O(\frac{R^{1/3}}{N})$, Algorithm \ref{alg:1} ensures that 
\begin{equation}
\E\bigg[\frac{1}{R}\sum_{r=1}^R \|\nabla F(\wb^{r-1})\|^2\bigg] \leq \bigg(\frac{1}{R^{2/3}}\bigg).
\end{equation} 
\end{theorem} 

\textbf{Remark.} To get $\E[\frac{1}{R}\sum_{r=1}^R \|\nabla F(\wb^{r-1})\|^2] \leq \epsilon^2$, we just need to set $R=O(\frac{1}{\epsilon^3})$, $\eta=N\epsilon^2$ and  $K=\frac{1}{N\epsilon}$. The number of communications is much less than the total number of iterations i.e., $O(\frac{1}{N\epsilon^4})$ as long as $N\leq O(\frac{1}{\epsilon})$. And the sample complexity on each machine is $\frac{1}{N\epsilon^4}$, which is linearly reduced by the number of machines $N$. 

{\bf Novelty of Analysis.} As the passive parts are computed in different machines  in a previous round, the gradient estimators $G^r_{i,k,1}$ and $G^r_{i, k, 2}$ will involve the dependency between the local model parameter $\w^r_{i,k}$ and the historical data contained in $\xi, \zeta$ used for computing $G^r_{i,k,1}$ and $G^r_{i, k, 2}$, which makes the analysis more involved.  We need to make sure that using the gradient estimator based on them can still result in ``good" results. To this end, we borrow an analysis technique in \citep{yang2021simple} to decouple the dependence between the current model parameter and the data used for computing the current gradient estimator, in which they used data in previous iteration to couple the data in the current iteration in order to compute a gradient of the pairwise loss $\ell(h(\w_t; \z_t), h(\w_t; \z_{t-1}))$.  Nevertheless, in federated DXO controlling the error  brought by the passive parts is more challenging since the delay is much longer and they were computed on different machines. In our analysis, we replace $\w^r_{i,k}$ with $\bar\w^{r-1}$ to decouple the dependence between the model parameter $\bar\w^{r-1}$ and the historical data $\xi, \zeta$,  then we need to control the latency error $\|\bar\w^{r-1}-\bar\w^r\|^2$ and the gap error between different machines $\sum_i\sum_k\E\|\wb^{r} - \w^r_{i,k}\|^2$ such that the complexities are not compromised. 

\subsection{FeDXL2 for optimizing DXO with nonlinear $f$}
With nonlinear $f$, we consider the following FL problem of DXO minimization, 
\begin{small} 
\begin{equation}
\begin{split}
F(\w) = \frac{1}{N} \sum\limits_{i = 1}^N \E_{\z \in \S^i_1} f\bigg(\underbrace{{\frac{1}{N} \sum\limits_{j = 1}^N \E_{\z' \in \S^j_2} \ell(h(\w,\z),  h(\w,\z'))}}\limits_{g(\w,\z,\S_2)}\bigg). 
\end{split} 
\label{eq:prob_DXO_nonlinear} 
\end{equation} 
\end{small}

We compute the gradient and decompose it into:
\begin{equation}
\begin{split}
&\nabla F(\w) = \frac{1}{N} \sum\limits_{i = 1}^N (\Delta_{i,1} + \Delta_{i,2}),
\end{split} 
\end{equation}  
where
\begin{equation}  
\begin{split} 
 &\Delta_{i,1}=\E_{\z \in \S^i_1}  
 \frac{1}{N}\sum\limits_{j = 1}^N \E_{\z'\in \S^j_2} \bigg[\colorbox{green!15}{$\nabla f(g(\w,\z, \S_2))$} \cdot\\
 &~~~~~~~~~~~~~~~~~~~~~
 \nabla_1 \ell(h(\w,\z), h(\w,\z')) \nabla h(\w,\z) \bigg]\\  
& \Delta_{i,2} = \E_{\z'\in \S^i_2} \frac{1}{N}\sum\limits_{j = 1}^N  \E_{\z \in \S^j_1}\bigg[\colorbox{blue!15}{$\nabla f(g(\w,\z, \S_2))$}\cdot\\
&~~~~~~~~~~~~~~~~~~~~~
\nabla_2 \ell(h(\w,\z),  h(\w,\z')) \nabla h(\w,\z')\bigg]. 
\end{split} 
\end{equation}  

\begin{algorithm}[H]  %[htbp]  
\caption {FeDXL2: Federated Learning for DXO with non-linear $f$} \label{alg:FeDXL2}     
\begin{algorithmic}[H]  
\STATE{On Client $i$: {\bf Require} parameters $\eta, K$} 
\STATE{Initialize model $\w_{i,K}^0$, $\mathcal U_i^{0}=\{u^0(\z)=0,\z\in\S^i_1\}$, $G^0_{i,K}=\mathbf{0}$, and buffer $\B_{i,1}, \B_{i, 2}, \mathcal C_i=\emptyset$}
\STATE{Sample $K$ points from $S_1^i$, compute their predictions using model $\w_{i,K}^{0}$ denoted by $\H^{0}_{i,1}$} 
\STATE{Sample $K$ points from $S_2^i$, compute their predictions using model $\w_{i,K}^{0}$ denoted by $\H^{0}_{i,2}$}
\FOR{$r=1,..., R$} 
\STATE Sends $\w^{r-1}_{i,K}, G^{r-1}_{i,K}$ to the server
\STATE Receives $\wb^{r}, \bar{G}^r$ from the server and set $\w^{r}_{i,0} = \wb^{r}, G^{r}_{i,0}=\bar{G}^r$ 
\STATE{Send $\H^{r-1}_{i,1}, \H^{r-1}_{i,2}, \mathcal U_i^{r-1}$ to the server} 
\STATE{Receive $\mathcal R^{r-1}_{i,1}, \mathcal R^{r-1}_{i,2}, \mathcal P^{r-1}$ from the server}
\STATE {Update  the buffer $\B_{i,1}, \B_{i,2}, \mathcal C_i$ using $\mathcal R^{r-1}_{i,1}, \mathcal R^{r-1}_{i,2}, \mathcal P^{r-1}$ with shuffling, respectively} 
 \STATE{Set $\H^r_{i,1}=\emptyset$, $\H^r_{i,2}=\emptyset, \mathcal U_i^r=\emptyset$}
\FOR{$k=0, .., K-1$}
\STATE{Sample $\z^r_{i, k, 1}$ from $\S^i_{1}$, sample $\z^r_{i,k,2}$ from $\S^i_2$} \hfill $\diamond$ or sample two mini-batches of data 
\STATE{Take next $h^{r-1}_\xi$,  $h^{r-1}_{\zeta}$ and $u^{r-1}_\zeta$ from  $\B_{i,1}$ and $\B_{i,2}$ and $\mathcal C_i$, respectively} 
\STATE{Compute $h(\w^r_{i,k}, \z^r_{i,k,1})$ and $h(\w^r_{i,k}, \z^r_{i,k,2})$} 
\STATE{Compute $h(\w^r_{i,k}, \hat{\z}^r_{i,k,1})$ and $h(\w^r_{i,k}, \hat{\z}^r_{i,k,2})$} and add them to $\H^r_{i,1}, \H^{r}_{i, 2}$, respectively
\STATE Compute $\u^r_{i,k}(\z^r_{i,k,1})$ according to~(\ref{eqn:u}) and add  $\u^r_{i,k}(\hat{\z}^r_{i,k,1})$ to $\mathcal U_i^{r}$
\STATE {Compute $G^r_{i,k,1}$ and $G^r_{i, k, 2}$ according to~(\ref{eqn:G1G2_G1},\ref{eqn:G1G2_G2})}   
\STATE{$G^r_{i,k} = (1-\beta) G^r_{i,k-1} + \beta (G^r_{i,k,1} + G^r_{i, k, 2})$}    
\STATE{$\w^r_{i, k+1} = \w^r_{i, k} - \eta G^r_{i,k}$}
\ENDFOR  
\ENDFOR
\vspace*{0.1in}
\hrule
\vspace*{0.05in}
\STATE{On Server} 
\FOR{$r=1,..., R$}  
\STATE Receive $\w^{r-1}_{i,K}$,$G^{r-1}_{i,K}$ from client $i\in [N]$, compute $\wb^{r} = \frac{1}{N}\sum_{i=1}^N\w^{r}_{i, K}$, $G^{r} = \frac{1}{N}\sum_{i=1}^N G^{r}_{i, K}$ and broadcast them to all clients. 
\STATE Collects $\H^{r-1}_{*}=\H^{r-1}_{1,*}\cup\H^{r-1}_{2,*}\ldots\cup\H^{r-1}_{N,*}$ and $\mathcal U^{r-1}=\mathcal U^{r-1}_{1}\cup\mathcal U^{r-1}_{1}\ldots\cup\mathcal U^{r-1}_{N}$, where $*=1, 2$
\STATE Set $\mathcal R^{r-1}_{i,1}=\H^{r-1}_1, \mathcal R^{r-1}_{i,2}=\H^{r-1}_2, \mathcal P^{r-1}_i=\mathcal U^{r-1}$ and send them to Client $i$ for all $i\in[N]$ 
\ENDFOR  
\end{algorithmic}  
\label{alg:2}
\end{algorithm}  
Let $\nabla F_i(\w) = \Delta_{i,1}+\Delta_{i,2}$. 
Then we have $\nabla F(\w) = \frac{1}{N}\sum\limits_{i=1}^{N} \nabla F_i(\w)$. 

Compared to that in~(\ref{eq:G1}) for DXO with linear $f$, the $\Delta_{i1}$ term above involves another factor $\nabla f(g(\w,\z, \S_2))$, which cannot be computed locally as it depends on $\S_2$ distributed over all machines. 
Similarly, the $\Delta_{i2}$ term above involves another non-locally computable factor $\nabla f(g(\w,\z, \S_2))$. To address the challenge of estimating $g(\w,\z, \S_2)$, we leverage the similar technique in the centralized setting~\citep{wang2022finite} by tracking it using a moving average estimator based on random samples. In a centralized setting, one can maintain and update $\u(\z)$ for estimating $g(\w, \z, \S_2)$ by
\begin{align*}
   \u(\z) \leftarrow (1-\gamma)\u(\z) + \gamma \ell(h(\w, \z), h(\w, \z')),
\end{align*}
where $\z'$ is a random sample from $\S_2$. However, this is not possible in an FL setting as $\S_2$ is distributed over many machines. To tackle this, we leverage the delay communication technique used in the last subsection. At the $k$-th iteration in the $r$-th round, we update $\u(\z^r_{i,k,1})$ for a sampled $\z^r_{i,k,1}$ by 
\begin{equation}\label{eqn:u}
    \u^r_{i, k}(\z^r_{i, k, 1}) = (1-\gamma)\u^r_{i, k}(\z^r_{i, k, 1}) + \gamma \ell(h(\w^r_{i, k}, \z^r_{i,k, 1}), h^{r-1}_{\xi, 2}),
\end{equation}
where $h^{r-1}_{\xi, 2}$ is a random sample from $\H^{r-1}_2$ where $\xi=(j',t',\hat{\z}^{r-1}_{j', t',2})$ captures the randomness in client, iteration index and data sample in the last round. Then, we can use $\nabla f(\u^r_{i,k}(\z^r_{i,k,1}))$ in place of $\nabla f(g(\w^r_{i,k}, \z^r_{i, k, 1}, \S_2))$ for estimating $\Delta_{i1}$. However, it is more nuanced for estimating $\nabla f(g(\w,\z, \S_2))$ in $\Delta_{2i}$ since $\z\in\S^2_j$ is not local random data. To address this, we propose to communicate $\mathcal U^{r-1}=\{\u^{r-1}_{i, k}(\hat{\z}^{r-1}_{i, k, 1}), i\in[N], k\in[K]-1\}$. Then at the $k$-iteration in the $r$-th round of the $i$-th client, we can estimate $\nabla f(g(\w,\z, \S_2))$ with a random sample from $\mathcal U^{r-1}$ denoted by $u^{r-1}_{\zeta}$, where $\zeta=(j', t', \hat{\z}^{r-1}_{j', t',1})$, i.e., by using $\nabla f(\u_\zeta^{r-1})$. Then we estimate $\Delta_{1i}$ and $\Delta_{2i}$ by  
\begin{small}
\begin{equation}\label{eqn:G1G2_G1}
\begin{aligned}
    &G^r_{i,k,1} := \\ 
    &\!\underbrace{\colorbox{green!30}{$\nabla f(\u^r_{i,k}(\z^r_{i,k,1}))$}}\limits_{\text{active}}\nabla_1 \ell(\underbrace{\colorbox{green!30}{$h(\w^r_{i,k}, \z^r_{i,k,1})$}}\limits_{\text{active}}\!,\!  \underbrace{\colorbox{blue!30}{$h^{r-1}_{2,\xi}$}}\limits_{\text{passive}}\!)\! \underbrace{\!\colorbox{green!30}{$\nabla h(\w^r_{i,k}, \z^r_{i,k,1})\!$}}\limits_{\text{active}}\\ 
\end{aligned}
\end{equation} 
\begin{equation}\label{eqn:G1G2_G2} 
\begin{aligned} 
    &G^r_{i,k,2} = \\
    &\underbrace{\colorbox{blue!30}{$\nabla f(\u^{r-1}_\zeta)$}}\limits_{\text{passive}} \nabla_2 \ell(\underbrace{\colorbox{blue!30}{$h^{r-1}_{1,\zeta}$}}\limits_{\text{passive}}, \underbrace{\colorbox{green!30}{$h(\w^r_{i,k}, \z^r_{i,k,2})$}}\limits_{\text{active}}) \underbrace{\colorbox{green!30}{$\nabla h(\w^r_{i,k}, \z^r_{i,k,2})$}}\limits_{\text{active}} 
\end{aligned}
\end{equation} 
\end{small}
where $j,\xi, j', \zeta$ are random variables. Another difference from DXO with linear $f$ is that even in the centralized setting directly using $G^r_{i,k,1}+G^r_{i,k,2}$ will lead to a worse complexity due to that non-linear $f$ make the stochastic gradient estimator biased~\citep{DBLP:journals/mp/WangFL17}. Hence, in order to improve the convergence, we follow existing state-of-the-art algorithms for stochastic compositional optimization~\citep{DBLP:journals/siamjo/GhadimiRW20,wang2022finite} to compute a moving average estimator for the gradient at local machines, i.e., Step 17 in Algorithm~\ref{alg:FeDXL2}. 
 With these changes, we present the detailed steps of FeDXL2 for solving DXO with non-linear $f$ in Algorithm~\ref{alg:FeDXL2}. The buffers $\mathcal B_{i,*}$ and $\mathcal C_i$ are updated similar to that for FeDXL1. Different from FeDXL1, there is an additional communication cost for communicating $\mathcal U_i^{r-1}$ and an additional buffer $\mathcal C_i$ at each local machine to store the received $\mathcal P^{r-1}_i$ from aggregated $\mathcal U^{r-1}$. Nevertheless, these additional costs are marginal compared with communicating $\mathcal H^{r-1}_{*}$ and maintaining the buffer $\mathcal B_{i,*}$.

We make the following assumptions regarding problem (\ref{eq:prob_DXO_nonlinear}). 
% the DXO with non-linear $f$, i.e., problem (\ref{eq:prob_DXO_nonlinear}). 
\begin{assumption}
\begin{itemize}
    \item $\ell(\cdot)$ is differentiable, $L_{\ell}$-smooth and $C_{\ell}$-Lipschitz. $|\ell(\cdot)| \leq C_0$.
     \item $f(\cdot)$ is differentiable, $L_{f}$-smooth and $C_{f}$-Lipschitz. 
    \item $h(\cdot, \z)$ is differentiable, $L_{h}$-smooth and $C_{h}$-Lipschitz on $\w$ for any $\z \in \mathcal{S}_1 \cup \mathcal{S}_2$.
    \item $\E_{\z\in\mathcal{S}_1^i} \E_{j\in[1:N]} \E_{\z' \in \mathcal{S}_2^j} \\
    \|\nabla f(g(\w,\z, \S_2))\nabla_1\ell(h(\w,\z), h(\w,\z'))\nabla h(\w,\z) \\ 
    ~
    + \nabla f(g(\w,\z, \S_2))\nabla_2\ell(h(\w,\z), h(\w,\z'))\nabla h(\w,\z) - \nabla F_{i}(\w)  \|^2  \leq \sigma^2$.
    \item $\exists D$ such that $\|\nabla F_{i}(\w) - \nabla F(\w)\|^2 \leq D^2, \forall i$. 
\end{itemize} 
\label{ass:non_linear} 
\end{assumption} 

We present the convergence result of FeDXL2 below.
\begin{theorem}
Under Assumption \ref{ass:non_linear},
denoting $M = \max_i |\S^1_i|$ as the largest number of data on a single machine, by setting $\gamma=O(\frac{M^{1/3}}{R^{2/3}})$, $\beta=O(\frac{1}{M^{1/6} R^{2/3}})$, $\eta = O(\frac{1}{M^{2/3} R^{2/3}})$ and $K=O(M^{1/3} R^{1/3})$,  Algorithm \ref{alg:2} ensures that 
$$\frac{1}{R}\sum_{r=1}^R \E\|\nabla F(\wb^{r})\|^2 \leq O(\frac{1}{R^{2/3}}).$$
\label{thm:nonlinear_informal}
\end{theorem} 
\vspace{-0.2in}
\textbf{Remark.} To get $\E[\frac{1}{R}\sum_{r=1}^R \|\nabla F(\wb^{r})\|^2] \leq \epsilon^2$, we just set $R=O(\frac{M^{1/2}}{\epsilon^3})$,  $\eta=O(\frac{\epsilon^2}{M})$, $\gamma=O(\epsilon^2)$, $\beta=\frac{\epsilon^2}{\sqrt{M}}$ and  $K=\frac{M^{1/2}}{\epsilon}$. 
The number of communications $R=O(\frac{M^{1/2}}{\epsilon^3})$ is less than the total number of iterations i.e., $O(\frac{M}{\epsilon^4})$ by a factor of $ O(M^{1/2}/\epsilon)$. And the sample complexity on each machine is $\frac{M}{\epsilon^4}$, which is less than that in \cite{wang2022finite} which has a sample complexity of $O(\sum\nolimits_{i=1}^{N} |\mathcal{S}^1_i|/\epsilon^4)$. When the data are evenly distributed on different machines, we have achieved a linear speedup property.  
And in an extreme case where all data are on one machine, the sample complexity of FeDXL2  matches that in  \citep{wang2022finite}, which is expected.
Compared with FeDXL1, the analysis of FeDXL2 has to deal with extra difficulties. 
First, with non-linear $f$, the coupling between the inner function and outer function adds to the complexity of interdependence between different rounds and machines. 
Second, we have to deal with the error for the passive part related to $\u$.  

Our analysis for FeDXL2 with moving average gradient estimator is different from previous studies for local momentum methods for ERM problems\citep{yu2019linear,karimireddy2020mime}, 
which used a fixed momentum parameter. In contrast, in FeDXL2 the momentum parameter $\beta$ is decreasing as $R$ increases, which is similar to centralized algorithms compositional problems~\citep{DBLP:journals/siamjo/GhadimiRW20,wang2022finite}.

\begin{table*}[!htbp] 
	\caption{Comparison for Federated Deep Partial AUC Maximization. All reported results are partial AUC scores on testing data. 
}\label{tab:exp_pauc_1} 
	\centering 
	\scalebox{0.7}{\begin{tabular}{l c|c||c| c|c|c} 
			\toprule 
{$K=32$, $N=16$}&&\cellcolor{gray!30}\makecell{Centralized \\(OPAUC Loss)} &
\makecell{Local SGD \\ (CE Loss)} & \makecell{CODASCA \\ (Min-Max AUC)} & \makecell{Local Pair \\ (OPAUC Loss)} & \makecell{FeDXL2 \\ (OPAUC Loss)} \\
		\hline  
\multirow{2}{*}{Cifar10}  
& FPR $\leq$ 0.3 &  \cellcolor{gray!30}{0.7655$\pm$0.0039}& 0.6825$\pm$0.0047&0.7288$\pm$0.0035
& 0.7487$\pm$0.0059
&{\bf 0.7580$\pm$0.0034}\\ 
& FPR $\leq$ 0.5 &\cellcolor{gray!30}{0.8032$\pm$0.0039}& 0.7279$\pm$0.0050&0.7702$\pm$0.0029 
&0.7888$\pm$0.0052
&{\bf 0.7978$\pm$0.0026}\\
\hline 
\multirow{2}{*}{Cifar100} 
& FPR $\leq$ 0.3   &\cellcolor{gray!30}{0.6287$\pm$0.0037} & 0.5875$\pm$0.0016& 0.6131$\pm$0.0054 
&0.6281$\pm$0.0032&{\bf 0.6332$\pm$0.0024}\\ 
& FPR $\leq$ 0.5 & \cellcolor{gray!30}{0.6487$\pm$0.0026}
& 0.6124$\pm$0.0021& 0.6406$\pm$0.0041
&0.6569$\pm$0.0017&{\bf 0.6623$\pm$0.0022}\\
\hline 
\multirow{2}{*}{CheXpert} 
& FPR $\leq$ 0.3 & \cellcolor{gray!30}{0.7220$\pm$0.0035} &0.6495$\pm$0.0039& 0.6903$\pm$0.0059 
&0.6902$\pm$0.0053&{\bf 0.7344$\pm$0.0042} \\ 
& FPR $\leq$ 0.5 & \cellcolor{gray!30}{0.7861$\pm$0.0040} &0.7017$\pm$0.0042& 0.7770$\pm$0.0071 
&0.7483$\pm$0.0033&{\bf 0.7918$\pm$0.0037}\\
\hline 
\multirow{2}{*}{ChestMNIST} 
& FPR $\leq$ 0.3 & \cellcolor{gray!30}{0.6344$\pm$0.0053} &0.5904$\pm$0.0012 & 0.6071$\pm$0.0040  &0.5802$\pm$0.0039 &{\bf 0.6228$\pm$0.0048}  \\ 
& FPR $\leq$ 0.5 & \cellcolor{gray!30}{0.6622$\pm$0.0029} &0.6072$\pm$0.0034 
& 0.6272$\pm$0.0038  &0.6026$\pm$0.0025
&{\bf 0.6490$\pm$0.0039}\\
		\bottomrule 
	\end{tabular}} 
	\caption{Comparison for Federated Deep AUC maximization under corrupted labels.  All reported results are  AUC scores on testing data. }\label{tab:exp_auc_1} 
	\centering 
	\scalebox{0.75}{\begin{tabular}{l|c||c|c|c|c} 
	\toprule 
		$K=32$, $N=16$
		&\cellcolor{gray!30}{\makecell{~Centralized~~~~\\(PSM Loss)}}& \makecell{Local SGD \\ (CE Loss)} & \makecell{CODASCA \\ ( Min-Max AUC)} & \makecell{Local Pair \\ (PSM Loss)} & \makecell{FeDXL1 \\ (PSM Loss)} \\ 
		\hline 
{Cifar10} &\cellcolor{gray!30}{0.7352$\pm$0.0043}
&0.6501$\pm$0.0024&0.6407$\pm$0.0044
&0.7287$\pm$0.0027
&{\bf 0.7344$\pm$0.0038}\\
\hline 
{Cifar100} &\cellcolor{gray!30}{0.6114$\pm$0.0038}
&0.5700$\pm$0.0031&0.5950$\pm$0.0039
&0.6175$\pm$0.0045
&{\bf 0.6208$\pm$0.0041}\\ 
\hline {CheXpert} &\cellcolor{gray!30}{0.8149$\pm$0.0031}
&0.6782$\pm$0.0032&0.7062$\pm$0.0085
&0.7924$\pm$0.0043
&{\bf 0.8431$\pm$0.0027} \\
\hline 
{ChestMNIST}  
&\cellcolor{gray!30}{0.7227$\pm$0.0026} &0.5642$\pm$0.0041 &0.6509$\pm$0.0033 &0.6766$\pm$0.0019&{\bf 0.6925$\pm$0.0030}  \\
		\bottomrule 
	\end{tabular}} 
\end{table*} 
%\vspace{-0.1in} 
\section{Experiments} 
\label{sec:experiment} 
To verify our theories, we experiment on two tasks: federated deep partial AUC maximization and  federated deep AUC maximization with a pairwise surrogate loss, which corresponds to~(\ref{eqn:DXO}) with non-linear and linear $f$, respectively. Code is released at \url{https://github.com/Optimization-AI/ICML2023_FeDXL}.

\noindent{\bf Datasets and Neural Networks.} We use four datasets: Cifar10, Cifar100 \citep{krizhevsky2009learning}, CheXpert \citep{DBLP:conf/aaai/IrvinRKYCCMHBSS19}, and ChestMNIST \citep{yang2021medmnist}, where the latter two datasets are large-scale medical image data. For Cifar10 and Cifar100, we sample 20\% of the training data as validation set, and construct imbalanced binary versions with positive:negative = 1:5 in the training set similar to \citep{DBLP:conf/iccv/Yuan0SY21}. For CheXpert, we consider the task of predicting Consolidation and use the last 1000 images in the training set as the validation set and use the original validation set as the testing set. For ChestMNIST, we consider the task of Mass prediction and use the provided train/valid/test split. We distribute training data to $N=16$ machines unless specified otherwise. To increase the heterogeneity of data on different machines,  we add random Gaussian noise of $\mathcal{N}(\mu,0.04)$ to all training images, where $\mu \in \{-0.08:0.01:0.08\}$ that varies on different machines, i.e., for the $i$-th machine out of the $N=16$ machines, its $\mu=-0.08+i*0.01$. We train ResNet18 from scratch for CIFAR-10 and CIFAR-100 data, and initialize DenseNet121 by an ImageNet pretrained model for CheXpert and ChestMNIST. We use the PyTorch framework \citep{paszke2019pytorch}.

\noindent{\bf Baselines.} We compare our algorithms with three local baselines: 1) \textit{Local SGD} which optimizes a Cross-Entropy loss using classical local SGD algorithm; 2) \textit{CODASCA} - a state-of-the-art FL algorithm for optimizing a min-max formulated AUC  loss \citep{yuan2021federated};  and 3) \textit{Local Pair} which optimizes the X-risk using only local pairs. As a reference, we also compare with  the \textit{Centralized} methods, i.e., mini-batch SGD for DXO with linear $f$ and SOX for DXO with non-linear $f$.  %For each algorithm, 
We tune the initial step size in $[1e^{-3}, 1]$ using grid search
and decay it by a factor of 0.1 every 5K iterations.
All algorithms are run for 20k iterations. The mini-batch sizes $B_1, B_2$ (as in Step 11 of FeDXL1 and FeDXL2) are set to  32. The $\beta$ parameter of FeDXL2 (and corresponding Local Pair and Centralized method) is set to $0.1$.  In the Centralized method, we tune the batch size $B_1$ and $B_2$ from $\{32,64,128, 256,512\}$ in an effort to benchmark the best performance.% of the centralized setting. 
For CODASCA and Local SGD which are not using pairwise losses, we set the batch size to 64 for fair comparison with FeDXL. For all the non-centralized algorithms, we set the communication interval $K=32$ unless specified otherwise. In every run, we use the validation set to select the best performing model and finally use the selected model to evaluate on the testing set. For each algorithm, we repeat 3 times with different random seeds and report the averaged performance.

\noindent{\bf FeDXL2 for Federated Deep Partial AUC Maximization.}\\
We consider the task of one way partial AUC maximization, which refers to the area under the ROC curve with false positive rate (FPR) restricted to be less than a threshold. We consider the  KL-OPAUC loss function proposed in \citep{DBLP:journals/corr/abs-2203-00176}, 
%\vspace{-0.1in} 
\begin{small} 
\begin{equation} 
    \min_{\w\in\R^d}\frac{1}{N} \sum_{i=1}^N\E_{\z\in\S^i_1}\lambda \log\bigg(\frac{1}{N} \sum_{j=1}^N\E_{\z'\in\S^j_2}\ell(\w,\z,\z')\bigg),
\end{equation} 
\end{small} 
%\vspace{-0.05in} 
where $\S_1^i$ denotes the set of positive data, $\S_2^j$ denotes the set of negative data and $\ell(\w, \z, \z') = \exp((h(\w,\z)+1-h(\w,\z'))_{+}^2/\lambda)$ where $\lambda$ is a parameter tuned in $[1:5]$.
The experimental results are reported in Table \ref{tab:exp_pauc_1}.  We can see: (i) FeDXL2 is better than all local methods (i.e., Local SGD, Local Pair and CODASCA), and achieves competitive performance as the Centralized method, which indicates the our algorithm can effectively utilize data on all machines. The better performance of FeDXL2 on CIFAR100 and CheXpert than the Centralized method is probably due to that the Centralized method may overfit the training data;  (ii) FeDXL2 is better than the Local Pair method, which implies that using data pairs from all machines are helpful for improving the performance in terms of partial AUC maximization; and (iii) FeDXL2 is better than CODASCA, which is not surprising since CODASCA is designed to optimize AUC loss, while FeDXL2 is used to  optimize partial AUC loss. 

\noindent{\bf FeDXL1 for Federated Deep AUC maximization with Corrupted Labels.} Second, we consider  the task of federated deep AUC maximization. Since deep AUC maximization for solving a min-max loss (an equivalent form for the pairwise square loss) has been developed in previous works~\citep{yuan2021federated}, we aim to justify the benefit of using the general pairwise loss formulation. According to~\citep{DBLP:conf/icml/CharoenphakdeeL19}, a symmetric loss can be more robust to data with corrupted labels for AUC maximization, where a symmetric loss is one such that $\ell(z) +  \ell(-z)$ is a constant. 
Since the square loss is not symmetric, we conjecture that that min-max federated deep AUC maximization algorithm  CODASCA is not robust to the noise in labels. In contrast, our algorithm FeDXL1 can optimize a symmetric pairwise loss; hence we expect FeDXL1 is better than CODASCA in the presence of corrupted labels. 
To verify this hypothesis, we generate corrupted data by  flipping the labels of  20\% of both the positive and negative training data.  We use FeDXL1/Local Pair to optimize the symmetric pairwise sigmoid (PSM) loss \citep{DBLP:conf/pkdd/CaldersJ07}, which corresponds to~(\ref{eqn:DXO}) with linear $f(s) = s$ and $\ell(a, b) = (1+\exp(a-b))^{-1}$, where $a$ is a positive data score and $b$ is a negative data score. 
Specifically,
\begin{equation*} 
    \min_{\w\in\R^d}\frac{1}{N} \sum_{i=1}^N\E_{\z\in\S^i_1}\frac{1}{N} \sum_{j=1}^N\E_{\z'\in\S^j_2}\ell(h(\w,\z),h(\w,\z')),
\end{equation*} 
where $\S_1^i$ denotes the set of positive data, $\S_2^j$ denotes the set of negative data and $\ell(h(\w,\z),h(\w,\z')) = (1+\exp(h(\w,\z)-h(\w,\z')))^{-1}$. 
The results are reported in Table \ref{tab:exp_auc_1}. We observe that FeDXL1 is more robust to label noises compared to other local methods, including Local SGD, Local Pair, and CODASCA that optimizes a min-max AUC loss.  As before, FeDXL1 has competitive  performance compared with the Centralized method. 

The running time comparison, statistics of data,
and ablation studies are in Appendix \ref{Appendix:experiment}. 
% with corrupted labels

%\vspace{-0.1in}
\section{Conclusion} 
We have considered federated learning (FL) for deep X-risk optimization. We have developed communication-efficient FL algorithms to alleviate the interdependence between different machines. Novel convergence analysis is performed to address the technical challenges and to improve both iteration and communication complexities of proposed algorithms. We have conducted empirical studies of the proposed FL algorithms for solving deep partial AUC maximization and deep AUC maximization and achieved promising results compared with several baselines. 

%\vspace{-0.2in} 
\section{Limitations and Potential Negative Societal Impacts}
%Our algorithms have the potential to solve the DXO problems in a federated setting. 
While the current communication complexity is 
$O(1/\epsilon^3)$, there may still be room for improvement to further reduce the communication cost because the state-of-the-art communication complexity for federated ERM problems is $O(1/\epsilon^2)$. Our experimental results indicate that FeDXL may offer better generalization performance than centralized algorithms. However, a more rigorous analysis is necessary to better understand this phenomenon and leverage it effectively. While this work has verified the performance of FeDXL on partial AUC maximization and AUC maximization problems, more studies are needed to test FeDXL on other federated DXO problems and beyond. We do not see any potential negative societal impact.

%\vspace{-0.1in} 
\section*{Acknowledgements}
We appreciate the feedback provided by the anonymous reviewers. 
This work has been partially supported by NSF Career Award 2246753, NSF Grant 2246757 and NSF Grant 2246756. 

\bibliography{ref}
\bibliographystyle{icml2023}

%%%%%%%%%%%%%%%%%%%%%%%%%%%%%%%%%%%%%%%%%%%%%%%%%%%%%%%%%%%%%%%%%%%%%%%%%%%%%%%
%%%%%%%%%%%%%%%%%%%%%%%%%%%%%%%%%%%%%%%%%%%%%%%%%%%%%%%%%%%%%%%%%%%%%%%%%%%%%%%
% APPENDIX
%%%%%%%%%%%%%%%%%%%%%%%%%%%%%%%%%%%%%%%%%%%%%%%%%%%%%%%%%%%%%%%%%%%%%%%%%%%%%%%
%%%%%%%%%%%%%%%%%%%%%%%%%%%%%%%%%%%%%%%%%%%%%%%%%%%%%%%%%%%%%%%%%%%%%%%%%%%%%%%
\newpage
\appendix
\onecolumn
\section{Notations}
\label{app:notations}
\begin{table*}[!htbp] 
	\caption{Notations}\label{tab:notations} 
	\centering 
	\scalebox{0.7}{\begin{tabular}{l l} 
			\toprule 
$\w$& Model parameters of the neural network, variables to be trained  \\
$\w^r_{i,k}$&Model parameters of machine $i$ at round $r$, iteration $k$ \\
$\z$&A data point\\
$\z_i$& A data point from machine $i$\\
$\z^r_{i,k}$& A data point sampled on machine $i$, at round $r$ iteration $k$ \\
$\z^r_{i,k,1}, \z^r_{i,k,2}$& Two independent data points sampled on machine $i$, at round $r$ iteration $k$ \\
$h(\w, \z)$& The prediction score of data $\z$ by network $\w$\\
$G^r_{i,k,1}, G^r_{i,k,2}$ & Stochastic estimators of components of gradient\\
$\H^r_{i,1}, \H^r_{i,2}$ & Collected historical prediction scores on machine $i$ at round $r$ \\
$\u(\z)$ & Moving average estimator of the inner function $g(\w, \z, \S_2)$\\
$\u^r_{i,k}(\z)$ & Moving average estimator of the inner function $g(\w, \z, \S_2)$ on machine $i$ at round $r$, iteration $k$ \\
$\mathcal U^r_{i}$ & Collected historical $\u$ on machine $i$ at round $r$  \\ 
$h^{r-1}_{\epsilon}, h^{r-1}_{\zeta}$ & Predictions scores sampled from the collected scores of round $r-1$\\
$u^{r-1}_{\zeta}$ & Moving average estimator sampled from the collected moving average estimator of round $r-1$ \\
		\bottomrule 
	\end{tabular}} 
\end{table*} 

\section{Applications of DXO Problems}  
\label{app:applicaitons}
% \vspace*{-0.05in}\noindent{\bf Applications of DXO Minimization.} 
We now present some concrete applications of the DXO  problems, including AUROC maximization, partial AUROC maximization and AUPRC maximization. A more comprehensive list of DXO  problems is discussed in the Intrduction section and can also be found in a recent survey \citep{DBLP:journals/corr/abs-2206-00439}. 
%\begin{enumerate}

     \underline{\it AUROC Maximization} The area under ROC curve (AUROC) is defined \citep{hanley1982meaning} as 
    \begin{equation}
        \text{AUROC}(\w) = \E[\I(h(\w, \z) \geq h(\w, \z'))|y=+1, y'=-1],
    \end{equation}
    where $\z, \z'$ are a pair of data features and $y,y'$ are the corresponding labels.
    To maximize the AUROC, there are a number of surrogate losses $\ell(\cdot)$, e.g. $\ell(\w; \z, \z') = (1-h(\w,\z) + h(\w,\z'))^2$, that have proposed in the literature \citep{DBLP:conf/icml/GaoJZZ13,DBLP:conf/icml/ZhaoHJY11,DBLP:conf/ijcai/GaoZ15,DBLP:conf/pkdd/CaldersJ07,DBLP:conf/icml/CharoenphakdeeL19,yang2021simple}, which formulates the problem into
    \begin{equation}
    \begin{split}
        \min\limits_{\w} \frac{1}{|\S_1|}  \sum\limits_{\z_i\in S_1} \frac{1}{|\S_2|}
         \sum\limits_{\z_j\in S_2} \ell(\w, \z_i, \z_j),
    \end{split}
    \end{equation}
    where $\S_1$ is the set of data with positive labels and $\S_2$ is the set of data with negative labels. This is a DXO problem of (\ref{eqn:DXO}) with $f(x)=x$.
    
    \underline{\it Partial AUROC Maximization} In medical diagnosis,  high false positive rates (FPR) and low true positive rates (TPR) may cause a large cost. To alleviate this, we will also consider optimizing partial AUC (pAUC). This task considers to maximize the area under ROC curve with the restriction that the false positive rate to be less than a certain level. 
    In \citep{DBLP:journals/corr/abs-2203-00176}, it has been shown that the partial AUROC maximization problem can be solved by the 
    \begin{equation}
    \begin{split}
        \min_{\w} \frac{1}{|\S_1|} \sum\limits_{\x_i\in \S_1} \lambda \log \left(\frac{1}{|\S_2|}\sum\limits_{\z_j \in \S_2} \exp(\frac{\tilde{\ell}(\w, \z_i, \z_j)}{\lambda}) \right),
    \end{split}
    \end{equation}
    where $\S_1$ is the set of positive data, $\S_2$ is the set of negative data, $\tilde{\ell}(\cdot)$ is surrogate loss, and $\lambda$ is associated with the tolerance level of false positive rate. This is a DXO problem of (\ref{eqn:DXO}) with $f(x) = \lambda \log(x)$, and $\ell(\w, \z_i, \z_j) = \exp(\frac{\tilde{\ell}(\w, \z_i, \z_j)}{\lambda})$.
   
    \underline{\it AUPRC Maximization} According to \citep{boyd2013area}, the area under the precision-recall curve (AUPRC) can be approximated by 
    \begin{equation} 
    \begin{split} 
        \frac{1}{|\S|} \sum\limits_{(\z_i, y_i)\in \S} \I(y_i=1) \frac{\sum\limits_{(\z_j, y_j)\in \S} \I(y_j=1) \I(h(\w, \z_i) \geq h(\w, \z_j))}{\sum\limits_{(\z_j, y_j)\in \S} \I(h(\w, \z_i) \geq h(\w, \z_j))}. 
    \end{split}
    \end{equation}
    Then using a surrogate loss, the AUPRC maximization problem becomes 
    \begin{equation}
    \begin{split}
         \min\limits_{\w} -\frac{1}{|\S|} \sum\limits_{(\z_i, y_i)\in \S} \I(y_i=1) \frac{\sum\limits_{(\z_j, y_j)\in \S} \I(y_j=1) \tilde{\ell}(\w, \z_i, \z_j))}{\sum\limits_{(\z_j, y_j)\in \S} \tilde{\ell}(\w, \z_i, \z_j)},
    \end{split} 
    \end{equation} 
    which is a DXO problem of (\ref{eqn:DXO}) with $\ell(\w, \z_i, \z_j) = 
    [(\I_{y_j=1}) \tilde{\ell}(\w, \z_i, \z_j), \tilde{\ell}(\w, \z_i, \z_j)]$ and $f(x_1, x_2) = \frac{x_1}{x_2}$ \citep{DBLP:conf/nips/QiLXJY21}. 

\section{Experiments}
\label{Appendix:experiment}
\subsection{\bf Statistics of Data}
Statistics of used data sets are summarized in Table \ref{tab:data_stats}. 

\begin{table*}[tbph] 
	\caption{Statistics of the Datasets 
}\label{tab:data_stats} 
\vspace{0.1in}
	\centering 
	%\label{tab:2}  
	\scalebox{1}{\begin{tabular}{l|c|c|c} 
	\toprule 
	& \# of Training Data & \# of Validation Data & \# of Testing Data  \\
	\hline
	%\hline
	Cifar10    &24000 &10000 &10000 \\
	\hline 
    Cifar100   &24000 &10000 &10000 \\
	\hline 
	CheXpert   &190027 &1000&202 \\
	\hline
	ChestMNIST &78468&11219&22433\\
	\bottomrule
	\end{tabular}} 
\end{table*}

\subsection{Running Time Comparison} 
Running time is reported in Tabel \ref{tab:runningtime}.
Each algorithm was run on 16 client machines connected by InfiniBand where each machine uses a NVIDIA A100 GPU. 

\begin{table*}[tbph] 
	\caption{Running time comparison of federated algorithm on partial AUC maximization task in \ref{sec:experiment}. We report the average number of communication rounds and runtime (in seconds) for each algorithm to converge to a region that for $\text{FR}\leq0.5$, the training pAUC $\geq$ its best training pAUC$-$0.01.  
}\label{tab:runningtime} 
	\centering 
	\scalebox{0.7}{\begin{tabular}{l |c|c|c| c} 
			\toprule 
 &
\makecell{Local SGD \\ (CE Loss)} & \makecell{CODASCA \\ (Min-Max AUC)} & \makecell{Local Pair \\ (OPAUC Loss)} & \makecell{FeDXL2 \\ (OPAUC Loss)} \\
		\hline  
{Cifar10}  
& 157 (664s) &	147 (955s) &168 (740s)&160 (819s) \\ 
\hline
{Cifar100} 
& 160 (644s)& 163 (974s)&162 (688s)&159 (758s)\\ 
\hline 
{CheXpert} 
& 162 (2465s)&151 (3501s)&175 (2838s)&182 (3246s)\\ 
\hline 
{ChestMNIST} 
&172 (1537s)&165 (3176s)&164 (1484s)&171 (1763s)\\  
\bottomrule 
	\end{tabular}} 
\end{table*}

\subsection{Ablation Study.} We show an ablation study to further verify our theory. In particular, we show the benefit of using multiple machines and the lower communication complexity by using $K>1$ local updates between two communications. To verify the first effect, we fix $K$ and vary $N$, and for the latter  we fix $N$ and vary $K$. We conduct experiments on the CIFAR-10 data for optimizing the X risk corresponding to partial AUC loss and the results are plotted in Figure \ref{fig:ablation_vary}. The left two figures demonstrate that our algorithm can tolerate a certain value of $K$ for skipping communications without harming the performance; and the right two figures demonstrate the advantage of FL by using FeDXL2, i.e.,  using data from more sources can dramatically improve the performance. 

\begin{figure}[H]
   \vspace*{-0.1in}   \centering
    \hspace*{-0.1in} \includegraphics[width=0.24\textwidth]{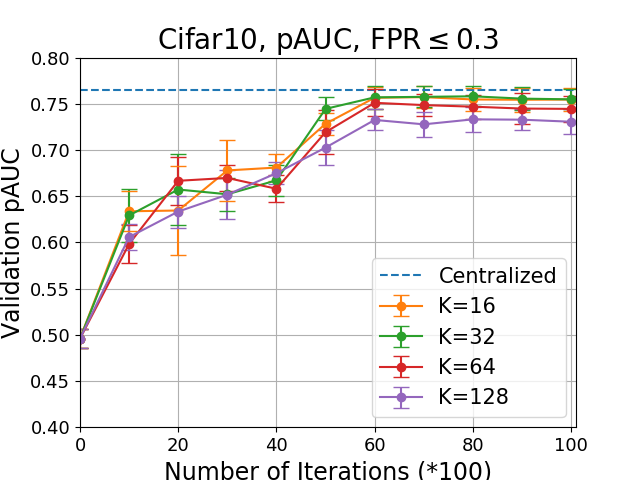} 
     \includegraphics[width=0.24\textwidth]{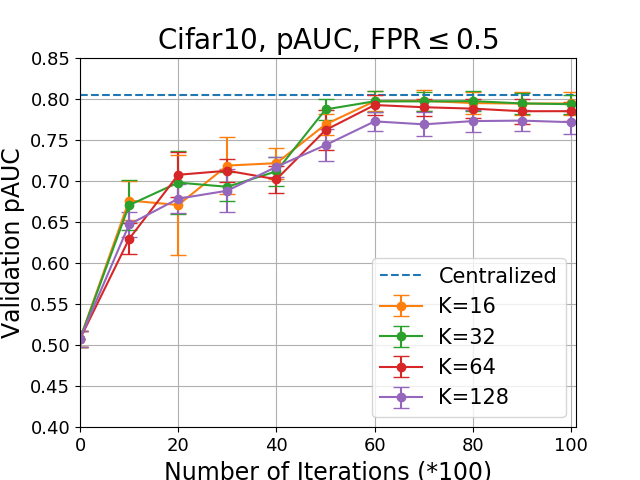} 
    \centering 
    \hspace{-0.1in} \includegraphics[width=0.23\textwidth]{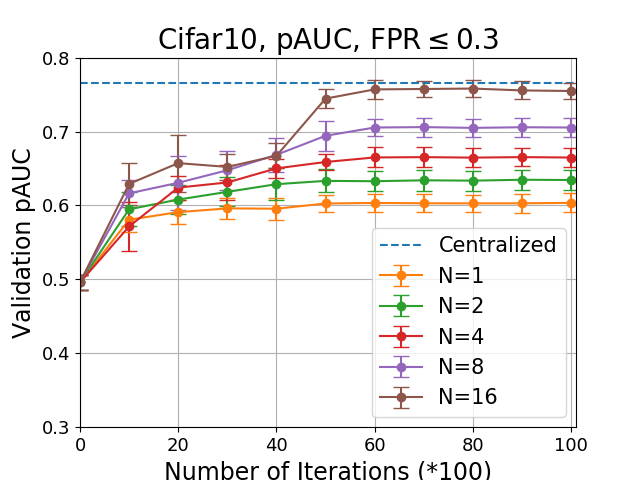}
     \includegraphics[width=0.23\textwidth]{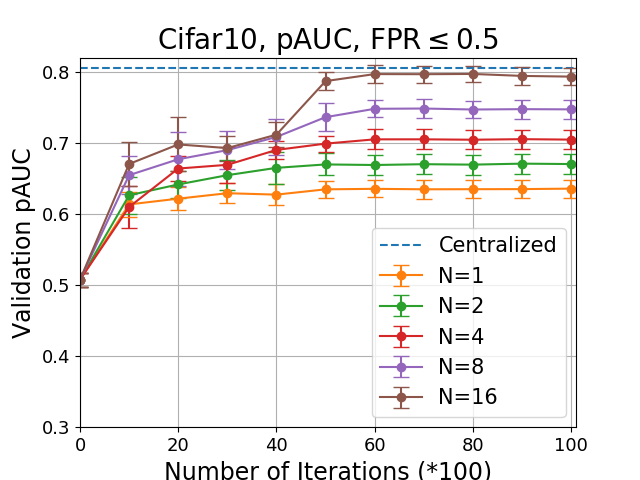} 
     \vspace*{-0.1in}
    \caption{Ablation study: Left two: Fix $N$ and Vary $K$; Right two: Fix $K$ and Vary $N$}  
    \label{fig:ablation_vary} 
         \vspace*{0.05in}
\end{figure}

\section{Analysis of FeDXL1 for solving DXO with Linear $f$}  
\label{app:sec:FeDXL}
In this section, we present the analysis of the FeDXL1 algorithm. For $\z\in \S_1^i$ and $\z'\in \S_2^j$, we define 
\begin{equation}
\begin{split}
&G_1(\w, \z, \w', \z') = \nabla_1 \ell(h(\w,\z), h(\w,\z'))^\top \nabla h(\w,\z)\\
&G_2(\w, \z, \w', \z') = \nabla_2 \ell(h(\w, \z), h(\w,\z'))^\top \nabla h(\w,\z').
\end{split}    
\end{equation} 
Therefore, the 
\begin{align*} 
    G^r_{i,k,1} =  \nabla_1  \ell(h(\w^r_{i,k}, \z^r_{i,k,1}),  h^{r-1}_{2,\xi}) \nabla h(\w^r_{i,k}, \z^r_{i,k,1}), 
\end{align*} 
defined in (\ref{eq:G1}) is equivalent to $G_1(\w^r_{i,k}, \z^r_{i,k,1}, \w^{r-1}_{j,t},\z^{r-1}_{j,t,2})$, 
where $h^{r-1}_{2, \xi} = h(\w^{r-1}_{j,t}, \z^{r-1}_{j,t,2})$ is a scored of a randomly sampled data that in computed in the round $r-1$ at machine $j$ and iteration $t$. Technically, notations $j$ and $t$ are associated with $i$ and $k$, but we omit this dependence when the context is clear to simplify notations.

Similarly,
the  
\begin{align*}
    G^r_{i,k,2} =  \nabla_2 \ell(h^{r-1}_{1,\zeta}, h(\w^r_{i,k}, \z^r_{i,k,2}),  \nabla h(\w^r_{i,k}, \z^r_{i,k,2}) ),
\end{align*} 
defined in (\ref{eqn:G2}) is equivalent to $G_2(\w^{r-1}_{j',t'}, \z^{r-1}_{j',t',1}, \w^r_{i,k}, \z^r_{i,k,2})$. 
\vspace{-0.1in} 
\begin{proof} 
Under Assumption \ref{ass:linear}, it follows that $F(\cdot)$ is $L_F$-smooth, with $L_F:= 2(L_\ell C_h + C_\ell L_h)$. Simiarly, $G_1, G_2$ also Lipschtz in $\w$ and $\w'$ with some constant $L_1$ that depend on $C_h, C_\ell, L_\ell, L_h$. Let $\tL:=\max\{L_F, L_1\}$. 

Denote $\teta=\eta K$ and suppose $\teta \tL \leq O(1)$by proper setting of $\eta$ and $K$. Using the $\tL$-smoothness of $F(\w)$, we have 
\begin{small}
\begin{equation} 
\begin{split}
&F(\wb^{r+1}) - F(\wb^r) \leq \nabla F(\wb^r)^\top (\wb^{r+1} - \wb^r) + \frac{\tL}{2} \|\wb^{r+1} - \wb^r\|^2\\
&=-\teta \nabla F(\wb^r)^\top \left( \frac{1}{N K} \sum_i\sum_k (G^r_{i, k, 1} + G^r_{i, k, 2}) \right) +  \frac{\tL}{2} \|\wb^{r+1} - \wb^r \|^2 \\  
&=-\teta (\nabla F(\wb^r) - \nabla F(\wb^{r-1}) + \nabla F(\wb^{r-1}) )^\top  \left( \frac{1}{N K} \sum_i\sum_k (G^r_{i, k, 1} + G^r_{i, k, 2}) \right) 
+  \frac{\tL}{2} \|\wb^{r+1} - \wb^r \|^2 \\  
& \leq \frac{1}{2\tL}\|\nabla F(\wb^r) - \nabla F(\wb^{r-1})\|^2
+ 2\teta^2 \tL \| \frac{1}{N K} \sum_i\sum_k (G^r_{i, k, 1} + G^r_{i, k, 2}) \|^2 \\
&~~~ -\teta \nabla F(\wb^{r-1})^\top \left( \frac{1}{N K} \sum_i\sum_k (G^r_{i, k, 1} + G^r_{i, k, 2}) \right) 
+  \frac{\tL}{2} \|\wb^{r+1} - \wb^r \|^2\\
&\leq \frac{\tL}{2} \|\wb^r - \wb^{r-1}\|^2 +  2\teta^2 \tL \| \frac{1}{N K} \sum_i\sum_k (G^r_{i, k, 1} + G^r_{i, k, 2}) \|^2  
-\teta \nabla F(\wb^{r-1})^\top \left( \frac{1}{N K} \sum_i\sum_k (G^r_{i, k, 1} + G^r_{i, k, 2}) \right) \\
&~~~ + \frac{\tL}{2} \|\wb^{r+1} - \wb^r \|^2, 
\end{split} 
\label{eq:smooth_F}
\end{equation}  
\end{small}
where 
\begin{small}
\begin{equation}
\label{eq:smooth_F_2}
\begin{split}
&-\E\left[\teta \nabla F(\wb^{r-1})^\top \left( \frac{1}{N K} \sum_i\sum_k (G^r_{i, k, 1} + G^r_{i, k, 2}) \right) \right] \\
&= -\E\bigg[\teta \nabla F(\wb^{r-1})^\top \bigg( \frac{1}{N K} \sum_i\sum_k (G_1(\w^r_{i,k},\z^r_{i,k,1},\w^{r-1}_{j,t}, \z^{r-1}_{j,t,2}) + G_2(\w^{r-1}_{j',t'}, \z^{r-1}_{j',t',1}, \w^r_{i,k}, \z^r_{i,k,2}) \\
&~~~~~~~~~~~~~~~~~~~~~~~~~~~~~~~~~~~~~~~~~~~~~
-  G_1(\wb^{r-1},\z^r_{i,k,1},\wb^{r-1}, \z^{r-1}_{j,t,2}) - G_2(\wb^{r-1}, \z^{r-1}_{j',t',1}, \wb^{r-1}, \z^r_{i,k,2}) 
\\
&~~~~~~~~~~~~~~~~~~~~~~~~~~~~~~~~~~~~~~~~~~~~~
+ G_1(\wb^{r-1},\z^r_{i,k,1},\wb^{r-1}, \z^{r-1}_{j,t,2}) + G_2(\wb^{r-1}, \z^{r-1}_{j',t',1}, \wb^{r-1}, \z^r_{i,k,2})) \bigg) \bigg] \\
&\leq  4\teta \tL^2 \frac{1}{NK}\sum_{i=1}^N \sum_{k=1}^K \E(\|\w^r_{i,k} - \wb^{r-1}\|^2 +  \|\w^{r-1}_{j,t} - \wb^{r-1}\|^2  
+  \|\w^{r-1}_{j',t'} - \wb^{r-1}\|^2 
+  \|\w^{r}_{i,k} - \wb^{r-1}\|^2 )\\  
&~~ + \frac{\teta}{4}\E\|\nabla F(\wb^{r-1})\|^2 
-\E\bigg[\teta \nabla F(\wb^{r-1})^\top \bigg( \frac{1}{N K} \sum_i\sum_k \nabla F_i(\wb^{r-1} )\bigg)\bigg] \\  
&\leq 16\teta \tL^2 \E\|\wb^r - \wb^{r-1}\|^2 + 8\teta\tL^2 \frac{1}{NK}\sum_i\sum_k\E\|\wb^{r} - \w^r_{i,k}\|^2
+8\teta\tL^2 \frac{1}{NK}\sum_i\sum_k\E\|\wb^{r-1} - \w^{r-1}_{i,k}\|^2 
-\frac{\teta}{2} \E\|\nabla F(\wb^{r-1})\|^2, 
\end{split}
\end{equation} 
\end{small} 
where first inequality uses Young's inequality, Lipschitz of $G_1, G_2$, and the fact that data samples $\z^r_{i,k,1}, \z^{r-1}_{j,t}, \z^{r-1}_{j',t',1}, \z^r_{i,k,2}$ are independent samples after $\wb^{r-1}$, therefore 
\begin{equation}
    \begin{split}
        \E[(G_1(\wb^{r-1},\z^r_{i,k,1},\wb^{r-1}, \z^{r-1}_{j,t,2}) + G_2(\wb^{r-1}, \z^{r-1}_{j',t',1}, \wb^{r-1}, \z^r_{i,k,2}) -  \nabla F_i(\wb^{r-1})] = \mathbf{0}.  
    \end{split} 
\end{equation}

To bound the updates of $\wb^r$ after one round, we have
\begin{equation}  
\begin{split}     
&\E\|\wb^{r+1} - \wb^r\|^2 = \teta^2 \E\| \frac{1}{NK}\sum_i \sum_k (G^r_{i,k,1} + G^r_{i,k,2}) \|^2\\
&= \teta^2 \E\|\frac{1}{NK} \sum_i \sum_k (G_1(\w^r_{i,k}, \z^r_{i,k,1}, \w^{r-1}_{j,t}, \z^{r-1}_{j,t,2}) + G_2(\w^{r-1}_{j',t'}, \z^{r-1}_{j',t',1}, \w^{r}_{i,k}, \z^r_{i,k,2}) )\|^2 \\
&\leq 3\teta^2 \E\bigg\|\frac{1}{NK} \sum_i \sum_k [G_1(\w^r_{i,k}, \z^r_{i,k,1}, \w^{r-1}_{j,t}, \z^{r-1}_{j,t,2}) + G_2(\w^{r-1}_{j',t'}, \z^{r-1}_{j',t',1}, \w^{r}_{i,k}, \z^r_{i,k,2}) ] \\ 
&~~~~~~~~~ - \frac{1}{NK} \sum_i \sum_k [G_1(\wb^{r-1}, \z^r_{i,k,1}, \wb^{r-1}, \z^{r-1}_{j,t,2}) + G_2(\wb^{r-1}, \z^{r-1}_{j',t',1}, \wb^{r-1}, \z^r_{i,k,2}) ]  \bigg \|^2 \\
&~~ + 3\teta^2\E \bigg\| \frac{1}{NK} \sum_i \sum_k [G_1(\wb^{r-1}, \z^r_{i,k,1}, \wb^{r-1}, \z^{r-1}_{j,t,2}) + G_2(\wb^{r-1}, \z^{r-1}_{j',t',1}, \wb^{r-1}, \z^r_{i,k,2})  
\!-\! \nabla F_i(\wb^{r-1})]  \bigg\|^2 \\
&~~ + 3\teta^2\E \left\| \nabla F(\wb^{r-1}) \right\|^2 \\
\end{split}   
\end{equation} 
Using the Lipschtz property of $G_1, G_2$, we continue this inequality as  
\begin{small} 
\begin{equation*} 
\begin{split} 
&\E\|\wb^{r+1} - \wb^r\|^2 \\
&\leq 6\teta^2 \frac{\tL^2}{NK} \sum_i \sum_k \E\|\w^r_{i,k} - \wb^r\|^2 + 6\teta^2 \frac{\tL^2}{NK} \sum_i \sum_k \E\|\w^{r-1}_{i,k} - \wb^{r-1}\|^2 + 6\teta^2 \tL^2 \E\|\wb^r - \wb^{r-1}\|^2 \\
&~~ + 3\teta^2 \frac{1}{NK} \E \bigg\|  [G_1(\wb^{r-1}, \z^r_{i,k,1}, \wb^{r-1}, \z^{r-1}_{j,t,2}) + G_2(\wb^{r-1}, \z^{r-1}_{j',t',1}, \wb^{r-1}, \z^r_{i,k,2})  
- \nabla F_i(\wb^{r-1})]  \bigg\|^2 
 + 3\teta^2 \E\|F(\wb^{r-1})\|^2 \\
&\leq 6\teta^2 \frac{\tL^2}{NK} \sum_i \sum_k \E\|\w^r_{i,k} - \wb^r\|^2 + 6\teta^2 \frac{\tL^2}{NK} \sum_i \sum_k \E\|\w^{r-1}_{i,k} - \wb^{r-1}\|^2 + 6\teta^2 \tL^2 \E\|\wb^r - \wb^{r-1}\|^2 \\ 
&~~ + 3\teta^2 \frac{\sigma^2}{NK} + 3\teta^2 \E\|F(\wb^{r-1})\|^2.  
\end{split}   
\end{equation*} 
\end{small}

Thus,
\begin{equation}
\begin{split}
&\frac{1}{R} \sum_{r=1}^R \E\|\wb^{r+1} - \wb^r\|^2 \\
&\leq \frac{1}{R} \sum_{r=1}^R \bigg[ 10\teta^2 \tL^2 \frac{1}{NK} \sum_i \sum_k \E\|\w^r_{i,k} - \wb^r\|^2 + 6 \teta^2 \frac{\sigma^2}{NK} + 6\teta^2 \E\|F(\wb^{r-1})\|^2 \bigg].  
\end{split} 
\end{equation} 

Using Assumption \ref{ass:linear}, we know that $\|G_1\|^2, \|G_2\|^2$ are both less than $C_\ell^2 C_h^2$.
Then, to bound the updates in one round of one machine as
\begin{equation}
\begin{split}
&\E\|\wb^r - \w^r_{i,k}\|^2 \leq 2 \teta^2 C_\ell^2 C_h^2.
% = \|\w^r_{i,k-1} - \eta (G_1(\w^r_{i,k-1}, \z^r_{i,k-1,1}, \w^{r-1}_{j,t}, \z^{r-1}_{j,t,2}) + G_2(\w^{r-1}_{j',t'}, \z^{r-1}_{j',t',1}, \w^{r}_{i,k-1}, \z^r_{i,k-1,2})) - \wb^r\|^2 \\  
% & \leq  \|\w^r_{i, k-1} - \wb^r - \eta (G_1(\wb^{r-1}, \z^r_{i,k-1,1}, \wb^{r-1}, \z^{r-1}_{j,t,2}) + G_2(\wb^{r-1}, \z^{r-1}_{j',t',1}, \wb^{r-1},  \z^r_{i,k-1,2}))   \\
% &~~~ + \eta ([G_1(\wb^{r-1}, \z^r_{i,k-1,1}, \wb^{r-1}, \z^{r-1}_{j,t,2}) + G_2(\wb^{r-1}, \z^{r-1}_{j',t',1}, \wb^{r-1},  \z^r_{i,k-1,2})] \\  
%& ~~~~~~~~~~~~  - [G_1(\wb^{r}, \z^r_{i,k-1,1}, \wb^{r-1}, \z^{r-1}_{j,t,2}) + G_2(\wb^{r-1}, \z^{r-1}_{j',t',1}, \wb^{r},  \z^r_{i,k-1,2})]) \\
%& ~~~ + \eta ([G_1(\wb^{r}, \z^r_{i,k-1,1}, \wb^{r-1}, \z^{r-1}_{j,t,2}) + G_2(\wb^{r-1}, \z^{r-1}_{j',t',1}, \wb^{r},  \z^r_{i,k-1,2})] \\
%&~~~~~~~~~~~~ - [G_1(\w^{r}_{i,k-1}, \z^r_{i,k-1,1}, \wb^{r-1}, \z^{r-1}_{j,t,2}) + G_2(\wb^{r-1}, \z^{r-1}_{j',t',1}, \w^{r}_{i,k-1},  \z^r_{i,k-1,2})]) \\
%&~~~ + \eta ([G_1(\w^{r}_{i,k-1}, \z^r_{i,k-1,1}, \wb^{r-1}, \z^{r-1}_{j,t,2}) + G_2(\wb^{r-1}, \z^{r-1}_{j',t',1}, \w^{r}_{i,k-1},  \z^r_{i,k-1,2})] \\
%&~~~~~~~~~~~~ - [G_1(\w^{r}_{i,k-1}, \z^r_{i,k-1,1}, \w^{r-1}_{j,t}, \z^{r-1}_{j,t,2}) + G_2(\w^{r-1}_{j',t'}, \z^{r-1}_{j',t',1}, \w^{r}_{i,k-1},  \z^r_{i,k-1,2})]) \|^2 \\ 
\end{split}
\end{equation}

Recalling (\ref{eq:smooth_F}) and (\ref{eq:smooth_F_2}), we obtain
\begin{equation}
\begin{split}
\frac{1}{R} \sum\limits_{r=1}^{R} \E\|F(\wb^{r-1})\|^2 \leq O\left(\frac{2(F(\wb^1) - F_*)}{\teta R} +  \teta^2 \tL^2 C_\ell^2 C_h^2 + \teta \frac{\sigma^2}{NK}\right). 
\end{split}
\end{equation} 
By setting parameters as in the theorem, we conclude the proof. 
Besides, if we set $\eta = O(N\epsilon^2)$, $K = O(1/ N\epsilon)$, thus $\teta = O(\epsilon)$, to ensure $\frac{1}{R} \sum\limits_{r=1}^{R} \E\|F(\wb^{r-1})\|^2\leq \epsilon^2$, it takes 
communication rounds of $R=O(\frac{1}{\epsilon^3})$, and sample complexity on each machine $O(\frac{1}{N\epsilon^4})$. 
\end{proof}

%\newpage 
\section{FeDXL2 for Solving DXO with Non-Linear $f$}  
\label{app:sec:FeDXL2}
In this section, we define the following notations:  
\begin{equation} 
\begin{split}    
&G_{i,1}(\w_1, \z_1, \u, \w_2, \z_2) = \nabla f(\u) \nabla_1 \ell(h(\w_1, \z_1), h(\w_2, \z_2)) \nabla h(\w_1, \z_1), \\  
&G_{i,2}(\w_1, \z_1, \u, \w_2, \z_2) = \nabla f(\u) \nabla_2 \ell(h(\w_1, \z_1), h(\w_2, \z_2)) \nabla h(\w_2, \z_2). 
\end{split} 
\end{equation} 

Based on Assumption \ref{ass:non_linear}, it follows that $G_{i, 1}, G_{i, 2}$ are Lipschitz with some constant modulus $L_1$ and $\|G_{i, 1}\|^2, \|G_{i, 2}\|^2$ are bounded by $C_f^2 C_\ell^2 C_h^2$, $F$ is $L_F$-smooth, where $L_1, L_F$ are some proper constants depend on Assumption \ref{ass:non_linear}. We denote $\tL=\max\{L_1, L_F\}$ to simplify notations. 

For $\z_1\in \S_1^i, \z_2\in \S_2^j$, define $g(\w_1, \z_1, \w_2, \z_2) = \ell(h(\w_1; \z_1), h(\w_2, \z_2))$ and for $\z_1\in \S_1^i$, we define
\begin{equation} 
\begin{split}
    g(\w_1, \z_1, \w_2, \S_2) = \frac{1}{N}\sum\limits_{j=1}^{N}\E_{\z'\in \S_2^j}\ell(h(\w_1; \z_1), h(\w_2, \z')) 
\end{split}
\end{equation}
It follows that $g$ is also $\tL$-Lipschitz in $\w_1$ and $\w_2$. 

\subsection{Analysis of the moving average estimator $\u$}  
\begin{lemma}
\label{lem:nonlinear_u}
Under Assumption \ref{ass:non_linear}, the moving average estimator $\u$ satisfies 
\begin{equation*} 
\begin{split}
&\frac{1}{N} \sum\limits_{i=1}^{N} \frac{1}{|\S_1^i|} \sum\limits_{\z \in |\S_1^i|} \E\|\u^r_{i,k}(\z) - g(\wb^r_{k}, \z, \wb^r_k, \S_2)\|^2 \\
&\leq (1-\frac{\gamma}{16|\S_1^i|})\frac{1}{N} \sum\limits_{i=1}^{N} \frac{1}{|\S_1^i|}  \sum\limits_{\z \in |\S_1^i|} [\E\|\u^r_{i,k-1}(\z) - g(\wb^r_{k-1}, \z, \wb^r_{k-1}, \S_2)\|^2 \\
& + \frac{20|\S_1^i|}{\gamma} \tL^2\|\wb^r_{k-1} - \wb^r_k\|^2 ]  + 8\frac{\gamma^2}{|\S_1^i|} (\sigma^2+C_0^2)   + \frac{16\gamma \beta^2 K^2 C_0^2}{|\S_1^i|} \\
& + 8 \tL^2 \|\wb^r - \wb^{r-1}\|^2 + 8 \tL^2 \|\wb^r - \wb^{r}_k\|^2 \\
&+ 8(\gamma^2 + \frac{\gamma}{|\S_1^i|} )\tL^2 \frac{1}{N}\sum_{i} \|\wb^{r} - \w^r_{i,k}\|^2 
%+ \frac{\gamma \tL^2}{|\S_1^i|} \|\wb^r - \wb^r_k\|^2  
+ 2(\gamma^2 + \frac{\gamma}{|\S_1^i|} ) \tL^2 \frac{1}{NK}\sum\limits_{i=1}^N \sum\limits_{k=1}^{K} \E\|\wb^{r-1} - \wb^{r-1}_{i,k}\|^2. 
\end{split} 
\end{equation*} 
\end{lemma}

\begin{proof}
By update rules of $\u$, we have 
\begin{equation}
\begin{split} 
\u_{i, k}^{r} (\z) = \left\{ \begin{array}{cc}
    \u_{i, k-1}^r (\z) - \gamma (\u^r_{i,k-1}(\z) - \ell(h(\w^r_{i,k}, \z^r_{i,k,1}), h(\w^{r-1}_{j,t}, \hat{\z}^{r-1}_{j,t,2})))  & \z=\z^r_{i,k,1}  \\
      \u_{i, k-1}^r (\z)  & \z \neq \z^r_{i,k,1}.
\end{array} \right. 
\end{split}  
\end{equation}  

Or equivalently, 
\begin{equation}
\begin{split}
\u_{i, k}^{r} (\z) = \left\{ \begin{array}{cc}
    \u_{i, k-1}^r (\z) - \gamma (\u^r_{i,k-1}(\z) - g(\w^r_{i,k}, \z^r_{i,k,1}, \w^{r-1}_{j,t}, \hat{\z}^{r-1}_{j,t,2}))  & \z=\z^r_{i,k,1}  \\ 
      \u_{i, k-1}^r (\z)  & \z\neq \z^r_{i,k,1} 
\end{array} \right. 
\end{split}  
\end{equation}

Define $\bar{\u}^r_k = (\u^r_{1, k}, \u^r_{2, k}, ..., \u^r_{N, k})$, $\wb^r_k = \frac{1}{N} \sum\limits_{i=1}^{N} \w^r_{i,k}$.
%and 
%\begin{equation} 
%\begin{split} 
%&\phi^r_{k} (\bar{\u}^r_{k}) 
%= \frac{1}{2N} \sum\limits_{i=1}^N \frac{1}{|\S_i|} \sum_{\z\in \S_1^i} \|\u^r_{i,k} (\z) - g(\wb^r_{k}, \z, \wb^r_k, \S_2)\|^2.
% &= \frac{1}{2N} \sum\limits_{i=1}^N \frac{1}{|\S_1^i|} \sum_{\z\in \S_1^i}  \left\|\u^r_{i,k}(\z) - \frac{1}{N}\sum\limits_{i=1}^N \frac{1}{Q_i} \sum_{q\in Q_i} \ell(f(\wb^r_{k}; p) - f(\wb^r_{k}; q)) \right\|^2.
%\end{split}     
%\end{equation}  
Then it follows that
\begin{small} 
\begin{equation} 
\begin{split}
% &\frac{1}{2}\phi^r_{k}(\bar{\u}^r_{k}) = 
&\frac{1}{2N} \sum\limits_{i=1}^{N} \frac{1}{|\S_1^i|} \sum\limits_{\z \in |\S_1^i|} \E\|\u^r_{i,k}(\z) - g(\wb^r_{k}, \z, \wb^r_k, \S_2)\|^2 \\ 
&=\frac{1}{N} \sum_i \frac{1}{|\S_1^i|} \sum_{\z\in |\S_1^i|} \E\bigg[ \frac{1}{2} \|\u^r_{i,k-1}(\z) - g(\wb^r_{k}, \z, \wb^r_k, \S_2)\|^2 \\
&~~~~~~ +  \langle \u^r_{i,k-1}(\z) - g(\wb^r_{k}, \z, \wb^r_k, \S_2), \u^r_{i,k}(\z) - \u^r_{i,k-1}(\z) \rangle 
%&~~~~~~~~~~~~~~~~~~~~~~~~~~~~~~~~~~~~~~~~~~~~~~~~~~~~~
+ \frac{1}{2}\|\u^r_{i,k}(\z) - \u^r_{i,k-1}(\z)\|^2 \bigg],
\end{split} 
\end{equation} 
which is 
\begin{equation} 
\begin{split}
&\frac{1}{2N} \sum\limits_{i=1}^{N} \frac{1}{|\S_1^i|} \sum\limits_{\z \in |\S_1^i|} \E\|\u^r_{i,k}(\z) - g(\wb^r_{k}, \z, \wb^r_k, \S_2)\|^2 \\ 
&=\frac{1}{2N} \sum_i \frac{1}{|\S_i|} \sum_{\z\in \S_1^i} \E\|\u^r_{i,k-1}(\z) - g(\wb^r_{k}, \z, \wb^r_k, \S_2)\|^2\\
%&~~~~~~~~~~~~~~~~~~~~~~~~~~~~~~~~~~~~
&~~~ + \frac{1}{N} \sum_i
\frac{1}{|\S_1^i|} \E\langle \u^r_{i,k-1}(\z^r_{i,k,1}) - g(\wb^r_{k}, \z^r_{i,k,1}, \wb^r_k, \S_2),  \u^r_{i,k}(\z^r_{i,k,1}) - \u^r_{i,k-1}(\z^r_{i,k,1}) \rangle \\ 
% &~~~~~~~~~~~~ ~~~~~~~~~~~~ ~~~~~~~~~~~  
&~~~ + \frac{1}{N} \sum_i \frac{1}{2|\S_1^i|}\E\|\u^r_{i,k}(\z^r_{i,k,1}) - \u^r_{i,k-1}(\z^r_{i,k,1})\|^2 \\  
& = \frac{1}{2N} \sum_i \frac{1}{|\S_i|} \sum_{\z\in \S_1^i} \E\|\u^r_{i,k-1}(\z) - g(\wb^r_{k}, \z, \wb^r_k, \S_2)\|^2 \\
&~~~ + \frac{1}{N} \sum_i\frac{1}{|\S_1^i|} \E\langle \u^r_{i,k-1}(\z^r_{i,k,1}) -  g(\w^r_{i,k}, \z^r_{i,k,1}, \w^{r-1}_{j,t}, \hat{\z}^{r-1}_{j,t,2}), \u^r_{i,k}(\z^r_{i,k,1}) - \u^r_{i,k-1}(\z^r_{i,k,1}) \rangle  \\ 
&~~~ + \frac{1}{N} \sum_i \frac{1}{|\S_1^i|} \E\langle  g(\w^r_{i,k}, \z^r_{i,k,1}, \w^{r-1}_{j,t}, \hat{\z}^{r-1}_{j,t,2}) - g(\wb^r_{k}, \z^r_{i,k,1}, \wb^r_k, \S_2), \u^r_{i, k}(\z^r_{i,k,1}) - \u^r_{i,k-1}(\z^r_{i,k,1}) \rangle
\\
&~~~ + \frac{1}{N} \sum_i \frac{1}{2 |\S_i|} \E\|\u^r_{i,k}(\z^r_{i,k,1}) - \u^r_{i, k-1}(\z^r_{i,k,1})\|^2,
\end{split} 
\end{equation} 
\end{small} 
where 
\begin{equation}
\begin{split}
& \langle \u^r_{i,k-1}(\z^r_{i,k,1}) - g(\w^r_{i,k}, \z^r_{i,k,1}, \w^{r-1}_{j,t}, \hat{\z}^{r-1}_{j,t,2}), \u^r_{i,k}(\z^r_{i,k,1}) - \u^r_{i,k-1}(\z^r_{i,k,1}) \rangle \\ 
&= \langle \u^r_{i,k-1}(\z^r_{i,k,1}) - g(\w^r_{i,k}, \z^r_{i,k,1}, \w^{r-1}_{j,t}, \hat{\z}^{r-1}_{j,t,2}), g(\wb^r_{k},\z^r_{i,k,1},\wb^r_{k}, \S_2) - \u^r_{i,k-1}(\z^r_{i,k,1}) \rangle \\
&~~~ 
+ \langle \u^r_{i,k-1}(\z^r_{i,k,1}) - g(\w^r_{i,k}, \z^r_{i,k,1}, \w^{r-1}_{j,t}, \hat{\z}^{r-1}_{j,t,2}), \u^r_{i,k}(\z^r_{i,k,1}) - g(\wb^r_{k},\z^r_{i,k,1},\wb^r_{k}, \S_2)  \rangle \\ 
& = \langle \u^r_{i,k-1}(\z^r_{i,k,1}) -  g(\w^r_{i,k}, \z^r_{i,k,1}, \w^{r-1}_{j,t}, \hat{\z}^{r-1}_{j,t,2}),  g(\wb^r_{k},\z^r_{i,k,1},\wb^r_{k}, \S_2)  - \u^r_{i,k-1}(\z^r_{i,k,1}) \rangle \\
&~~~ 
+\frac{1}{\gamma}\langle \u^r_{i,k-1}(\z^r_{i,k,1}) - \u^r_{i,k}(\z^r_{i,k,1}), \u^r_{i,k}(\z^r_{i,k,1}) - g(\wb^r_{k},\z^r_{i,k,1},\wb^r_{k}, \S_2)  \rangle \\
& = \langle \u^r_{i,k-1}(\z^r_{i,k,1}) - g(\w^r_{i,k}, \z^r_{i,k,1}, \w^{r-1}_{j,t}, \hat{\z}^{r-1}_{j,t,2}), g(\wb^r_{k},\z^r_{i,k,1},\wb^r_{k},\S_2) - \u^r_{i,k-1}(\z^r_{i,k,1}) \rangle \\  
&~~~ 
+ \frac{1}{2\gamma} (\|\u^r_{i,k-1}(\z^r_{i,k,1}) - g(\wb^r_{k},\z^r_{i,k,1},\wb^r_{k}, \S_2) \|^2 - \|\u^r_{i,k}(\z^r_{i,k,1}) - \u^r_{i, k-1}(\z^r_{i,k,1})\|^2 \\
&~~~ ~~~ ~~~~~~ - \|\u^r_{i,k}(\z^r_{i,k,1}) -  g(\wb^r_{k},\z^r_{i,k,1},\wb^r_{k}, \S_2) \|^2)
\end{split}      
\end{equation}   

If $\gamma \leq \frac{1}{5}$, we have  
\begin{equation}
\begin{split}
& -\frac{1}{2} \left( \frac{1}{\gamma} - 1 - \frac{\gamma+1}{4\gamma}  \right) \E\|\u^r_{i,k}(\z^r_{i,k,1}) -   \u^r_{i,k-1}(\z^r_{i,k,1})\|^2 \\
&~~~ + \E \langle  g(\w^r_{i,k}, \z^r_{i,k,1}, \w^{r-1}_{j,t}, \hat{\z}^{r-1}_{j,t,2}) - g(\wb^r_{k}, \z^r_{i,k,1},\wb^r_k, \S_2), \u^r_{i,k}(\z^r_{i,k,1}) - \u^r_{i,k-1}(\z^r_{i,k,1}) \rangle \\
&\leq  -\frac{1}{4\gamma}   \E\|\u^r_{i,k}(\z^r_{i,k,1}) - \u^r_{i,k-1}(\z^r_{i,k,1})\|^2 
+ \gamma \E\|g(\w^r_{i,k}, \z^r_{i,k,1}, \w^{r-1}_{j,t}, \hat{\z}^{r-1}_{j,t,2}) - g(\wb^r_{k}, \z^r_{i,k,1},\wb^r_k, \S_2)\|^2 \\
&~~~ 
+ \frac{1}{4\gamma} \E \|\u^r_{i,k}(\z^r_{i,k,1}) - \u^r_{i,k-1}(\z^r_{i,k,1})\|^2 \\ 
& \leq  \gamma \E\|g(\w^r_{i,k}, \z^r_{i,k,1}, \w^{r-1}_{j,t}, \hat{\z}^{r-1}_{j,t,2}) - g(\wb^r_{k}, \z^r_{i,k,1},\wb^r_k, \S_2)\|^2 \\
&\leq 4\gamma  \E\|g(\wb^{r-1}, \z^r_{i,k,1}, \wb^{r-1}, \hat{\z}^{r-1}_{j,t,2}) - g(\wb^{r-1},\z^r_{i,k,1}, \wb^{r-1}, \S_2)\|^2 + 4\gamma\tL^2 \E\|\wb^{r} - \wb^{r-1}\|^2 \\
&~~~ + 4\gamma\tL^2 \E\|\w^r_{i,k} - \wb^{r}\|^2 + 4\gamma\tL^2 \E\|\w^{r-1}_{j,t} - \wb^{r-1}\|^2 \\
&\leq 4\gamma  \sigma^2 + 4\gamma\tL^2 \E\|\wb^{r} - \wb^{r-1}\|^2 + 4\gamma\tL^2 \E\|\w^r_{i,k} - \wb^{r}\|^2 + 4\gamma\tL^2 \E\|\w^{r-1}_{j,t} - \wb^{r-1}\|^2.
\end{split} 
\end{equation}

Then, we have
\begin{equation}
\begin{split}
&\frac{1}{2N} \sum\limits_{i=1}^{N} \frac{1}{|\S_1^i|} \sum\limits_{\z \in |\S_1^i|} \E\|\u^r_{i,k}(\z) - g(\wb^r_{k}, \z, \wb^r_k, \S_2)\|^2 \\
&\leq \frac{1}{2N} \sum\limits_{i=1}^{N} \frac{1}{|\S_1^i|} \sum\limits_{\z \in |\S_1^i|} \E\|\u^r_{i,k-1}(\z) - g(\wb^r_{k}, \z, \wb^r_k, \S_2)\|^2 \\
& + \frac{1}{N}\sum_i \frac{1}{|\S_1^i|} \Bigg[ \frac{1}{2\gamma}  \E\|\u^r_{i,k-1}(\z^r_{i,k,1}) - g(\wb^{r}_k,\z^r_{i,k,1}, \wb^{r}_k, \S_2)\|^2 \\
& - \frac{1}{2\gamma} \E\|\u^r_{i,k}(\z^r_{i,k,1})- g(\wb^{r}_{k},\z^r_{i,k,1}, \wb^{r}_k, \S_2)\|^2\! -\! \frac{\gamma+1}{8\gamma} \|\u^r_{i,k}(\z^r_{i,k,1})  - \u^r_{i,k-1}(\z^r_{i,k,1}) \|^2 \!+\! 4\gamma \sigma^2 \\
 % + \gamma  \|g(\wb^{r-1}, \z^r_{i,k,1}, \wb^{r-1}, \hat{\z}^{r-1}_{j,t,2}) - g(\wb^{r-1}, \z^r_{i,k,1}, \wb^{r-1}, \S_2)\|^2 \\ 
 & + 4\gamma \tL^2 \E\|\wb^r-\wb^{r-1}\|^2
 + 4\gamma \tL^2 \E\|\w^r_{i,k} - \wb^r\|^2
 +4 \gamma\tL^2\E\|\w^{r-1}_{j,t} - \wb^{r-1}\|^2 \\
&+ \E \langle \u^r_{i,k-1}(\z^r_{i,k,1}) - g(\w^r_{i,k}, \z^r_{i,k,1}, \w^{r-1}_{j,t}, \hat{\z}^{r-1}_{j,t,2}), g(\wb^r_{k}, \z^r_{i,k,1}, \wb^r_k, \S_2) - \u^r_{i, k-1}(\z^r_{i,k,1}) \rangle  \Bigg].
\end{split}
\end{equation}  

Note that $\sum_{\z\neq \z^r_{i,k,1}}\|\u^r_{i,k-1}(\z) - g(\wb^r_{k}, \z, \wb^r_{k}, \S_2 )\|^2 = \sum_{\z\neq \z^r_{i,k,1}}\|\u^r_{i, k}(\z) - g(\w^r_{k}, \z, \wb^r_{k}, \S_2)\|^2 $, which implies
\begin{equation}  
\begin{split}     
&\frac{1}{2\gamma} \left( \|\u^r_{i,k-1}(\z^r_{i,k,1}) - g(\wb^r_{k}, \z, \wb^r_{k}, \S_2 )\|^2 - \|\u^r_{i, k}(\z^r_{i,k,1}) - g(\wb^r_{k}, \z, \wb^r_{k}, \S_2 )\|^2 \right) \\ 
&= \frac{1}{2\gamma}\sum_{\z\in \S_1^i} \left( \|\u^r_{i,k-1}(\z) - g(\wb^r_{k}, \z, \wb^r_{k}, \S_2 )\|^2 - \|\u^r_{i,k}(\z) - g(\wb^r_{k}, \z, \wb^r_{k}, \S_2 )\|^2 \right).  
\end{split}    
\end{equation} 

Since $\ell(\cdot) \leq C_0$, we have that $\|g(\cdot)\|^2 \leq C_0^2$, $\|\u^r_{i,k}(\z)\|^2 \leq C_0^2$ and $$\|\u^r_{i,k}(\z) - \u^{r}_{i,0}(\z)\|^2 \leq \beta^2 K^2 C_0^2$$.
Besides, we have 
%\begin{small} 
\begin{equation}
\begin{split}
& \E\langle \u^r_{i,k-1}(\z^r_{i,k,1}) -  g(\w^r_{i,k}, \z^r_{i,k,1}, \w^{r-1}_{j,t}, \hat{\z}^{r-1}_{j,t,2}), g(\wb^r_{k}, \z^r_{i,k,1},\wb^r_{k},\S_2) -  \u^r_{i,k-1}(\z^r_{i,k,1}) \rangle \\ 
& = \E\langle \u^r_{i,k-1}(\z^r_{i,k,1}) -  g(\wb^{r-1}, \z^r_{i,k,1}, \wb^{r-1}, \hat{\z}^{r-1}_{j,t,2}), g(\wb^r_{k}, \z^r_{i,k,1},\wb^r_{k},\S_2) -  \u^r_{i,k-1}(\z^r_{i,k,1}) \rangle \\
&  + \E\langle g(\wb^{r-1}, \z^r_{i,k,1}, \wb^{r-1}, \hat{\z}^{r-1}_{j,t,2})  -  g(\w^r_{i,k}, \z^r_{i,k,1}, \w^{r-1}_{j,t}, \hat{\z}^{r-1}_{j,t,2}), g(\wb^r_{k}, \z^r_{i,k,1},\wb^r_{k},\S_2) -  \u^r_{i,k-1}(\z^r_{i,k,1}) \rangle \\
&\leq \E\langle \u^r_{i,k-1}(\z^r_{i,k,1}) -  g(\wb^{r-1}, \z^r_{i,k,1}, \wb^{r-1}, \hat{\z}^{r-1}_{j,t,2}) , 
g(\wb^r_{k}, \z^r_{i,k,1},\wb^r_{k},\S_2) - g(\wb^{r-1}, \z^r_{i,k,1},\wb^{r-1},\S_2) 
\rangle \\
&+ \E\langle \u^r_{i,k-1}(\z^r_{i,k,1}) -  g(\wb^{r-1}, \z^r_{i,k,1}, \wb^{r-1}, \hat{\z}^{r-1}_{j,t,2}), g(\wb^{r-1}, \z^r_{i,k,1},\wb^{r-1},\S_2) -  \u^r_{i,k-1}(\z^r_{i,k,1}) 
\rangle \\
& + 2\tL^2\E\|\wb^{r} - \wb^{r-1}\|^2 + 2\tL^2 \E\|\wb^{r} - \w^r_{i,k}\|^2 
+ \tL^2 \E\|\wb^{r-1} - \w^{r-1}_{j,t}\|^2 \\
& + \frac{1}{4}\E\|g(\wb^r, \z^r_{i,k,1}, \wb^r_k, \S_2) - \u^r_{i,k-1}(\z^r_{i,k,1}) \|^2 \\
&\leq 2\gamma  C_0^2 + \frac{1}{\gamma} \|\wb^r_k - \wb^{r-1}\|^2 \\
& + \E\langle \u^r_{i,k-1}(\z^r_{i,k,1}) -  g(\wb^{r-1}, \z^r_{i,k,1}, \wb^{r-1}, \hat{\z}^{r-1}_{j,t,2}), g(\wb^{r-1}, \z^r_{i,k,1},\wb^{r-1},\S_2) -  \u^r_{i,k-1}(\z^r_{i,k,1}) 
\rangle \\
&  + 2\tL^2\E\|\wb^{r} - \wb^{r-1}\|^2 + 2\tL^2 \E\|\wb^{r} - \w^r_{i,k}\|^2 
+ \tL^2 \E\|\wb^{r-1} - \w^{r-1}_{j,t}\|^2 \\
& + \frac{1}{4}\E\|g(\wb^r, \z^r_{i,k,1}, \wb^r_k, \S_2) - \u^r_{i,k-1}(\z^r_{i,k,1}) \|^2,
\end{split}
\end{equation} 
%\end{small}
% \|g(\wb^{r-1}, \z^r_{i,k,1}, \wb^{r-1}, S_2) - \u^r_{i, k-1}(\z^r_{i,k,1})\|^2 \\
where
%\begin{small} 
\begin{equation}
\begin{split}
&\E\langle \u^r_{i,k-1}(\z^r_{i,k,1}) -  g(\wb^{r-1}, \z^r_{i,k,1}, \wb^{r-1}, \hat{\z}^{r-1}_{j,t,2}), g(\wb^{r-1}, \z^r_{i,k,1},\wb^{r-1},\S_2) -  \u^r_{i,k-1}(\z^r_{i,k,1}) 
\rangle \\
&=\E\langle \u^r_{i,k-1}(\z^r_{i,k,1}) - \u^{r-1}_{i,0}(\z^r_{i,k,1}) + \u^{r-1}_{i,0}(\z^r_{i,k,1})  -  g(\wb^{r-1}, \z^r_{i,k,1}, \wb^{r-1}, \hat{\z}^{r-1}_{j,t,2}), \\
&~~~~~~~~~ g(\wb^{r-1}, \z^r_{i,k,1},\wb^{r-1},\S_2) - \u^{r-1}_{i,0}(\z^r_{i,k,1}) + \u^{r-1}_{i,0}(\z^r_{i,k,1}) -  \u^r_{i,k-1}(\z^r_{i,k,1}) 
\rangle \\
&\leq \E\langle \u^r_{i,k-1}(\z^r_{i,k,1}) - \u^{r-1}_{i,0}(\z^r_{i,k,1}), g(\wb^{r-1}, \z^r_{i,k,1},\wb^{r-1},\S_2) - \u^{r-1}_{i,0}(\z^r_{i,k,1})\rangle \\
& + \E\langle \u^r_{i,k-1}(\z^r_{i,k,1}) - \u^{r-1}_{i,0}(\z^r_{i,k,1}),  \u^{r-1}_{i,0}(\z^r_{i,k,1}) -  \u^r_{i,k-1}(\z^r_{i,k,1})  \rangle \\
& + \E\langle \u^{r-1}_{i,0}(\z^r_{i,k,1})  -  g(\wb^{r-1}, \z^r_{i,k,1}, \wb^{r-1}, \hat{\z}^{r-1}_{j,t,2}), g(\wb^{r-1}, \z^r_{i,k,1},\wb^{r-1},\S_2) - \u^{r-1}_{i,0}(\z^r_{i,k,1})\rangle \\
& + \E\langle  \u^{r-1}_{i,0}(\z^r_{i,k,1})  -  g(\wb^{r-1}, \z^r_{i,k,1}, \wb^{r-1}, \hat{\z}^{r-1}_{j,t,2}), \u^{r-1}_{i,0}(\z^r_{i,k,1}) -  \u^r_{i,k-1}(\z^r_{i,k,1}) \rangle \\
&\leq 4 \E\|\u^r_{i,k-1}(\z^r_{i,k,1}) - \u^{r-1}_{i,0}(\z^r_{i,k,1})\|^2
+ \frac{1}{4} \E\|g(\wb^{r-1}, \z^r_{i,k,1},\wb^{r-1},\S_2) - \u^{r-1}_{i,0}(\z^r_{i,k,1})\|^2 \\
& - \E\|g(\wb^{r-1}, \z^r_{i,k,1},\wb^{r-1},\S_2) - \u^{r-1}_{i,0}(\z^r_{i,k,1})\|^2 \\
& + \frac{1}{4}\E\|g(\wb^{r-1}, \z^r_{i,k,1},\wb^{r-1},\S_2) - \u^{r-1}_{i,0}(\z^r_{i,k,1})\|^2 
+ 4 \E\|\u^r_{i,k-1}(\z^r_{i,k,1}) - \u^{r-1}_{i,0}(\z^r_{i,k,1})\|^2 \\
&\leq  4 \E\|\u^r_{i,k-1}(\z^r_{i,k,1}) - \u^{r-1}_{i,0}(\z^r_{i,k,1})\|^2 
% + \frac{1}{4} \E\|g(\wb^{r-1}, \z^r_{i,k,1},\wb^{r-1},\S_2) - \u^{r-1}_{i,0}(\z^r_{i,k,1})\|^2 \\
-\frac{1}{2}\E\|g(\wb^{r-1}, \z^r_{i,k,1},\wb^{r-1},\S_2) - \u^{r-1}_{i,0}(\z^r_{i,k,1})\|^2
+ 8\beta^2 K^2 C_0^2. 
\end{split}
\end{equation}
%\end{small} 
Noting 
\begin{equation}
\begin{split}
&-\E\|g(\wb^{r-1}, \z^r_{i,k,1},\wb^{r-1},\S_2) - \u^{r-1}_{i,0}(\z^r_{i,k,1})\|^2 \\
&=-\E\|g(\wb^{r-1}, \z^r_{i,k,1},\wb^{r-1},\S_2) - \u^{r}_{i,k-1}(\z^r_{i,k,1}) + \u^{r}_{i,k-1}(\z^r_{i,k,1}) - \u^{r-1}_{i,0}(\z^r_{i,k,1})\|^2 \\
&=-\E\|g(\wb^{r-1}, \z^r_{i,k,1},\wb^{r-1},\S_2) - \u^{r}_{i,k-1}(\z^r_{i,k,1})\|^2 - \E \| \u^{r}_{i,k-1}(\z^r_{i,k,1}) - \u^{r-1}_{i,0}(\z^r_{i,k,1})\|^2 \\
&~~~ +2\E\langle g(\wb^{r-1}, \z^r_{i,k,1},\wb^{r-1},\S_2) - \u^{r}_{i,k-1}(\z^r_{i,k,1}),  \u^{r}_{i,k-1}(\z^r_{i,k,1}) - \u^{r-1}_{i,0}(\z^r_{i,k,1}) \rangle \\
&\leq -\frac{1}{2} \E\|g(\wb^{r-1}, \z^r_{i,k,1},\wb^{r-1},\S_2) - \u^{r}_{i,k-1}(\z^r_{i,k,1})\|^2 + 8 \| \u^{r}_{i,k-1}(\z^r_{i,k,1}) - \u^{r-1}_{i,0}(\z^r_{i,k,1})\|^2 \\
&\leq -\frac{1}{2} \E\|g(\wb^{r-1}, \z^r_{i,k,1},\wb^{r-1},\S_2) - \u^{r}_{i,k-1}(\z^r_{i,k,1})\|^2 + 8\beta^2K^2 C_0^2 \\
&\leq -\frac{1}{4} \E\|g(\wb^{r}_k, \z^r_{i,k,1},\wb^{r}_k,\S_2) - \u^{r}_{i,k-1}(\z^r_{i,k,1})\|^2
+ \frac{1}{2} \tL^2\|\wb^{r-1} - \wb^r_k\|^2
+ 8\beta^2K^2 C_0^2
\end{split} 
\end{equation} 

Then by multiplying $\gamma$ to every term and rearranging terms using the setting of $\gamma \leq O(1)$, 
we can obtain    
\begin{equation} 
\begin{split}
& \frac{\gamma+1}{2} \frac{1}{N} \sum\limits_{i=1}^{N} \frac{1}{|\S_1^i|} \sum\limits_{\z \in |\S_1^i|} \E\|\u^r_{i,k}(\z) - g(\wb^r_{k}, \z, \wb^r_k, \S_2)\|^2 \\
&\leq \frac{\gamma(1-\frac{1}{8|\S_1^i|}) + 1}{2}  \frac{1}{N} \sum\limits_{i=1}^{N} \frac{1}{|\S_1^i|} \sum\limits_{\z \in |\S_1^i|} \E\|\u^r_{i,k-1}(\z) - g(\wb^r_{k}, \z, \wb^r_k, \S_2)\|^2 \\  
& + \frac{4\gamma^2}{|\S_1^i|} (\sigma^2+C_0^2) 
+ \frac{8\gamma \beta^2 K^2 C_0^2}{|\S_1^i|} + 4\tL^2 \E\|\wb^r - \wb^{r-1}\|^2 + 4\tL^2 \E\|\wb^r - \wb^r_k\|^2  \\
& 
+ 4(\gamma^2 + \frac{\gamma}{|\S_1^i|} )\tL^2 \frac{1}{N}\sum_{i} \E\|\wb^{r} - \w^r_{i,k}\|^2 
%+ \frac{\gamma \tL^2}{|\S_1^i|} \|\wb^r - \wb^r_k\|^2  
+ (\gamma^2 + \frac{\gamma}{|\S_1^i|} ) \tL^2 \frac{1}{NK}\sum\limits_{i=1}^N \sum\limits_{k=1}^{K} \E\|\wb^{r-1} - \w^{r-1}_{i,k}\|^2. 
\end{split} 
\end{equation} 
Dividing $\frac{\gamma+1}{2}$ on both sides gives
\begin{equation}
\begin{split}
&  \frac{1}{N} \sum\limits_{i=1}^{N} \frac{1}{|\S_1^i|} \sum\limits_{\z \in |\S_1^i|} \E\|\u^r_{i,k}(\z) - g(\wb^r_{k}, \z, \wb^r_k, \S_2)\|^2 \\
&\leq \frac{\gamma(1-\frac{1}{8|\S_1^i|}) + 1}{\gamma+1}  \frac{1}{N} \sum\limits_{i=1}^{N} \frac{1}{|\S_1^i|} \sum\limits_{\z \in |\S_1^i|} \E\|\u^r_{i,k-1}(\z) - g(\wb^r_{k}, \z, \wb^r_k, \S_2)\|^2 \\
& + 8\frac{\gamma^2}{|\S_1^i|} (\sigma^2+C_0^2)  + \frac{16\gamma \beta^2 K^2 C_0^2}{|\S_1^i|} 
+ 8\tL^2 \|\wb^r - \wb^{r-1}\|^2 +  8\tL^2 \|\wb^r - \wb^{r}_k\|^2 \\
&+ 8(\gamma^2 + \frac{\gamma}{|\S_1^i|} )\tL^2 \frac{1}{N}\sum_{i} \|\wb^{r} - \w^r_{i,k}\|^2 
%+ \frac{\gamma \tL^2}{|\S_1^i|} \|\wb^r - \wb^r_k\|^2  
+ 2(\gamma^2 + \frac{\gamma}{|\S_1^i|} ) \tL^2 \frac{1}{NK}\sum\limits_{i=1}^N \sum\limits_{k=1}^{K} \E\|\wb^{r-1} - \wb^{r-1}_{i,k}\|^2. 
\end{split} 
\end{equation} 
Using Young's inequality, 
\begin{equation*} 
\begin{split}
&\frac{1}{N} \sum\limits_{i=1}^{N} \frac{1}{|\S_1^i|} \sum\limits_{\z \in |\S_1^i|} \E\|\u^r_{i,k}(\z) - g(\wb^r_{k}, \z, \wb^r_k, \S_2)\|^2 \\
&\leq (1-\frac{\gamma}{8|\S_1^i|}) \frac{1}{N} \sum\limits_{i=1}^{N} \frac{1}{|\S_1^i|}  \sum\limits_{\z \in |\S_1^i|} \bigg[(1+\frac{\gamma}{16|\S_1^i|})\E\|\u^r_{i,k-1}(\z) - g(\wb^r_{k-1}, \z, \wb^r_{k-1}, \S_2)\|^2 \\
&~~~  +(1+\frac{16|\S_1^i|}{\gamma})  \tL^2\|\wb^r_{k-1} - \wb^r_k\|^2 \bigg]  \\
& + 8\frac{\gamma^2}{|\S_1^i|} (\sigma^2+C_0^2)  + \frac{16\gamma \beta^2 K^2 C_0^2}{|\S_1^i|} 
+ 8 \tL^2 \|\wb^r - \wb^{r-1}\|^2 + 8 \tL^2 \|\wb^r - \wb^{r}_k\|^2 \\
&+ 8(\gamma^2 + \frac{\gamma}{|\S_1^i|} )\tL^2 \frac{1}{N}\sum_{i} \|\wb^{r} - \w^r_{i,k}\|^2 
%+ \frac{\gamma \tL^2}{|\S_1^i|} \|\wb^r - \wb^r_k\|^2  
+ 2(\gamma^2 + \frac{\gamma}{|\S_1^i|} ) \tL^2 \frac{1}{NK}\sum\limits_{i=1}^N \sum\limits_{k=1}^{K} \E\|\wb^{r-1} - \wb^{r-1}_{i,k}\|^2 
\\
&\leq (1-\frac{\gamma}{16|\S_1^i|})\frac{1}{N} \sum\limits_{i=1}^{N} \frac{1}{|\S_1^i|}  \sum\limits_{\z \in |\S_1^i|} [\E\|\u^r_{i,k-1}(\z) - g(\wb^r_{k-1}, \z, \wb^r_{k-1}, \S_2)\|^2 \\
& + \frac{20|\S_1^i|}{\gamma} \tL^2\|\wb^r_{k-1} - \wb^r_k\|^2 ]  + 8\frac{\gamma^2}{|\S_1^i|} (\sigma^2+C_0^2)   + \frac{16\gamma \beta^2 K^2 C_0^2}{|\S_1^i|} \\
& + 8 \tL^2 \|\wb^r - \wb^{r-1}\|^2 + 8 \tL^2 \|\wb^r - \wb^{r}_k\|^2 \\
&+ 8(\gamma^2 + \frac{\gamma}{|\S_1^i|} )\tL^2 \frac{1}{N}\sum_{i} \|\wb^{r} - \w^r_{i,k}\|^2 
%+ \frac{\gamma \tL^2}{|\S_1^i|} \|\wb^r - \wb^r_k\|^2  
+ 2(\gamma^2 + \frac{\gamma}{|\S_1^i|} ) \tL^2 \frac{1}{NK}\sum\limits_{i=1}^N \sum\limits_{k=1}^{K} \E\|\wb^{r-1} - \wb^{r-1}_{i,k}\|^2. 
\end{split} 
\end{equation*} 
\end{proof}

\subsection{Analysis of the estimator of gradient} 
With update $G^r_{i,k} = (1-\beta) G^r_{i,k-1} + \beta (G^r_{i,k,1} + G^r_{i,k,2})$,
we define $\bG^r_k := \frac{1}{N} \sum\limits_{i=1}^{N} G^r_{i,k}$, and $\Delta^r_{k} := \|\bG^r_{k} - \nabla F(\wb^r_k)\|^2$. 
Then it follows that $\bar{G}^r_{k} = (1-\beta) \bG^r_{k-1} + \beta \frac{1}{N} \sum_i (G^r_{i,k,1} + G^r_{i,k,2})$. 

\begin{lemma}
\label{lem:nonlinear_lem_G}
Under Assumption \ref{ass:non_linear}, with setting of $\eta = O(\beta)$, Algorithm \ref{alg:FeDXL2} ensures that 
\begin{equation*}
\begin{split} 
&\Delta^r_k \leq (1-\frac{3\beta}{4}) \|\bG^r_{k-1} - \nabla F(\wb^r_{k-1})\|^2
+ \frac{\eta}{16} \|\nabla F(\wb^{r-1})\|^2 
+ \frac{2\beta^2 \sigma^2}{N} + 36 \frac{\eta^2}{\beta}\tL^2 \|\wb^r_k - \wb^{r-1}\|^2 \\  
& + 12\beta 
\left(\frac{1}{N}\sum_i 4\tL^2 \E\|\w^r_{i,k} - \wb^r\|^2 + 4\tL^2 \E\|\wb^r - \wb^{r-1}\|^2
+ \frac{1}{N}\sum_i 4\tL^2 \E\|\w^{r-1}_{j',t'} - \wb^{r-1}\|^2  \right) \\
& + 12\beta \frac{1}{N}\sum_i  \bigg( \tL^2 \E\|\u^r_{i,k}(\z^r_{i,k,1}) -  g(\wb^{r}_k,\z^r_{i,k,1},\wb^{r}_k, \S_2)\|^2  
+ \tL^2 \E\|\u^{r-1}_{j',t'}(\hat{\z}^{r-1}_{j',t',1})-g(\wb^{r-1}_{t'}, \hat{\z}^{r-1}_{j',t',1}, \wb^{r-1}_{t'}, \S_2)\|^2\bigg). 
\end{split}     
\end{equation*} 
\end{lemma} 
\begin{proof}
\begin{small}
\begin{equation}
\begin{split}
& \Delta^r_{k} = \|\bG^r_{k} - \nabla F(\wb^r_k)\|^2 \\ 
& = \|(1-\beta) \bG^r_{k-1} + \beta  \frac{1}{N} \sum_{i} (G^r_{i,k,1} + G^r_{i,k,2}) - \nabla F(\wb^r_{k})  \|^2 \\  
&= \bigg\|(1-\beta) (\bG^r_{k-1} - \nabla F(\wb^r_{k-1})) + (1 - \beta)(\nabla F(\wb^r_{k-1}) - \nabla F(\wb^r_{k})) \\
&~~~ + \beta \bigg(\frac{1}{N}\sum_{i} (G_1(\w^r_{i,k}, \z^r_{i,k,1}, \u^r_{i,k}(\z^r_{i,k,1}), \w^{r-1}_{j,t}, \hat{\z}^{r-1}_{j,t,2})
+ G_2(\w^{r-1}_{j', t'}, \hat{\z}^{r-1}_{j',t',1}, \u^{r-1}_{j',t'}(\hat{\z}^{r-1}_{j',t',1}), \w^r_{i,k}, \z^r_{i,k,2})) \\
%&~~~~~~~~~~~~~ -\frac{1}{N}\sum_{i} (G_1(\wb^{r-1}, \z^r_{i,k,1}, \u^{r}_{i,k}(\z^r_{i,k,1}), \w^{r-1}_{j,t},  \hat{\z}^{r-1}_{j,t,2}) 
%+ G_2(\wb^{r-1}, \hat{\z}^{r-1}_{j',t',1}, \u^{r-1}_{j',t'}(\hat{\z}^{r-1}_{j',t',1}), \w^r_{i,k}, \z^r_{i,k,2})) \bigg) \\ 
%& + \beta \bigg(\frac{1}{N}\sum_{i} (G_1(\wb^{r-1}, \z^r_{i,k,1}, \u^r_{i,k}(\z^r_{i,k,1}), \wb^{r-1}, \hat{\z}^{r-1}_{j,t,2}) 
%+ G_2(\wb^{r-1}, \hat{\z}^{r-1}_{j',t',1}, \u^{r-1}_{j',t'}(\hat{\z}^{r-1}_{j',t',1}), \wb^{r-1}, \z^r_{i,k,2}))  \bigg) \\
%& ~~~~~~~~~~~~~ - \frac{1}{N}\sum_{i} (G_1(\wb^{r-1}, \z^r_{i,k,1}, g(\wb^r, \z^r_{i,k,1}, \wb^r, \S_2), \wb^{r-1}, \hat{\z}^{r-1}_{j,t,2}) \\
%& ~~~~~~~~~~~~~ ~~~~~~~~~~~~~ ~~~~~
%+ G_2(\wb^{r-1}, \hat{\z}^{r-1}_{j',t',1}, g(\wb^{r-1}, \hat{\z}^{r-1}_{j',t',1}, \wb^{r-1}, \S_2), \wb^{r-1}, \z^r_{i,k,2}))  \bigg) \\
%& + \beta \bigg(\frac{1}{N}\sum_{i} (G_1(\wb^{r-1}, \z^r_{i,k,1}, g(\wb^r, \z^r_{i,k,1}, \wb^r, \S_2), \wb^{r-1}, \hat{\z}^{r-1}_{j,t,2}) \\
%& ~~~~~~~~~~~~~ ~~~~~~~~~~~~~ ~~~~~ 
%+ G_2(\wb^{r-1}, \hat{\z}^{r-1}_{j',t',1}, g(\wb^{r-1}, \hat{\z}^{r-1}_{j',t',1}, \wb^{r-1}, \S_2), \wb^{r-1}, \z^r_{i,k,2}))   \\  
& ~~~~~~~~~~~~~ - \frac{1}{N}\sum_{i} (G_1(\wb^{r-1}, \z^r_{i,k,1}, g(\wb^{r-1}, \z^r_{i,k,1}, \wb^{r-1}, \S_2), \wb^{r-1}, \hat{\z}^{r-1}_{j,t,2}) \\
& ~~~~~~~~~~~~~ ~~~~~~~~~~~~~ ~~~~~ 
+ G_2(\wb^{r-1}, \hat{\z}^{r-1}_{j',t',1}, g(\wb^{r-1}, \hat{\z}^{r-1}_{j',t',1}, \wb^{r-1}, \S_2), \wb^{r-1}, \z^r_{i,k,2})) \bigg) \\ 
& + \beta \bigg(\frac{1}{N}\sum_{i} (G_1(\wb^{r-1}, \z^r_{i,k,1}, g(\wb^{r-1}, \z^r_{i,k,1}, \wb^{r-1}, \S_2), \wb^{r-1}, \hat{\z}^{r-1}_{j,t,2}) \\
& ~~~~~~~~~~~~~ ~~~~~~~~~~~~~ ~~~~~  
+ G_2(\wb^{r-1}, \hat{\z}^{r-1}_{j',t',1}, g(\wb^{r-1}, \hat{\z}^{r-1}_{j',t',1}, \wb^{r-1}, \S_2), \wb^{r-1}, \z^r_{i,k,2})) 
- \nabla F(\wb^r_k) \bigg) 
\bigg\|^2.
\end{split}    
\end{equation}  
\end{small} 
% Denoting $g_{\z}(\w) = g(\w, \z, \w, \S_2)$. 
Using Young's inequality and $\tL$-Lipschtzness of $G_1, G_2$, we can then derive
\begin{small} 
\begin{equation}
\begin{split}
&\Delta^r_k \leq (1+\beta) \Bigg\| (1-\beta)(\bG^r_{k-1} - \nabla F(\wb^r_{k-1}) ) \\
&+ \beta \bigg(\frac{1}{N}\sum_{i} (G_1(\wb^{r-1}, \z^r_{i,k,1}, g(\wb^{r-1},\z^r_{i,k,1},\wb^{r-1}, \S_2), \wb^{r-1}, \hat{\z}^{r-1}_{j,t,2}) \\
&~~~~~~~~~~~~~~~~~~~
+ G_2(\wb^{r-1}, \hat{\z}^{r-1}_{j',t',1}, g(\wb^{r-1}, {\hat{\z}^{r-1}_{j',t',1}}, \wb^{r-1}, \S_2), \wb^{r-1}, \z^r_{i,k,2})) 
- \nabla F(\wb^{r-1})\bigg)\Bigg\|^2 \\
& + (1+\frac{10}{\beta})\beta^2 
\left(\frac{1}{N}\sum_i 4\tL^2 \E\|\w^r_{i,k} - \wb^r\|^2 + 4\tL^2 \E\|\wb^r - \wb^{r-1}\|^2
+ \frac{1}{N}\sum_i 4\tL^2 \E\|\w^{r-1}_{j',t'} - \wb^{r-1}\|^2  \right) \\
& + (1+\frac{10}{\beta}) \|\wb^r_{k-1} - \wb^r_k\|^2
+ (1+\frac{10}{\beta})\beta^2 \frac{1}{N}\sum_i  \bigg( \tL^2 \E\|\u^r_{i,k}(\z^r_{i,k,1}) -  g(\wb^{r}_k,\z^r_{i,k,1},\wb^{r}_k, \S_2)\|^2 \\
&~~~~~~~~~~~~~~~~~~~~~~~~~~~~
+ \tL^2 \E\|\u^{r-1}_{j',t'}(\hat{\z}^{r-1}_{j',t',1})-g(\wb^{r-1}_{t'}, \hat{\z}^{r-1}_{j',t',1}, \wb^{r-1}_{t'}, \S_2)\|^2
\bigg). 
\end{split}
\end{equation}
\end{small}
By the fact that 
\begin{equation}
\begin{split}
&\E[\frac{1}{N}\sum_{i} (G_1(\wb^{r-1}, \z^r_{i,k,1}, g(\wb^{r-1},\z^r_{i,k,1},\wb^{r-1},\S_2), \wb^{r-1}, \hat{\z}^{r-1}_{j,t,2}) \\
&~~~~~~~~~~~
+ G_2(\wb^{r-1}, \hat{\z}^{r-1}_{j',t',1}, g(\wb^{r-1},{\hat{\z}^{r-1}_{j',t',1}},\wb^{r-1},\S_2), \wb^{r-1}, \z^r_{i,k,2})) 
- \nabla F(\wb^{r-1})]  = 0,
\end{split}
\end{equation} 
\begin{equation}
\begin{split}
&\E\|\frac{1}{N} \sum_{i} (G_1(\wb^{r-1}, \z^r_{i,k,1}, g(\wb^{r-1},\z^r_{i,k,1},\wb^{r-1},\S_2), \wb^{r-1}, \hat{\z}^{r-1}_{j,t,2})  \\
&~~~~~~~
+ G_2(\wb^{r-1}, \hat{\z}^{r-1}_{j',t',1}, g(\wb^{r-1},{\hat{\z}^{r-1}_{j',t',1}},\wb^{r-1},\S_2), \wb^{r-1}, \z^r_{i,k,2})) 
- \nabla F(\wb^{r-1})  \|^2 
\leq \frac{\sigma^2}{N},  
\end{split} 
\end{equation} 
and 
\begin{equation}
\begin{split}
\|\wb^r_{k-1} - \wb^r_k\|^2 = \eta^2 \|\bar{G}^r_k\|^2 \leq 3\eta^2 \|\bar{G}^r_k - \nabla F(\wb^r_k) \|^2
+ 3\eta^2 \|\nabla F(\wb^r_k) - \nabla F(\wb^{r-1}) \|^2 
+ 3\eta^2 \|\nabla F(\wb^{r-1})\|^2,  
\end{split} 
\end{equation}
we obtain 
%\begin{equation*}
%\begin{split}
%&\Delta^r_k \leq (1-\beta) \|\bG^r_{k-1} - \nabla F(\wb^r_{k-1})\|^2 + 36\frac{\eta^2}{\beta} \| \bar{G}^r_k - \nabla F(\wb^r_k) \|^2 + 36\frac{\eta^2}{\beta} \tL^2\|\wb^r_{k} - \wb^{r-1}\|^2  
%+ 36\frac{\eta^2}{\beta} \|\nabla F(\wb^{r-1})\|^2 \\
%& + \frac{\beta^2 \sigma^2}{N}   
% + 12\beta 
%\left(\frac{1}{N}\sum_i 4\tL^2 \E\|\w^r_{i,k} - \wb^r\|^2 + 4\tL^2 \E\|\wb^r - \wb^{r-1}\|^2
%+ \frac{1}{N}\sum_i 4\tL^2 \E\|\w^{r-1}_{j',t'} - \wb^{r-1}\|^2  \right) \\
%& + 12\beta \frac{1}{N}\sum_i  \bigg( \tL^2 \E\|\u^r_{i,k}(\z^r_{i,k,1}) -  g(\wb^{r}_k,\z^r_{i,k,1},\wb^{r}_k, \S_2)\|^2  
%+ \tL^2 \E\|\u^{r-1}_{j',t'}(\hat{\z}^{r-1}_{j',t',1})-g(\wb^{r-1}_{t'}, \hat{\z}^{r-1}_{j',t',1}, \wb^{r-1}_{t'}, \S_2)\|^2\bigg).  
%\end{split} 
%\end{equation*}
%Therefore,
\begin{equation*}
\begin{split}
&\Delta^r_k \leq (1-\frac{3\beta}{4}) \|\bar{G}_{k-1} - \nabla F(\wb^r_{k-1})\|^2 + \frac{\eta}{16} \|\nabla F(\wb^{r-1})\|^2 
+ 36 \frac{\eta^2}{\beta}\tL^2 \|\wb^r_k - \wb^{r-1}\|^2 \\    
& + \frac{2\beta^2 \sigma^2}{N}   
 + 12\beta 
\left(\frac{1}{N}\sum_i 4\tL^2 \E\|\w^r_{i,k} - \wb^r\|^2 + 4\tL^2 \E\|\wb^r - \wb^{r-1}\|^2
+ \frac{1}{N}\sum_i 4\tL^2 \E\|\w^{r-1}_{j',t'} - \wb^{r-1}\|^2  \right) \\
& + 12\beta \frac{1}{N}\sum_i  \bigg( \tL^2 \E\|\u^r_{i,k}(\z^r_{i,k,1}) -  g(\wb^{r}_k,\z^r_{i,k,1},\wb^{r}_k, \S_2)\|^2  
+ \tL^2 \E\|\u^{r-1}_{j',t'}(\hat{\z}^{r-1}_{j',t',1})-g(\wb^{r-1}_{t'}, \hat{\z}^{r-1}_{j',t',1}, \wb^{r-1}_{t'}, \S_2)\|^2\bigg).  
\end{split}      
\end{equation*}

\end{proof}

%Then,
%\begin{small}
%\begin{equation}
%\begin{split}
%&\frac{1}{K}\sum_k \Delta^r_k \\
%&\leq (1-\frac{\beta}{2}) \frac{1}{K} \sum\limits_{k}\|\bG^r_{k-1} - \nabla F(\wb^r_{k-1})\|^2
%+ 4\frac{\beta^2 \sigma^2}{N} + 5\beta \frac{1}{NK}\sum_i\sum_k\|\wb^{r-1} - \w^r_{i,k}\|^2 + 10\beta \|\u^r_{i,k}(\z^r_{i,k,1}) - g_{\z^r_{i,k,1}}(\wb^r))\|^2. 
%\end{split}
%\end{equation}
%\end{small} 

\subsection{Analysis of Theorem \ref{thm:nonlinear_informal}}
%We re-present Theorem \ref{thm:nonlinear_informal} as below.
%\begin{theorem}
%\label{thm:nonlinear_formal}
%Suppose Assumption \ref{ass:non_linear} holds,
%denoting $M = \max_i |\S^1_i|$ as the largest number of data on a single machine, by setting $\gamma=O(\frac{M^{1/3}}{R^{2/3}})$, $\beta=O(\frac{1}{M^{1/6} R^{2/3}})$, $\eta = O(\frac{1}{M^{2/3} R^{2/3}})$ and $K=O(M^{1/3} R^{1/3})$,  Algorithm \ref{alg:1} ensures that 
%\begin{equation} 
%$\E\left[\frac{1}{R}\sum_{r=1}^R \|\nabla F(\wb^r)\|^2\right] \leq O(\frac{1}{R^{2/3}})$.
%\end{equation}
%\end{theorem}

\begin{proof}
By updating rules, 
\begin{equation}
\begin{split}
&\|\wb^r - \w^r_{i,k}\|^2 
\leq  \eta^2 K^2 C_f^2 C_\ell^2 C_g^2,
\end{split}
\label{app:thm3_eq_1}
\end{equation}
%\begin{equation}
%\begin{split}
%&\|\wb^{r-1} - \wb^r\|^2 = \teta^2 \|\frac{1}{NK}\sum_{i=1}^N\sum_{k=1}^K \bG^{r-1}_{k}\|^2  
%\leq \teta^2 \frac{1}{K} \sum_{k=1}^K \|\bG^{r-1}_k - \nabla F(\wb^{r-1}_{k}) + \nabla F(\wb^{r-1}_{k})\|^2. 
%\end{split}
%\label{app:thm3_eq_3}
%\end{equation} 
and 
\begin{equation}
\begin{split}
&\|\wb^{r}_k - \wb^r\|^2 = \teta^2 \|\frac{1}{NK}\sum_{i=1}^{N} \sum\limits_{m=1}^{k} \bG^r_{m}\|^2 \leq \teta^2 \frac{1}{K} \sum_{m=1}^K \|\bG^r_m - \nabla F(\wb^r_m) + \nabla F(\wb^r_m)\|^2.
\end{split}
\label{app:thm3_eq_2}
\end{equation} 

Similarly, we also have
\begin{equation}
\begin{split}
&\|\wb^{r-1} - \wb^r\|^2 = \teta^2 \|\frac{1}{NK}\sum_{i=1}^N\sum_{k=1}^K \bG^{r-1}_{k}\|^2  
\leq \teta^2 \frac{1}{K} \sum_{k=1}^K \|\bG^{r-1}_k - \nabla F(\wb^{r-1}_{k}) + \nabla F(\wb^{r-1}_{k})\|^2. 
\end{split}
\label{app:thm3_eq_3}
\end{equation}

Lemma \ref{lem:nonlinear_lem_G} gives that
\begin{small}
\begin{equation}
\begin{split}
& \frac{1}{RK}\sum_{r,k} \E\|\bG^r_{k} - \nabla F(\wb^r_{k})\|^2 \leq \frac{\Delta^0_0}{\beta RK} 
+ \frac{1}{8R}\sum_r \|\nabla F(\wb^{r-1})\|^2 + \frac{1}{RK}\sum_{r,k} 50 \frac{\eta^2}{\beta}\tL^2 \|\wb^r_k - \wb^{r-1}\|^2
+ \frac{2\beta\sigma^2}{N}  \\
& + 18 \left(\frac{1}{N}\sum_i 4\tL^2 \E\|\w^r_{i,k} - \wb^r\|^2 + 4\tL^2 \E\|\wb^r - \wb^{r-1}\|^2
+ \frac{1}{N}\sum_i 4\tL^2 \E\|\w^{r-1}_{j',t'} - \wb^{r-1}\|^2  \right) 
\\ 
&+ 18 \frac{1}{R}\sum_r \frac{1}{NK}\sum_{i,k} \frac{1}{|\S_1^i|} \sum_{\z\in\S_1^i} \E\|\u^r_{i,k}(\z) - g(\wb^r, \z, \wb^r, \S_2)\|^2 \\
&~~ + 18\frac{1}{R}\sum_r  \frac{1}{NK} \sum_{j',t'}  \frac{1}{|\S_1^i|} \sum_{\z\in\S_1^i} \|\u^{r-1}_{j',t'}(\z) -  g(\wb^{r-1}_{t'}, \z, \wb^{r-1}_{t'}, \S_2))\|^2, 
\end{split}    
\end{equation} 
\end{small}
which by setting of $\eta$ and $\beta$ leads to
\begin{equation*}
\begin{split}
&\frac{1}{RK}\sum_{r,k} \E\|\bG^r_{k} - \nabla F(\wb^r_{k})\|^2 \leq \frac{2\Delta^0_0}{\beta RK}  
+ \frac{8\beta\sigma^2}{N} 
+ \frac{1}{4R}\sum_r \|\nabla F(\wb^{r-1})\|^2
+ 10 \teta^2 C_\ell^2 C_g^2  \\
&+ 16\frac{1}{R}\sum_r\frac{1}{NK}\sum_{i,k} \frac{1}{|\S_1^i|} \sum_{\z\in\S_1^i} \E\|\u^r_{i,k}(\z) - g(\wb^r; \z, \S_2)\|^2 \\
&+ 32\frac{1}{R}\sum_r \frac{1}{NK} \sum_{j',t'}  \frac{1}{|\S_1^i|} \sum_{\z\in\S_1^i} \|\u^{r-1}_{j',t'}(\hat{\z}^{r-1}_{j',t',1}) -  g(\wb^{r-1};\hat{\z}^{r-1}_{j',t',1}, \S_2))\|^2\\
&+ 32C_g^2 \frac{1}{R} \sum_r\frac{1}{K}\sum_{t'} \|\wb^{r-1} - \wb^{r-1}_{t'}\|^2.  
\end{split} 
\end{equation*} 
Using Lemma \ref{lem:nonlinear_u} yields
\begin{equation*}
\begin{split}
&\frac{1}{R}\sum_r\frac{1}{NK}\sum\limits_{i=1}^{N}\sum\limits_{k=1}^K
\frac{1}{|\S_1^i|}\sum\limits_{\z\in \S_1^i} \E \|\u^r_{i,k}(\z) - g(\wb^r_{k}, \z, \wb^r_{k}, \S_2)\|^2 \\
&\leq \frac{16M}{\gamma} \frac{1}{R} \frac{1}{NK}\sum\limits_{i=1}^{N} 
\frac{1}{|\S_1^i|}\sum\limits_{\z\in \S_1^i}  \E \|\u^0_{i,0}(\z) - g(\wb^0_{0}, \z, \wb^0_{0}, \S_2)\|^2 \\
& + \frac{400M^2}{\gamma^2} \frac{1}{RK}\sum_{r,k} \tL^2\|\wb^r_{k-1} - \wb^r_k\|^2  + 150\gamma (\sigma^2+C_0^2) + 256 \beta^2 K^2 C_0^2  \\ 
& + 128 \tL^2 \frac{|\S_1^i|}{\gamma} (\|\wb^r-\wb^{r-1}\|^2 + \|\wb^r-\wb^{r-1} \|^2) \\
& + 150(\gamma |\S_1^i| + 1)\tL^2 \frac{1}{N}\sum_{i} \|\wb^{r} - \w^r_{i,k}\|^2 
%+ \frac{\gamma \tL^2}{|\S_1^i|} \|\wb^r - \wb^r_k\|^2  
+ 32(\gamma |\S_1^i| + 1) \tL^2 \frac{1}{NK}\sum\limits_{i=1}^N \sum\limits_{k=1}^{K} \E\|\wb^{r-1} - \wb^{r-1}_{i,k}\|^2. 
%&~~~ + 8\gamma M \frac{1}{R}\sum_r \|\wb^r - \wb^{r-1}\|^2 
%+ 8\gamma |\S_1^i| \frac{1}{RNK}\sum_{r,i,k} \|\wb^{r} - \w^r_{i,k}\|^2 
%+ 8\frac{|\S_1^i|}{\gamma}\frac{1}{RK}\sum_{r,k}.  \|\wb^r - \wb^r_k\|^2. 
\end{split}
\end{equation*}
Combining this with previous five inequalities and noting the parameters settings, we obtain
\begin{equation*} 
\begin{split} 
&\frac{1}{R}\sum_r\frac{1}{NK}\sum\limits_{i=1}^{N}\sum\limits_{k=1}^K
\frac{1}{|\S_1^i|}\sum\limits_{\z\in \S_1^i} \E \|\u^r_{i,k}(\z) - g(\wb^r_{k}, \z, \wb^r_{k}, \S_2)\|^2 \\
&\leq O\bigg(\frac{M}{\gamma R K}+ \eta^2 \frac{M^2}{\gamma^2} \frac{1}{RK}\sum_{r,k} \E\|\bG^r_{k} - \nabla F(\wb^r_{k})\|^2 
+\gamma + \beta^2K^2  + \frac{M}{\gamma} \teta^2 (\frac{1}{\beta RK}+ \frac{\beta}{N}) \\
&~~~~~~~~~+ \gamma M \eta^2 K^2  + \frac{1}{4R} \sum_r  \|\nabla F(\wb^{r-1})\|^2 \bigg) 
\end{split}
\label{app:lem1_corollary}
\end{equation*}
and 
\begin{equation}
\begin{split}
&\frac{1}{RK}\sum_{r,k} \E\|\bG^r_{k} - \nabla F(\wb^r_{k})\|^2 \\
&\leq O\left(\frac{M}{\gamma R K}
+\gamma + \beta^2K^2  + \frac{M}{\gamma} \teta^2 (\frac{1}{\beta RK}+ \frac{\beta}{N}) + \gamma M \eta^2 K^2  + \frac{1}{4R} \sum_r \|\nabla F(\wb^{r-1})\|^2 \right). 
\end{split} 
\label{app:lem2_corollary}
\end{equation} 
Then using the standard analysis of smooth function, we derive
\begin{equation}
\begin{split}
&F(\wb^{r+1}) - F(\wb^r) \leq \nabla F(\wb^r)^\top (\wb^{r+1} - \wb^r) + \frac{\tL}{2} \|\wb^{r+1} - \wb^r\|^2 \\
% & = - \teta \nabla F(\wb^r)^\top 
% \left( \frac{1}{NK} \sum_i \sum_k G^r_{i,k}  \right) + \frac{\tL}{2} \|\wb^{r+1} - \wb^r\|^2 \\
& = - \teta \nabla F(\wb^r)^\top 
\left( \frac{1}{NK} \sum_i \sum_k G^r_{i,k} 
 - \nabla F(\wb^r) + \nabla F(\wb^r)
\right) + \frac{\tL}{2} \|\wb^{r+1} - \wb^r\|^2 \\
& = -\teta \|\nabla F(\wb^r)\|^2 + \frac{\teta}{2} \|\nabla F(\wb^r)\|^2 + \frac{\teta}{2} \|  \frac{1}{NK} \sum_i \sum_k G^r_{i,k} 
 - \nabla F(\wb^r) \|^2 \\
&~~~ + \frac{\tL}{2} \|\wb^{r+1} - \wb^r\|^2 \\
& \leq -\frac{\teta}{2} \|\nabla F(\wb^r)\|^2
+\teta \|\frac{1}{NK}\sum_i\sum_k (G^r_{i,k} - \nabla F(\wb^r_k))\|^2 \\
&~~~+ \teta \|\frac{1}{K} \sum_k (\nabla F(\wb^r_k) - \nabla F(\wb^r))\|^2  + \frac{\tL}{2} \|\wb^{r+1} - \wb^r\|^2 \\
&\leq -\frac{\teta}{2} \|\nabla F(\wb^r)\|^2
+\teta \frac{1}{K} \sum_k \|\frac{1}{N}\sum_i (G^r_{i,k} - \nabla F(\wb^r_k))\|^2 \\
&~~~ +\teta \frac{\tL^2}{K} \sum_k \|\wb^r_{k} - \wb^r\|^2 + \frac{\tL}{2} \|\wb^{r+1} - \wb^r\|^2.
\end{split} 
\end{equation}

Combining with (\ref{app:lem2_corollary}), (\ref{app:thm3_eq_1}), (\ref{app:thm3_eq_2}), and (\ref{app:thm3_eq_3}), we derive 
\begin{equation*}
\begin{split}
& \frac{1}{R} \sum_r \E\|\nabla F(\wb^{r})\|^2 \leq O\left(\frac{M}{\gamma R K} +\gamma + \beta^2K^2  + \frac{M}{\gamma} \teta^2 (\frac{1}{\beta RK}+ \frac{\beta}{N}) + \gamma M \eta^2 K^2 \right). 
% &\leq \frac{F(\wb^0) - F(\wb^{R+1})}{\teta R} + \frac{\beta\sigma^2}{N} + \eta^2K^2 P_i + \gamma \sigma^2
\end{split}
\end{equation*}
By setting parameters as in the theorem, we can conclude the proof. 
Further, to get  $\frac{1}{R} \sum_r \E\|\nabla F(\wb^{r})\|^2\leq \epsilon^2$,
we just need to set $\gamma = O(\epsilon^2)$, $\beta = O(\frac{\epsilon^2}{\sqrt{M}})$, $K=O(\frac{\sqrt{M}}{\epsilon})$, $\eta=O(\frac{\epsilon^2}{M})$,
$R=O(\frac{\sqrt{M}}{\epsilon^3})$. 
\end{proof}

\section{FeDXL with Partial Client Participation} 
Considering that not all client machines are available to work at each round,
in this section, we provide an algorithm that allows partial client participation in every round.
The algorithm is given in Algorithm \ref{alg:fedx2_pc}. We use the Assumption \ref{ass:non_linear}.
The convergence results will be presented in Theorem \ref{thm:nonlinear_formal_pc}. 

\begin{algorithm}[ht]  
\caption {FeDXL2: Federated Learning for DXO with non-linear $f$} \label{alg:fedx2}
\begin{algorithmic}[1]
\STATE{On Client $i$: {\bf Require} parameters $\eta, K$} 
\STATE{Initialize model $\w_{i,K}^0$, $\mathcal U_i^{0}=\{u^0(\z)=0,\z\in\S^i_1\}$, $G^0_{i,K}=0$, and buffer $\B_{i,1}, \B_{i, 2}, \mathcal C_i=\emptyset$}
%, \eta, \eta_g, K, \teta = \eta \eta_g K$} 
\STATE{Send $\H^{0}_{i,1}, \H^{0}_{i,2}, \mathcal U^{0}_i$ to the server}%\hfill $\diamond$ or sample $O(1)$ predictions
\STATE{Sample $K$ points from $S_1^i$, compute their predictions using model $\w_{i,0}^{0}$ denoted by $\H^{0}_{i,1}$} %split into $N$ folds and broadcast}
%\STATE{~~~ Each machine receives $P_i^{-1}$}
\STATE{Sample $K$ points from $S_2^i$, compute their predictions using model $\w_{i,0}^{0}$ denoted by $\H^{0}_{i,2}$}
\FOR{$r=1,..., R$} 
\STATE{if $i \not\in P^r$ then skip this round, otherwise do the following} 
\STATE Receives $\wb^{r}, \bar{G}^r$ from the server and set $\w^{r+1}_{i,0} = \wb_{r}, G^{r+1}_{i,0} =  \bar{G}^r$ 
\STATE{Receive $\mathcal R^{r-1}_{i,1}, \mathcal R^{r-1}_{i,2}, \mathcal P^{r-1}$ from the server} 
\STATE {Update  the buffer $\B_{i,1}, \B_{i,2}, \mathcal C_i$ using $\mathcal R^{r-1}_{i,1}, \mathcal R^{r-1}_{i,2}, \mathcal P^{r-1}$ with shuffling, respectively} 
 \STATE{Set $\H^r_{i,1}=\emptyset$, $\H^r_{i,2}=\emptyset, \mathcal U_i^r=\emptyset$}
\FOR{$k=0, .., K-1$}
%\STATE{Each machine $i$:}  
\STATE{Sample $\z^r_{i, k, 1}$ from $\S^i_{1}$, sample $\z^r_{i,k,2}$ from $\S^i_2$} \hfill $\diamond$ or sample two mini-batches of data 
\STATE{Take next $h^{r-1}_\xi$,  $h^{r-1}_{\zeta}$ and $u^{r-1}_\zeta$ from  $\B_{i,1}$ and $\B_{i,2}$ and $\mathcal C_i$, respectively} 
% , respectively, w/o replacement}   
\STATE{Compute $h(\w^r_{i,k}, \z^r_{i,k,1})$ and $h(\w^r_{i,k}, \z^r_{i,k,2})$} 
\STATE{Compute $h(\w^r_{i,k}, \hat{\z}^r_{i,k,1})$ and $h(\w^r_{i,k}, \hat{\z}^r_{i,k,2})$} and add them to $\H^r_{i,1}, \H^{r}_{i, 2}$, respectively
%\STATE{Add $h(\w^r_{i,k}, \z^r_{i,k,1})$ into $\H^r_{i,1}$ and add $h(\w^r_{i,k}, \z^r_{i,k,2})$ into $\H^r_{i,2}$} 
\STATE Compute $\u^r_{i,k}(\z^r_{i,k,1})$ according to~(\ref{eqn:u} and add it to $\mathcal U_i^{r}$
% \STATE{~~~ For $p \in B_{1, t}^k$: Update $\u_{p, t+1} = \u_{p, t} - \gamma (\u_{p, t} - \sum_{q\in B^k_{2, t}}\ell(f(\w_t^k;p) - f(\w_{t_0}; q)) )$ } 
\STATE {Compute $G^r_{i,k,1}$ and $G^r_{i, k, 2}$ according to~(\ref{eqn:G1G2_G1}), (\ref{eqn:G1G2_G2})}%{$G^r_{i,k,1} = G_1(\w^r_{i, k}, p^r_{i, k}, \w^{r-1}_{i', k'}, q^{r-1}_{i', k'}) $ }
%\STATE{($i', k'$ denotes unknown machine index $i$ and unknown iteration index $k$)}  
%\STATE{$G^r_{i, k, 2} = G_2(\w^{r-1}_{j,t}, p^{r-1}_{j,t}, \w^r_{i, k}, q^r_{i, k}) $}  
\STATE{$G^r_{i,k} = (1-\beta) G^r_{i,k-1} + \beta (G^r_{i,k,1} + G^r_{i, k, 2})$}    
\STATE{$\w^r_{i, k+1} = \w^r_{i, k} - \eta G^r_{i,k}$}
\ENDFOR  
\STATE Sends $\w^{r}_{i,K}, G^r_{i,k}$ to the server
\STATE{Send $\H^{r}_{i,1}, \H^{r}_{i,2}, \mathcal U_i^{r}$ to the server}%\hfill $\diamond$ or sample $O(1)$ predictions
\ENDFOR
\vspace*{0.1in}
\hrule
\vspace*{0.05in}
\STATE{On Server} 
\STATE Collects $\H^{0}_{*}=\H^{0}_{1,*}\cup\H^{0}_{2,*}\ldots\cup\H^{0}_{N,*}$ and $\mathcal U^{0}=\mathcal U^{0}_{1}\cup\mathcal U^{0}_{1}\ldots\cup\mathcal U^{0}_{N}$, where $*=1, 2$ 
\FOR{$r=1,..., R$} 
\STATE{Sample a set $P^r$ of clients to participant this round} 
\STATE Receive $\w^{r-1}_{i,K}$,$G^{r-1}_{i,K}$ from client $i \in P^{r-1}$, compute $\wb^{r} = \frac{1}{|P^{r-1}|}\sum_{i\in P^{r-1}} \w^{r-1}_{i, K}$, $G^{r} = \frac{1}{|P^{r-1}|}\sum_{i\in P^{r-1}} G^{r-1}_{i, K}$.  
\STATE{Broadcast $\wb^{r}$ and $G^r$ to clients in $P^r$ }
\STATE Set $\mathcal R^{r-1}_{i,1}=\H^{r-1}_1, \mathcal R^{r-1}_{i,2}=\H^{r-1}_2, \mathcal P^{r-1}_i=\mathcal U^{r-1}$ and send them to Client $i$ for all $i\in P^r$
\STATE Collects $\H^{r}_{*}=\cup \H^{r}_{i,*}, \forall i\in P^r$  and $\mathcal U^{r}=\cup \mathcal U^{r}_{i}, \forall i \in P^r$, where $*=1, 2$ 
\ENDFOR  
\end{algorithmic}  
\label{alg:fedx2_pc}
\end{algorithm}

\subsection{Analysis of the moving average estimator $\u$}  
\begin{lemma}
\label{lem:nonlinear_u_pc}
Under Assumption \ref{ass:non_linear}, the moving average estimator $\u$ satisfies 
\begin{equation*} 
\begin{split}
&\frac{1}{N} \sum\limits_{i=1}^{N} \frac{1}{|\S_1^i|} \sum\limits_{\z \in |\S_1^i|} \E\|\u^r_{i,k}(\z) - g(\wb^r_{k}, \z, \wb^r_k, \S_2)\|^2 \\
&\leq (1-\frac{\gamma |P^r|}{16|\S_1^i|N})\frac{1}{N} \sum\limits_{i=1}^{N} \frac{1}{|\S_1^i|}  \sum\limits_{\z \in |\S_1^i|} [\E\|\u^r_{i,k-1}(\z) - g(\wb^r_{k-1}, \z, \wb^r_{k-1}, \S_2)\|^2 \\
& + \frac{20|\S_1^i|N}{\gamma |P^r|} \tL^2\|\wb^r_{k-1} - \wb^r_k\|^2 ]  + 8\frac{\gamma^2}{|\S_1^i|} \frac{|P^r|}{N} (\sigma^2+C_0^2)   + \frac{16\gamma \beta^2 K^2 C_0^2 |P^r|}{|\S_1^i| N} \\
& + 8 \frac{|P^r|}{N} \tL^2 \|\wb^r - \wb^{r-1}\|^2 + 8 \tL^2 \frac{|P^r|}{N} \|\wb^r - \wb^{r}_k\|^2 \\
&+ 8(\gamma^2 + \frac{\gamma}{|\S_1^i|} )\tL^2 \frac{1}{N}\sum_{i\in P^r} \|\wb^{r} - \w^r_{i,k}\|^2 
%+ \frac{\gamma \tL^2}{|\S_1^i|} \|\wb^r - \wb^r_k\|^2  
+ 2(\gamma^2 + \frac{\gamma}{|\S_1^i|} ) \tL^2 \frac{1}{NK}\sum\limits_{i\in P^r} \sum\limits_{k=1}^{K} \E\|\wb^{r-1} - \wb^{r-1}_{i,k}\|^2. 
\end{split} 
\end{equation*} 
\end{lemma}

\begin{proof}
Denote $P^r$ as the clients that are sampled to take participation in the $r$-th round. 
By update rules of $\u$, we have 
\begin{equation} 
\begin{split} 
\u_{i, k}^{r} (\z) = \left\{ \begin{array}{cc} 
    \u_{i, k-1}^r (\z) - \gamma (\u^r_{i,k-1}(\z) - \ell(h(\w^r_{i,k}, \z^r_{i,k,1}), h(\w^{r-1}_{j,t}, \hat{\z}^{r-1}_{j,t,2}))),  & i\in P^r ~and~ \z=\z^r_{i,k,1}  \\ 
      \u_{i, k-1}^r (\z),  & otherwise. 
\end{array} \right. 
\end{split}  
\end{equation}  

Or equivalently, 
\begin{equation}
\begin{split}
\u_{i, k}^{r} (\z) = \left\{ \begin{array}{cc}
    \u_{i, k-1}^r (\z) - \gamma (\u^r_{i,k-1}(\z) - g(\w^r_{i,k}, \z^r_{i,k,1}, \w^{r-1}_{j,t}, \hat{\z}^{r-1}_{j,t,2})),  & i\in P^r ~and~ \z=\z^r_{i,k,1}  \\ 
      \u_{i, k-1}^r (\z),  & otherwise.
\end{array} \right. 
\end{split}  
\end{equation}

Define $\bar{\u}^r_k = (\u^r_{1, k}, \u^r_{2, k}, ..., \u^r_{N, k})$, $\wb^r_k = \frac{1}{|P^r|} \sum\limits_{i\in P^r} \w^r_{i,k}$.
%and 
%\begin{equation} 
%\begin{split} 
%&\phi^r_{k} (\bar{\u}^r_{k}) 
%= \frac{1}{2N} \sum\limits_{i=1}^N \frac{1}{|\S_i|} \sum_{\z\in \S_1^i} \|\u^r_{i,k} (\z) - g(\wb^r_{k}, \z, \wb^r_k, \S_2)\|^2.
% &= \frac{1}{2N} \sum\limits_{i=1}^N \frac{1}{|\S_1^i|} \sum_{\z\in \S_1^i}  \left\|\u^r_{i,k}(\z) - \frac{1}{N}\sum\limits_{i=1}^N \frac{1}{Q_i} \sum_{q\in Q_i} \ell(f(\wb^r_{k}; p) - f(\wb^r_{k}; q)) \right\|^2.
%\end{split}     
%\end{equation}  
Then it follows that
%\begin{small} 
\begin{equation} 
\begin{split}
% &\frac{1}{2}\phi^r_{k}(\bar{\u}^r_{k}) = 
&\frac{1}{2N} \sum\limits_{i=1}^{N} \frac{1}{|\S_1^i|} \sum\limits_{\z \in |\S_1^i|} \E\|\u^r_{i,k}(\z) - g(\wb^r_{k}, \z, \wb^r_k, \S_2)\|^2 \\ 
&=\frac{1}{N} \sum_i \frac{1}{|\S_1^i|} \sum_{\z\in |\S_1^i|} \E\bigg[ \frac{1}{2} \|\u^r_{i,k-1}(\z) - g(\wb^r_{k}, \z, \wb^r_k, \S_2)\|^2 \\
&~~~~~~ +  \langle \u^r_{i,k-1}(\z) - g(\wb^r_{k}, \z, \wb^r_k, \S_2), \u^r_{i,k}(\z) - \u^r_{i,k-1}(\z) \rangle 
%&~~~~~~~~~~~~~~~~~~~~~~~~~~~~~~~~~~~~~~~~~~~~~~~~~~~~~
+ \frac{1}{2}\|\u^r_{i,k}(\z) - \u^r_{i,k-1}(\z)\|^2 \bigg]\\
&=\frac{1}{2N} \sum_i \frac{1}{|\S_i|} \sum_{\z\in \S_1^i} \E\|\u^r_{i,k-1}(\z) - g(\wb^r_{k}, \z, \wb^r_k, \S_2)\|^2\\
%&~~~~~~~~~~~~~~~~~~~~~~~~~~~~~~~~~~~~
&~~~ +  \E \frac{1}{N} \sum_{i\in P^r} 
\frac{1}{|\S_1^i|}\langle \u^r_{i,k-1}(\z^r_{i,k,1}) - g(\wb^r_{k}, \z^r_{i,k,1}, \wb^r_k, \S_2),  \u^r_{i,k}(\z^r_{i,k,1}) - \u^r_{i,k-1}(\z^r_{i,k,1}) \rangle \\ 
% &~~~~~~~~~~~~ ~~~~~~~~~~~~ ~~~~~~~~~~~  
&~~~ + \frac{1}{N} \sum_i \frac{1}{2|\S_1^i|}\E\|\u^r_{i,k}(\z^r_{i,k,1}) - \u^r_{i,k-1}(\z^r_{i,k,1})\|^2 \\  
& = \frac{1}{2N} \sum_i \frac{1}{|\S_i|} \sum_{\z\in \S_1^i} \E\|\u^r_{i,k-1}(\z) - g(\wb^r_{k}, \z, \wb^r_k, \S_2)\|^2 \\
&~~~ + \E[\frac{1}{N} \sum_{i\in P^r} \frac{1}{|\S_1^i|} \langle \u^r_{i,k-1}(\z^r_{i,k,1}) -  g(\w^r_{i,k}, \z^r_{i,k,1}, \w^{r-1}_{j,t}, \hat{\z}^{r-1}_{j,t,2}), \u^r_{i,k}(\z^r_{i,k,1}) - \u^r_{i,k-1}(\z^r_{i,k,1}) \rangle ] \\ 
&~~~ + \E[\frac{1}{N} \sum_{i\in P^r} \frac{1}{|\S_1^i|} \langle  g(\w^r_{i,k}, \z^r_{i,k,1}, \w^{r-1}_{j,t}, \hat{\z}^{r-1}_{j,t,2}) - g(\wb^r_{k}, \z^r_{i,k,1}, \wb^r_k, \S_2), \u^r_{i, k}(\z^r_{i,k,1}) - \u^r_{i,k-1}(\z^r_{i,k,1}) \rangle]
\\
&~~~ + \E [\frac{1}{N} \sum_{i\in P^r} \frac{1}{2 |\S_i|} \|\u^r_{i,k}(\z^r_{i,k,1}) - \u^r_{i, k-1}(\z^r_{i,k,1})\|^2], 
\end{split} 
\end{equation} 
%\end{small} 
where for $i \in P^r$ it has
\begin{equation}
\begin{split}
& \langle \u^r_{i,k-1}(\z^r_{i,k,1}) - g(\w^r_{i,k}, \z^r_{i,k,1}, \w^{r-1}_{j,t}, \hat{\z}^{r-1}_{j,t,2}), \u^r_{i,k}(\z^r_{i,k,1}) - \u^r_{i,k-1}(\z^r_{i,k,1}) \rangle \\ 
&= \langle \u^r_{i,k-1}(\z^r_{i,k,1}) - g(\w^r_{i,k}, \z^r_{i,k,1}, \w^{r-1}_{j,t}, \hat{\z}^{r-1}_{j,t,2}), g(\wb^r_{k},\z^r_{i,k,1},\wb^r_{k}, \S_2) - \u^r_{i,k-1}(\z^r_{i,k,1}) \rangle \\
&~~~ 
+ \langle \u^r_{i,k-1}(\z^r_{i,k,1}) - g(\w^r_{i,k}, \z^r_{i,k,1}, \w^{r-1}_{j,t}, \hat{\z}^{r-1}_{j,t,2}), \u^r_{i,k}(\z^r_{i,k,1}) - g(\wb^r_{k},\z^r_{i,k,1},\wb^r_{k}, \S_2)  \rangle \\ 
& = \langle \u^r_{i,k-1}(\z^r_{i,k,1}) -  g(\w^r_{i,k}, \z^r_{i,k,1}, \w^{r-1}_{j,t}, \hat{\z}^{r-1}_{j,t,2}),  g(\wb^r_{k},\z^r_{i,k,1},\wb^r_{k}, \S_2)  - \u^r_{i,k-1}(\z^r_{i,k,1}) \rangle \\
&~~~ 
+\frac{1}{\gamma}\langle \u^r_{i,k-1}(\z^r_{i,k,1}) - \u^r_{i,k}(\z^r_{i,k,1}), \u^r_{i,k}(\z^r_{i,k,1}) - g(\wb^r_{k},\z^r_{i,k,1},\wb^r_{k}, \S_2)  \rangle \\
& = \langle \u^r_{i,k-1}(\z^r_{i,k,1}) - g(\w^r_{i,k}, \z^r_{i,k,1}, \w^{r-1}_{j,t}, \hat{\z}^{r-1}_{j,t,2}), g(\wb^r_{k},\z^r_{i,k,1},\wb^r_{k},\S_2) - \u^r_{i,k-1}(\z^r_{i,k,1}) \rangle \\  
&~~~ 
+ \frac{1}{2\gamma} (\|\u^r_{i,k-1}(\z^r_{i,k,1}) - g(\wb^r_{k},\z^r_{i,k,1},\wb^r_{k}, \S_2) \|^2 - \|\u^r_{i,k}(\z^r_{i,k,1}) - \u^r_{i, k-1}(\z^r_{i,k,1})\|^2 \\
&~~~ ~~~ ~~~~~~ - \|\u^r_{i,k}(\z^r_{i,k,1}) -  g(\wb^r_{k},\z^r_{i,k,1},\wb^r_{k}, \S_2) \|^2)
\end{split}      
\end{equation}   

If $\gamma \leq \frac{1}{5}$, we have  for $i\in P^r$
\begin{equation}
\begin{split}
& -\frac{1}{2} \left( \frac{1}{\gamma} - 1 - \frac{\gamma+1}{4\gamma}  \right) \E\|\u^r_{i,k}(\z^r_{i,k,1}) -   \u^r_{i,k-1}(\z^r_{i,k,1})\|^2 \\
&~~~ + \E \langle  g(\w^r_{i,k}, \z^r_{i,k,1}, \w^{r-1}_{j,t}, \hat{\z}^{r-1}_{j,t,2}) - g(\wb^r_{k}, \z^r_{i,k,1},\wb^r_k, \S_2), \u^r_{i,k}(\z^r_{i,k,1}) - \u^r_{i,k-1}(\z^r_{i,k,1}) \rangle \\
&\leq  -\frac{1}{4\gamma}   \E\|\u^r_{i,k}(\z^r_{i,k,1}) - \u^r_{i,k-1}(\z^r_{i,k,1})\|^2 \\
&~~~ + \gamma \E\|g(\w^r_{i,k}, \z^r_{i,k,1}, \w^{r-1}_{j,t}, \hat{\z}^{r-1}_{j,t,2}) - g(\wb^r_{k}, \z^r_{i,k,1},\wb^r_k, \S_2)\|^2 \\
&~~~ 
+ \frac{1}{4\gamma} \E \|\u^r_{i,k}(\z^r_{i,k,1}) - \u^r_{i,k-1}(\z^r_{i,k,1})\|^2 \\ 
& \leq  \gamma \E\|g(\w^r_{i,k}, \z^r_{i,k,1}, \w^{r-1}_{j,t}, \hat{\z}^{r-1}_{j,t,2}) - g(\wb^r_{k}, \z^r_{i,k,1},\wb^r_k, \S_2)\|^2 \\
&\leq 4\gamma  \E\|g(\wb^{r-1}, \z^r_{i,k,1}, \wb^{r-1}, \hat{\z}^{r-1}_{j,t,2}) - g(\wb^{r-1},\z^r_{i,k,1}, \wb^{r-1}, \S_2)\|^2 + 4\gamma\tL^2 \E\|\wb^{r} - \wb^{r-1}\|^2 \\
&~~~ + 4\gamma\tL^2 \E\|\w^r_{i,k} - \wb^{r}\|^2 + 4\gamma\tL^2 \E\|\w^{r-1}_{j,t} - \wb^{r-1}\|^2 \\
&\leq 4\gamma  \sigma^2 + 4\gamma\tL^2 \E\|\wb^{r} - \wb^{r-1}\|^2 + 4\gamma\tL^2 \E\|\w^r_{i,k} - \wb^{r}\|^2 + 4\gamma\tL^2 \E\|\w^{r-1}_{j,t} - \wb^{r-1}\|^2.
\end{split} 
\end{equation}

Then, we have
\begin{equation}
\begin{split}
&\frac{1}{2N} \sum\limits_{i=1}^{N} \frac{1}{|\S_1^i|} \sum\limits_{\z \in |\S_1^i|} \E\|\u^r_{i,k}(\z) - g(\wb^r_{k}, \z, \wb^r_k, \S_2)\|^2 \\
&\leq \frac{1}{2N} \sum\limits_{i=1}^{N} \frac{1}{|\S_1^i|} \sum\limits_{\z \in |\S_1^i|} \E\|\u^r_{i,k-1}(\z) - g(\wb^r_{k}, \z, \wb^r_k, \S_2)\|^2 \\
& + \frac{1}{N}\sum_{i\in P^r} \frac{1}{|\S_1^i|} \Bigg[ \frac{1}{2\gamma}  \E\|\u^r_{i,k-1}(\z^r_{i,k,1}) - g(\wb^{r}_k,\z^r_{i,k,1}, \wb^{r}_k, \S_2)\|^2 \\
& - \frac{1}{2\gamma} \E\|\u^r_{i,k}(\z^r_{i,k,1})- g(\wb^{r}_{k},\z^r_{i,k,1}, \wb^{r}_k, \S_2)\|^2\! -\! \frac{\gamma+1}{8\gamma} \|\u^r_{i,k}(\z^r_{i,k,1})  - \u^r_{i,k-1}(\z^r_{i,k,1}) \|^2 \!+\! 4\gamma \sigma^2 \\
 % + \gamma  \|g(\wb^{r-1}, \z^r_{i,k,1}, \wb^{r-1}, \hat{\z}^{r-1}_{j,t,2}) - g(\wb^{r-1}, \z^r_{i,k,1}, \wb^{r-1}, \S_2)\|^2 \\ 
 & + 4\gamma \tL^2 \E\|\wb^r-\wb^{r-1}\|^2
 + 4\gamma \tL^2 \E\|\w^r_{i,k} - \wb^r\|^2
 +4 \gamma\tL^2\E\|\w^{r-1}_{j,t} - \wb^{r-1}\|^2 \\
&+ \E \langle \u^r_{i,k-1}(\z^r_{i,k,1}) - g(\w^r_{i,k}, \z^r_{i,k,1}, \w^{r-1}_{j,t}, \hat{\z}^{r-1}_{j,t,2}), g(\wb^r_{k}, \z^r_{i,k,1}, \wb^r_k, \S_2) - \u^r_{i, k-1}(\z^r_{i,k,1}) \rangle  \Bigg].
\end{split}
\end{equation}  

Note that for $i\in P^r$, $\sum_{\z\neq \z^r_{i,k,1}}\|\u^r_{i,k-1}(\z) - g(\wb^r_{k}, \z, \wb^r_{k}, \S_2 )\|^2 = \sum_{\z\neq \z^r_{i,k,1}}\|\u^r_{i, k}(\z) - g(\w^r_{k}, \z, \wb^r_{k}, \S_2)\|^2 $, which implies for $i\in P^r$
\begin{equation}  
\begin{split}     
&\frac{1}{2\gamma} \left( \|\u^r_{i,k-1}(\z^r_{i,k,1}) - g(\wb^r_{k}, \z^r_{i,k,1}, \wb^r_{k}, \S_2 )\|^2 - \|\u^r_{i, k}(\z^r_{i,k,1}) - g(\wb^r_{k}, \z^r_{i,k,1}, \wb^r_{k}, \S_2 )\|^2 \right) \\ 
&= \frac{1}{2\gamma}\sum_{\z\in \S_1^i} \left( \|\u^r_{i,k-1}(\z) - g(\wb^r_{k}, \z, \wb^r_{k}, \S_2 )\|^2 - \|\u^r_{i,k}(\z) - g(\wb^r_{k}, \z, \wb^r_{k}, \S_2 )\|^2 \right).  
\end{split}    
\end{equation} 

Since $\ell(\cdot) \leq C_0$, we have that $\|g(\cdot)\|^2 \leq C_0^2$, $\|\u^r_{i,k}(\z)\|^2 \leq C_0^2$ and $$\|\u^r_{i,k}(\z) - \u^{r}_{i,0}(\z)\|^2 \leq \beta^2 K^2 C_0^2.$$
Besides, we have for $i\in P^r$ that
%\begin{small}   
\begin{equation}  
\begin{split}
& \E\langle \u^r_{i,k-1}(\z^r_{i,k,1}) -  g(\w^r_{i,k}, \z^r_{i,k,1}, \w^{r-1}_{j,t}, \hat{\z}^{r-1}_{j,t,2}), g(\wb^r_{k}, \z^r_{i,k,1},\wb^r_{k},\S_2) -  \u^r_{i,k-1}(\z^r_{i,k,1}) \rangle \\ 
& = \E\langle \u^r_{i,k-1}(\z^r_{i,k,1}) -  g(\wb^{r-1}, \z^r_{i,k,1}, \wb^{r-1}, \hat{\z}^{r-1}_{j,t,2}), g(\wb^r_{k}, \z^r_{i,k,1},\wb^r_{k},\S_2) -  \u^r_{i,k-1}(\z^r_{i,k,1}) \rangle \\
&  + \E\langle g(\wb^{r-1}, \z^r_{i,k,1}, \wb^{r-1}, \hat{\z}^{r-1}_{j,t,2})  -  g(\w^r_{i,k}, \z^r_{i,k,1}, \w^{r-1}_{j,t}, \hat{\z}^{r-1}_{j,t,2}), g(\wb^r_{k}, \z^r_{i,k,1},\wb^r_{k},\S_2) -  \u^r_{i,k-1}(\z^r_{i,k,1}) \rangle \\
&\leq \E\langle \u^r_{i,k-1}(\z^r_{i,k,1}) -  g(\wb^{r-1}, \z^r_{i,k,1}, \wb^{r-1}, \hat{\z}^{r-1}_{j,t,2}) , 
g(\wb^r_{k}, \z^r_{i,k,1},\wb^r_{k},\S_2) - g(\wb^{r-1}, \z^r_{i,k,1},\wb^{r-1},\S_2) 
\rangle \\
&+ \E\langle \u^r_{i,k-1}(\z^r_{i,k,1}) -  g(\wb^{r-1}, \z^r_{i,k,1}, \wb^{r-1}, \hat{\z}^{r-1}_{j,t,2}), g(\wb^{r-1}, \z^r_{i,k,1},\wb^{r-1},\S_2) -  \u^r_{i,k-1}(\z^r_{i,k,1}) 
\rangle \\
& + 2\tL^2\E\|\wb^{r} - \wb^{r-1}\|^2 + 2\tL^2 \E\|\wb^{r} - \w^r_{i,k}\|^2 
+ \tL^2 \E\|\wb^{r-1} - \w^{r-1}_{j,t}\|^2 \\
& + \frac{1}{4}\E\|g(\wb^r, \z^r_{i,k,1}, \wb^r_k, \S_2) - \u^r_{i,k-1}(\z^r_{i,k,1}) \|^2 \\
&\leq 2\gamma  C_0^2 + \frac{1}{\gamma} \|\wb^r_k - \wb^{r-1}\|^2 \\
& + \E\langle \u^r_{i,k-1}(\z^r_{i,k,1}) -  g(\wb^{r-1}, \z^r_{i,k,1}, \wb^{r-1}, \hat{\z}^{r-1}_{j,t,2}), g(\wb^{r-1}, \z^r_{i,k,1},\wb^{r-1},\S_2) -  \u^r_{i,k-1}(\z^r_{i,k,1}) 
\rangle \\
&  + 2\tL^2\E\|\wb^{r} - \wb^{r-1}\|^2 + 2\tL^2 \E\|\wb^{r} - \w^r_{i,k}\|^2 
+ \tL^2 \E\|\wb^{r-1} - \w^{r-1}_{j,t}\|^2 \\
& + \frac{1}{4}\E\|g(\wb^r, \z^r_{i,k,1}, \wb^r_k, \S_2) - \u^r_{i,k-1}(\z^r_{i,k,1}) \|^2,
\end{split}
\end{equation} 
%\end{small}
% \|g(\wb^{r-1}, \z^r_{i,k,1}, \wb^{r-1}, S_2) - \u^r_{i, k-1}(\z^r_{i,k,1})\|^2 \\
where
%\begin{small} 
\begin{equation}
\begin{split}
&\E\langle \u^r_{i,k-1}(\z^r_{i,k,1}) -  g(\wb^{r-1}, \z^r_{i,k,1}, \wb^{r-1}, \hat{\z}^{r-1}_{j,t,2}), g(\wb^{r-1}, \z^r_{i,k,1},\wb^{r-1},\S_2) -  \u^r_{i,k-1}(\z^r_{i,k,1}) 
\rangle \\
&=\E\langle \u^r_{i,k-1}(\z^r_{i,k,1}) - \u^{r-1}_{i,0}(\z^r_{i,k,1}) + \u^{r-1}_{i,0}(\z^r_{i,k,1})  -  g(\wb^{r-1}, \z^r_{i,k,1}, \wb^{r-1}, \hat{\z}^{r-1}_{j,t,2}), \\
&~~~~~~~~~ g(\wb^{r-1}, \z^r_{i,k,1},\wb^{r-1},\S_2) - \u^{r-1}_{i,0}(\z^r_{i,k,1}) + \u^{r-1}_{i,0}(\z^r_{i,k,1}) -  \u^r_{i,k-1}(\z^r_{i,k,1}) 
\rangle \\
&\leq \E\langle \u^r_{i,k-1}(\z^r_{i,k,1}) - \u^{r-1}_{i,0}(\z^r_{i,k,1}), g(\wb^{r-1}, \z^r_{i,k,1},\wb^{r-1},\S_2) - \u^{r-1}_{i,0}(\z^r_{i,k,1})\rangle \\
& + \E\langle \u^r_{i,k-1}(\z^r_{i,k,1}) - \u^{r-1}_{i,0}(\z^r_{i,k,1}),  \u^{r-1}_{i,0}(\z^r_{i,k,1}) -  \u^r_{i,k-1}(\z^r_{i,k,1})  \rangle \\
& + \E\langle \u^{r-1}_{i,0}(\z^r_{i,k,1})  -  g(\wb^{r-1}, \z^r_{i,k,1}, \wb^{r-1}, \hat{\z}^{r-1}_{j,t,2}), g(\wb^{r-1}, \z^r_{i,k,1},\wb^{r-1},\S_2) - \u^{r-1}_{i,0}(\z^r_{i,k,1})\rangle \\
& + \E\langle  \u^{r-1}_{i,0}(\z^r_{i,k,1})  -  g(\wb^{r-1}, \z^r_{i,k,1}, \wb^{r-1}, \hat{\z}^{r-1}_{j,t,2}), \u^{r-1}_{i,0}(\z^r_{i,k,1}) -  \u^r_{i,k-1}(\z^r_{i,k,1}) \rangle \\
&\leq 4 \E\|\u^r_{i,k-1}(\z^r_{i,k,1}) - \u^{r-1}_{i,0}(\z^r_{i,k,1})\|^2
+ \frac{1}{4} \E\|g(\wb^{r-1}, \z^r_{i,k,1},\wb^{r-1},\S_2) - \u^{r-1}_{i,0}(\z^r_{i,k,1})\|^2 \\
& - \E\|g(\wb^{r-1}, \z^r_{i,k,1},\wb^{r-1},\S_2) - \u^{r-1}_{i,0}(\z^r_{i,k,1})\|^2 \\
& + \frac{1}{4}\E\|g(\wb^{r-1}, \z^r_{i,k,1},\wb^{r-1},\S_2) - \u^{r-1}_{i,0}(\z^r_{i,k,1})\|^2 
+ 4 \E\|\u^r_{i,k-1}(\z^r_{i,k,1}) - \u^{r-1}_{i,0}(\z^r_{i,k,1})\|^2 \\
&\leq  4 \E\|\u^r_{i,k-1}(\z^r_{i,k,1}) - \u^{r-1}_{i,0}(\z^r_{i,k,1})\|^2 
% + \frac{1}{4} \E\|g(\wb^{r-1}, \z^r_{i,k,1},\wb^{r-1},\S_2) - \u^{r-1}_{i,0}(\z^r_{i,k,1})\|^2 \\
-\frac{1}{2}\E\|g(\wb^{r-1}, \z^r_{i,k,1},\wb^{r-1},\S_2) - \u^{r-1}_{i,0}(\z^r_{i,k,1})\|^2 
+ 8\beta^2 K^2 C_0^2. 
\end{split}
\end{equation}
%\end{small} 
Noting for $i\in P^r$,
\begin{equation}
\begin{split}
&-\E\|g(\wb^{r-1}, \z^r_{i,k,1},\wb^{r-1},\S_2) - \u^{r-1}_{i,0}(\z^r_{i,k,1})\|^2 \\
&=-\E\|g(\wb^{r-1}, \z^r_{i,k,1},\wb^{r-1},\S_2) - \u^{r}_{i,k-1}(\z^r_{i,k,1}) + \u^{r}_{i,k-1}(\z^r_{i,k,1}) - \u^{r-1}_{i,0}(\z^r_{i,k,1})\|^2 \\
&=-\E\|g(\wb^{r-1}, \z^r_{i,k,1},\wb^{r-1},\S_2) - \u^{r}_{i,k-1}(\z^r_{i,k,1})\|^2 - \E \| \u^{r}_{i,k-1}(\z^r_{i,k,1}) - \u^{r-1}_{i,0}(\z^r_{i,k,1})\|^2 \\
&~~~ +2\E\langle g(\wb^{r-1}, \z^r_{i,k,1},\wb^{r-1},\S_2) - \u^{r}_{i,k-1}(\z^r_{i,k,1}),  \u^{r}_{i,k-1}(\z^r_{i,k,1}) - \u^{r-1}_{i,0}(\z^r_{i,k,1}) \rangle \\
&\leq -\frac{1}{2} \E\|g(\wb^{r-1}, \z^r_{i,k,1},\wb^{r-1},\S_2) - \u^{r}_{i,k-1}(\z^r_{i,k,1})\|^2 + 8 \| \u^{r}_{i,k-1}(\z^r_{i,k,1}) - \u^{r-1}_{i,0}(\z^r_{i,k,1})\|^2 \\
&\leq -\frac{1}{2} \E\|g(\wb^{r-1}, \z^r_{i,k,1},\wb^{r-1},\S_2) - \u^{r}_{i,k-1}(\z^r_{i,k,1})\|^2 + 8\beta^2K^2 C_0^2 \\
&\leq -\frac{1}{4} \E\|g(\wb^{r}_k, \z^r_{i,k,1},\wb^{r}_k,\S_2) - \u^{r}_{i,k-1}(\z^r_{i,k,1})\|^2
+ \frac{1}{2} \tL^2\|\wb^{r-1} - \wb^r_k\|^2
+ 8\beta^2K^2 C_0^2.
\end{split} 
\end{equation} 

With the client sampling and data sampling, we observe that
\begin{equation}
\begin{split}
    &-\E \left[\frac{1}{N}\sum_{i\in P^r} \frac{1}{|\S^i_1|}\|g(\wb^r_k, \z^r_{i,k,1}, \wb^r_k, \S_2) - \u^r_{i,k-1}(\z^r_{i,k,1})\|^2 \right] \\
    &= - \frac{1}{N} \frac{|P^r|}{N}\sum_{i=1}^N \E_{\z^r_{i,k,1}\in \S^i_1} \left[\frac{1}{|\S^i_1|}\|g(\wb^r_k, \z^r_{i,k,1}, \wb^r_k, \S_2) - \u^r_{i,k-1}(\z^r_{i,k,1})\|^2 \right].
\end{split}
\end{equation}

Then by multiplying $\gamma$ to every term and rearranging terms using the setting of $\gamma \leq O(1)$, 
we can obtain    
\begin{equation} 
\begin{split}
& \frac{\gamma+1}{2} \frac{1}{N} \sum\limits_{i=1}^{N} \frac{1}{|\S_1^i|} \sum\limits_{\z \in |\S_1^i|} \E\|\u^r_{i,k}(\z) - g(\wb^r_{k}, \z, \wb^r_k, \S_2)\|^2 \\
&\leq \frac{\gamma(1-\frac{|P^r|}{8|\S_1^i|N}) + 1}{2}  \frac{1}{N} \sum\limits_{i=1}^{N} \frac{1}{|\S_1^i|} \sum\limits_{\z \in |\S_1^i|} \E\|\u^r_{i,k-1}(\z) - g(\wb^r_{k}, \z, \wb^r_k, \S_2)\|^2 \\  
& + \frac{4\gamma^2 |P^r|}{|\S_1^i|N} (\sigma^2+C_0^2) 
+ \frac{8\gamma \beta^2 K^2 C_0^2 |P^r|}{|\S_1^i| N} + 4\tL^2 \frac{|P^r|}{N}\E\|\wb^r - \wb^{r-1}\|^2 + 4\tL^2 \frac{|P^r|}{N} \E\|\wb^r - \wb^r_k\|^2  \\
& 
+ 4(\gamma^2 + \frac{\gamma}{|\S_1^i|} )\tL^2 \frac{1}{N}\sum_{i\in P^r} \E\|\wb^{r} - \w^r_{i,k}\|^2   
+ (\gamma^2 + \frac{\gamma}{|\S_1^i|} ) \tL^2 \frac{1}{NK}\sum\limits_{i\in P^r} \sum\limits_{k=1}^{K} \E\|\wb^{r-1} - \w^{r-1}_{i,k}\|^2. 
\end{split} 
\end{equation} 
Dividing $\frac{\gamma+1}{2}$ on both sides gives
\begin{equation}
\begin{split}
&  \frac{1}{N} \sum\limits_{i=1}^{N} \frac{1}{|\S_1^i|} \sum\limits_{\z \in |\S_1^i|} \E\|\u^r_{i,k}(\z) - g(\wb^r_{k}, \z, \wb^r_k, \S_2)\|^2 \\
&\leq \frac{\gamma(1-\frac{|P^r|}{8|\S_1^i|N}) + 1}{\gamma+1}  \frac{1}{N} \sum\limits_{i=1}^{N} \frac{1}{|\S_1^i|} \sum\limits_{\z \in |\S_1^i|} \E\|\u^r_{i,k-1}(\z) - g(\wb^r_{k}, \z, \wb^r_k, \S_2)\|^2 \\
& + 8\frac{\gamma^2 |P^r|}{|\S_1^i| N} (\sigma^2+C_0^2)  + \frac{16\gamma \beta^2 K^2 C_0^2 |P^r|}{|\S_1^i| N} 
+ 8\tL^2 \frac{|P^r|}{N} \|\wb^r - \wb^{r-1}\|^2 \\
&+  8\tL^2 \frac{|P^r|}{N} \|\wb^r - \wb^{r}_k\|^2 
+ 8(\gamma^2 + \frac{\gamma}{|\S_1^i|} )\tL^2 \frac{1}{N}\sum_{i\in P^r} \|\wb^{r} - \w^r_{i,k}\|^2 \\
&+ 2(\gamma^2 + \frac{\gamma}{|\S_1^i|} ) \tL^2 \frac{1}{NK}\sum\limits_{i\in P^r} \sum\limits_{k=1}^{K} \E\|\wb^{r-1} - \wb^{r-1}_{i,k}\|^2. 
\end{split} 
\end{equation} 
Using Young's inequality, 
\begin{equation*} 
\begin{split}
&\frac{1}{N} \sum\limits_{i=1}^{N} \frac{1}{|\S_1^i|} \sum\limits_{\z \in |\S_1^i|} \E\|\u^r_{i,k}(\z) - g(\wb^r_{k}, \z, \wb^r_k, \S_2)\|^2 \\
&\leq (1-\frac{\gamma |P^r|}{8|\S_1^i|N}) \frac{1}{N} \sum\limits_{i=1}^{N} \frac{1}{|\S_1^i|}  \sum\limits_{\z \in |\S_1^i|} \bigg[(1+\frac{\gamma |P^r|}{16|\S_1^i| N})\E\|\u^r_{i,k-1}(\z) - g(\wb^r_{k-1}, \z, \wb^r_{k-1}, \S_2)\|^2 \\
&~~~  +(1+\frac{16|\S_1^i| N}{\gamma |P^r|})  \tL^2\|\wb^r_{k-1} - \wb^r_k\|^2 \bigg]  \\
& + 8\frac{\gamma^2 |P^r|}{|\S_1^i| N} (\sigma^2+C_0^2)  + \frac{16\gamma \beta^2 K^2 C_0^2 |P^r|}{|\S_1^i| N} \\
& + 8 \tL^2 \frac{|P^r|}{N} \|\wb^r - \wb^{r-1}\|^2 + 8 \tL^2 \frac{|P^r|}{N} \|\wb^r - \wb^{r}_k\|^2 
+ 8(\gamma^2 + \frac{\gamma}{|\S_1^i|} )\tL^2 \frac{1}{N}\sum_{i\in P^r} \|\wb^{r} - \w^r_{i,k}\|^2 \\
%+ \frac{\gamma \tL^2}{|\S_1^i|} \|\wb^r - \wb^r_k\|^2  
&+ 2(\gamma^2 + \frac{\gamma}{|\S_1^i|} ) \tL^2 \frac{1}{NK}\sum\limits_{i\in P^r} \sum\limits_{k=1}^{K} \E\|\wb^{r-1} - \wb^{r-1}_{i,k}\|^2,
\end{split} 
\end{equation*} 
which yields 
\begin{equation*} 
\begin{split}
&\frac{1}{N} \sum\limits_{i=1}^{N} \frac{1}{|\S_1^i|} \sum\limits_{\z \in |\S_1^i|} \E\|\u^r_{i,k}(\z) - g(\wb^r_{k}, \z, \wb^r_k, \S_2)\|^2 \\
&\leq (1-\frac{\gamma |P^r|}{16|\S_1^i|N})\frac{1}{N} \sum\limits_{i=1}^{N} \frac{1}{|\S_1^i|}  \sum\limits_{\z \in |\S_1^i|} [\E\|\u^r_{i,k-1}(\z) - g(\wb^r_{k-1}, \z, \wb^r_{k-1}, \S_2)\|^2 \\
& + \frac{20|\S_1^i|N}{\gamma |P^r|} \tL^2\|\wb^r_{k-1} - \wb^r_k\|^2 ]  + 8\frac{\gamma^2}{|\S_1^i|} \frac{|P^r|}{N} (\sigma^2+C_0^2)   + \frac{16\gamma \beta^2 K^2 C_0^2 |P^r|}{|\S_1^i| N} \\
& + 8 \frac{|P^r|}{N} \tL^2 \|\wb^r - \wb^{r-1}\|^2 + 8 \tL^2 \frac{|P^r|}{N} \|\wb^r - \wb^{r}_k\|^2 \\
&+ 8(\gamma^2 + \frac{\gamma}{|\S_1^i|} )\tL^2 \frac{1}{N}\sum_{i\in P^r} \|\wb^{r} - \w^r_{i,k}\|^2 \\
& + 2(\gamma^2 + \frac{\gamma}{|\S_1^i|} ) \tL^2 \frac{1}{NK}\sum\limits_{i\in P^r} \sum\limits_{k=1}^{K} \E\|\wb^{r-1} - \wb^{r-1}_{i,k}\|^2. 
\end{split} 
\end{equation*} 
\end{proof}

\subsection{Analysis of the estimator of gradient} 
With update $G^r_{i,k} = (1-\beta) G^r_{i,k-1} + \beta (G^r_{i,k,1} + G^r_{i,k,2})$,
we define $\bG^r_k := \frac{1}{|P^r|} \sum\limits_{i\in P^r} G^r_{i,k}$, and $\Delta^r_{k} := \|\bG^r_{k} - \nabla F(\wb^r_k)\|^2$. 
Then it follows that $\bar{G}^r_{k} = (1-\beta) \bG^r_{k-1} + \beta \frac{1}{|P^r|} \sum\limits_{i\in P^r} (G^r_{i,k,1} + G^r_{i,k,2})$. 

\begin{lemma}
\label{lem:nonlinear_lem_G_pc}
Under Assumption \ref{ass:non_linear}, Algorithm \ref{alg:fedx2} ensures that 
\begin{equation*}
\begin{split}
&\Delta^r_k \leq (1-\beta) \|\bG^r_{k-1} - \nabla F(\wb^r_{k-1})\|^2 + \frac{\eta}{16}\|\nabla F(\wb^{r-1})\|^2 \\
& + \frac{2\beta^2 \sigma^2}{|P^r|} 
+ 36\frac{\eta^2}{\beta} \tL^2\|\wb^r_{k} - \wb^{r-1}\|^2 \\
& + 12\beta 
\left(\frac{1}{N}\sum_i 4\tL^2 \E\|\w^r_{i,k} - \wb^r\|^2 + 4\tL^2 \E\|\wb^r - \wb^{r-1}\|^2
+ \frac{1}{N}\sum_i 4\tL^2 \E\|\w^{r-1}_{j',t'} - \wb^{r-1}\|^2  \right) \\
& + 12\beta \frac{1}{N}\sum_i  \bigg( \tL^2 \E\|\u^r_{i,k}(\z^r_{i,k,1}) -  g(\wb^{r}_k,\z^r_{i,k,1},\wb^{r}_k, \S_2)\|^2 
+ \tL^2 \E\|\u^{r-1}_{j',t'}(\hat{\z}^{r-1}_{j',t',1})-g(\wb^{r-1}_{t'}, \hat{\z}^{r-1}_{j',t',1}, \wb^{r-1}_{t'}, \S_2)\|^2\bigg).
\end{split} 
\end{equation*}
\end{lemma} 
\begin{proof}
\begin{small}
\begin{equation}
\begin{split}
& \Delta^r_{k} = \|\bG^r_{k} - \nabla F(\wb^r_k)\|^2 \\ 
& = \|(1-\beta) \bG^r_{k-1} + \beta  \frac{1}{|P^r|} \sum_{i\in P^r} (G^r_{i,k,1} + G^r_{i,k,2}) - \nabla F(\wb^r_{k})  \|^2 \\  
&= \bigg\|(1-\beta) (\bG^r_{k-1} - \nabla F(\wb^r_{k-1})) + (1 - \beta)(\nabla F(\wb^r_{k-1}) - \nabla F(\wb^r_{k})) \\
&~~~ + \beta \bigg(\frac{1}{|P^r|}\sum_{i\in P^r} (G_1(\w^r_{i,k}, \z^r_{i,k,1}, \u^r_{i,k}(\z^r_{i,k,1}), \w^{r-1}_{j,t}, \hat{\z}^{r-1}_{j,t,2})
+ G_2(\w^{r-1}_{j', t'}, \hat{\z}^{r-1}_{j',t',1}, \u^{r-1}_{j',t'}(\hat{\z}^{r-1}_{j',t',1}), \w^r_{i,k}, \z^r_{i,k,2})) \\
%&~~~~~~~~~~~~~ -\frac{1}{N}\sum_{i} (G_1(\wb^{r-1}, \z^r_{i,k,1}, \u^{r}_{i,k}(\z^r_{i,k,1}), \w^{r-1}_{j,t},  \hat{\z}^{r-1}_{j,t,2}) 
%+ G_2(\wb^{r-1}, \hat{\z}^{r-1}_{j',t',1}, \u^{r-1}_{j',t'}(\hat{\z}^{r-1}_{j',t',1}), \w^r_{i,k}, \z^r_{i,k,2})) \bigg) \\ 
%& + \beta \bigg(\frac{1}{N}\sum_{i} (G_1(\wb^{r-1}, \z^r_{i,k,1}, \u^r_{i,k}(\z^r_{i,k,1}), \wb^{r-1}, \hat{\z}^{r-1}_{j,t,2}) 
%+ G_2(\wb^{r-1}, \hat{\z}^{r-1}_{j',t',1}, \u^{r-1}_{j',t'}(\hat{\z}^{r-1}_{j',t',1}), \wb^{r-1}, \z^r_{i,k,2}))  \bigg) \\
%& ~~~~~~~~~~~~~ - \frac{1}{N}\sum_{i} (G_1(\wb^{r-1}, \z^r_{i,k,1}, g(\wb^r, \z^r_{i,k,1}, \wb^r, \S_2), \wb^{r-1}, \hat{\z}^{r-1}_{j,t,2}) \\
%& ~~~~~~~~~~~~~ ~~~~~~~~~~~~~ ~~~~~
%+ G_2(\wb^{r-1}, \hat{\z}^{r-1}_{j',t',1}, g(\wb^{r-1}, \hat{\z}^{r-1}_{j',t',1}, \wb^{r-1}, \S_2), \wb^{r-1}, \z^r_{i,k,2}))  \bigg) \\
%& + \beta \bigg(\frac{1}{N}\sum_{i} (G_1(\wb^{r-1}, \z^r_{i,k,1}, g(\wb^r, \z^r_{i,k,1}, \wb^r, \S_2), \wb^{r-1}, \hat{\z}^{r-1}_{j,t,2}) \\
%& ~~~~~~~~~~~~~ ~~~~~~~~~~~~~ ~~~~~ 
%+ G_2(\wb^{r-1}, \hat{\z}^{r-1}_{j',t',1}, g(\wb^{r-1}, \hat{\z}^{r-1}_{j',t',1}, \wb^{r-1}, \S_2), \wb^{r-1}, \z^r_{i,k,2}))   \\  
& ~~~~~~~~~~~~~ - \frac{1}{P^r}\sum_{i\in P^r} (G_1(\wb^{r-1}, \z^r_{i,k,1}, g(\wb^{r-1}, \z^r_{i,k,1}, \wb^{r-1}, \S_2), \wb^{r-1}, \hat{\z}^{r-1}_{j,t,2}) \\
& ~~~~~~~~~~~~~ ~~~~~~~~~~~~~ ~~~~~ 
+ G_2(\wb^{r-1}, \hat{\z}^{r-1}_{j',t',1}, g(\wb^{r-1}, \hat{\z}^{r-1}_{j',t',1}, \wb^{r-1}, \S_2), \wb^{r-1}, \z^r_{i,k,2})) \bigg) \\ 
& + \beta \bigg(\frac{1}{|P^r|}\sum_{i\in P^r} (G_1(\wb^{r-1}, \z^r_{i,k,1}, g(\wb^{r-1}, \z^r_{i,k,1}, \wb^{r-1}, \S_2), \wb^{r-1}, \hat{\z}^{r-1}_{j,t,2}) \\
& ~~~~~~~~~~~~~ ~~~~~~~~~~~~~ ~~~~~  
+ G_2(\wb^{r-1}, \hat{\z}^{r-1}_{j',t',1}, g(\wb^{r-1}, \hat{\z}^{r-1}_{j',t',1}, \wb^{r-1}, \S_2), \wb^{r-1}, \z^r_{i,k,2})) 
- \nabla F(\wb^r_k) \bigg) 
\bigg\|^2.
\end{split}    
\end{equation}  
\end{small} 
% Denoting $g_{\z}(\w) = g(\w, \z, \w, \S_2)$. 
Using Young's inequality and $\tL$-Lipschtzness of $G_1, G_2$, we can then derive
%\begin{small} 
\begin{equation}
\begin{split}
&\Delta^r_k \leq (1+\beta) \Bigg\| (1-\beta)(\bG^r_{k-1} - \nabla F(\wb^r_{k-1}) ) \\
&+ \beta \bigg(\frac{1}{|P^r|}\sum_{i\in P^r} (G_1(\wb^{r-1}, \z^r_{i,k,1}, g(\wb^{r-1},\z^r_{i,k,1},\wb^{r-1}, \S_2), \wb^{r-1}, \hat{\z}^{r-1}_{j,t,2}) \\
&~~~~~~~~~~~~~~~~~~~
+ G_2(\wb^{r-1}, \hat{\z}^{r-1}_{j',t',1}, g(\wb^{r-1}, {\hat{\z}^{r-1}_{j',t',1}}, \wb^{r-1}, \S_2), \wb^{r-1}, \z^r_{i,k,2})) 
- \nabla F(\wb^{r-1})\bigg)\Bigg\|^2 \\
& + (1+\frac{10}{\beta})\beta^2 
\left(\frac{1}{|P^r|}\sum_{i\in P^r} 4\tL^2 \E\|\w^r_{i,k} - \wb^r\|^2 + 4\tL^2 \E\|\wb^r - \wb^{r-1}\|^2
+ \frac{1}{N}\sum_i 4\tL^2 \E\|\w^{r-1}_{j',t'} - \wb^{r-1}\|^2  \right) \\
& + (1+\frac{10}{\beta})\|\wb^r_{k-1} - \wb^r_k\|^2 + (1+\frac{10}{\beta})\beta^2 \frac{1}{|P^r|}\sum_{i\in P^r}  \bigg( \tL^2 \E\|\u^r_{i,k}(\z^r_{i,k,1}) -  g(\wb^{r}_k,\z^r_{i,k,1},\wb^{r}_k, \S_2)\|^2 \\
&~~~~~~~~~~~~~~~~~~~~~~~~~~~~
+ \tL^2 \E\|\u^{r-1}_{j',t'}(\hat{\z}^{r-1}_{j',t',1})-g(\wb^{r-1}_{t'}, \hat{\z}^{r-1}_{j',t',1}, \wb^{r-1}_{t'}, \S_2)\|^2
\bigg). 
\end{split}
\end{equation}
%\end{small}
By the fact that 
\begin{equation}
\begin{split}
&\E[\frac{1}{|P^r|}\sum_{i\in P^r} (G_1(\wb^{r-1}, \z^r_{i,k,1}, g(\wb^{r-1},\z^r_{i,k,1},\wb^{r-1},\S_2), \wb^{r-1}, \hat{\z}^{r-1}_{j,t,2}) \\
&~~~~~~~~~~~
+ G_2(\wb^{r-1}, \hat{\z}^{r-1}_{j',t',1}, g(\wb^{r-1},{\hat{\z}^{r-1}_{j',t',1}},\wb^{r-1},\S_2), \wb^{r-1}, \z^r_{i,k,2})) 
- \nabla F(\wb^{r-1})]  = 0,
\end{split}
\end{equation} 
\begin{equation}
\begin{split}
&\E\|\frac{1}{|P^r|} \sum_{i\in P^r} (G_1(\wb^{r-1}, \z^r_{i,k,1}, g(\wb^{r-1},\z^r_{i,k,1},\wb^{r-1},\S_2), \wb^{r-1}, \hat{\z}^{r-1}_{j,t,2})  \\
&~~~~~~~
+ G_2(\wb^{r-1}, \hat{\z}^{r-1}_{j',t',1}, g(\wb^{r-1},{\hat{\z}^{r-1}_{j',t',1}},\wb^{r-1},\S_2), \wb^{r-1}, \z^r_{i,k,2})) 
- \nabla F(\wb^{r-1})  \|^2 
\leq \frac{\sigma^2}{|P^r|}  
\end{split} 
\end{equation} 
and
\begin{equation}
\begin{split}
\|\wb^r_{k-1}-\wb^r_k\|^2 = \eta^2\|\bar{G}^r_k\|^2
\leq 3\eta^2 \|\bar{G}^r_k - \nabla F(\wb^r_k) \|^2
+ 3\eta^2 \|\nabla F(\wb^r_k) - \nabla F(\wb^{r-1}) \|^2 
+ 3\eta^2 \|\nabla F(\wb^{r-1})\|^2, 
\end{split}
\end{equation}
we obtain 
\begin{equation*}
\begin{split}
&\Delta^r_k \leq (1-\frac{3\beta}{4}) \|\bG^r_{k-1} - \nabla F(\wb^r_{k-1})\|^2
+ \frac{2\beta^2 \sigma^2}{|P^r|} 
+ 36\frac{\eta^2}{\beta} \tL^2\|\wb^r_{k} - \wb^{r-1}\|^2  
+ \frac{\eta}{16} \|\nabla F(\wb^{r-1})\|^2 \\
& + 12\beta 
\left(\frac{1}{|P^r|}\sum_{i\in P^r} 4\tL^2 \E\|\w^r_{i,k} - \wb^r\|^2 + 4\tL^2 \E\|\wb^r - \wb^{r-1}\|^2
+ \frac{1}{|P^r|}\sum_{i\in P^r} 4\tL^2 \E\|\w^{r-1}_{j',t'} - \wb^{r-1}\|^2  \right) \\
& + 12\beta \frac{1}{|P^r|}\sum_{i\in P^r}  \bigg( \tL^2 \E\|\u^r_{i,k}(\z^r_{i,k,1}) -  g(\wb^{r}_k,\z^r_{i,k,1},\wb^{r}_k, \S_2)\|^2
% \\
%&~~~~~~~~~~~~~~~~~~~ 
+ \tL^2 \E\|\u^{r-1}_{j',t'}(\hat{\z}^{r-1}_{j',t',1})-g(\wb^{r-1}_{t'}, \hat{\z}^{r-1}_{j',t',1}, \wb^{r-1}_{t'}, \S_2)\|^2\bigg).
\end{split} 
\end{equation*}
\end{proof} 

%Then,
%\begin{small}
%\begin{equation}
%\begin{split}
%&\frac{1}{K}\sum_k \Delta^r_k \\
%&\leq (1-\frac{\beta}{2}) \frac{1}{K} \sum\limits_{k}\|\bG^r_{k-1} - \nabla F(\wb^r_{k-1})\|^2
%+ 4\frac{\beta^2 \sigma^2}{N} + 5\beta \frac{1}{NK}\sum_i\sum_k\|\wb^{r-1} - \w^r_{i,k}\|^2 + 10\beta \|\u^r_{i,k}(\z^r_{i,k,1}) - g_{\z^r_{i,k,1}}(\wb^r))\|^2. 
%\end{split}
%\end{equation}
%\end{small} 

\subsection{Convergence Result}

\begin{theorem}
\label{thm:nonlinear_formal_pc}
Suppose Assumption \ref{ass:non_linear} holds, and assume there are at least $|P|$ machines take participation in each round.
Denoting $M = \max_i |\S^1_i|$ as the largest number of data on a single machine, by setting $\gamma=O(\frac{M^{1/3}}{R^{2/3}})$, $\beta=O(\frac{1}{M^{1/6} R^{2/3}})$, $\eta = O(\frac{|P|}{N M^{2/3} R^{2/3}})$ and $K=O(\frac{N M^{1/3} R^{1/3}}{|P|})$,  Algorithm \ref{alg:1} ensures that 
%\begin{equation} 
$\E\left[\frac{1}{R}\sum_{r=1}^R \|\nabla F(\wb^r)\|^2\right] \leq O(\frac{1}{R^{2/3}})$.  
%\end{equation}
\end{theorem}

%We re-present Theorem \ref{thm:nonlinear_informal} as below.
%\begin{theorem}
%\label{thm:nonlinear_formal}
%Suppose Assumption \ref{ass:non_linear} holds,
%denoting $M = \max_i |\S^1_i|$ as the largest number of data on a single machine, by setting $\gamma=O(\frac{M^{1/3}}{R^{2/3}})$, $\beta=O(\frac{1}{M^{1/6} R^{2/3}})$, $\eta = O(\frac{1}{M^{2/3} R^{2/3}})$ and $K=O(M^{1/3} R^{1/3})$,  Algorithm \ref{alg:1} ensures that 
%\begin{equation} 
%$\E\left[\frac{1}{R}\sum_{r=1}^R \|\nabla F(\wb^r)\|^2\right] \leq O(\frac{1}{R^{2/3}})$.
%\end{equation}
%\end{theorem}

\begin{proof}
By updating rules, we have that for $i\in P^r$,
\begin{equation}
\begin{split}
&\|\wb^r - \w^r_{i,k}\|^2 
\leq  \eta^2 K^2 C_f^2 C_\ell^2 C_g^2,
\end{split}
\label{app:thm3_eq_1_pc}
\end{equation}
and 
\begin{equation}
\begin{split}
&\|\wb^{r}_k - \wb^r\|^2 = \teta^2 \|\frac{1}{|P^r| K}\sum_{i\in P^r} \sum\limits_{m=1}^{k} \bG^r_{m}\|^2 \leq \teta^2 \frac{1}{K} \sum_{m=1}^K \|\bG^r_m - \nabla F(\wb^r_m) + \nabla F(\wb^r_m)\|^2.  
\end{split}
\label{app:thm3_eq_2_pc}
\end{equation} 

Similarly, we also have
\begin{equation}
\begin{split}
&\|\wb^{r-1} - \wb^r\|^2 = \teta^2 \|\frac{1}{|P^r| K}\sum_{i\in P^r} \sum_{k=1}^K \bG^{r-1}_{k}\|^2\leq \teta^2 \frac{1}{K} \sum_{k=1}^K \|\bG^{r-1}_k - \nabla F(\wb^{r-1}_{k}) + \nabla F(\wb^{r-1}_{k})\|^2.
\end{split}
\label{app:thm3_eq_3_pc}
\end{equation}

Lemma \ref{lem:nonlinear_lem_G_pc} yields that
\begin{small}
\begin{equation}
\begin{split} 
& \frac{1}{RK}\sum_{r,k} \E\|\bG^r_{k} - \nabla F(\wb^r_{k})\|^2 \leq \frac{\Delta^0_0}{\beta RK} 
+ \frac{\beta\sigma^2}{|P^r|}  \\
& + 2 \left(\frac{1}{|P^r|}\sum_{i\in P^i} 4\tL^2 \E\|\w^r_{i,k} - \wb^r\|^2 + 4\tL^2 \E\|\wb^r - \wb^{r-1}\|^2
+ \frac{1}{|P^r|}\sum_{i\in P^r} 4\tL^2 \E\|\w^{r-1}_{j',t'} - \wb^{r-1}\|^2  \right)
\\
&+ 2\E\left[\frac{1}{R}\sum_r \frac{1}{|P^r|K}\sum_{i\in P^r,k} \frac{1}{|\S_1^i|} \sum_{\z\in\S_1^i} \|\u^r_{i,k}(\z) - g(\wb^r, \z, \wb^r, \S_2)\|^2 \right] \\
&~~ + 2\E\left[\frac{1}{R}\sum_r  \frac{1}{|P^r|K} \sum_{j',t'}  \frac{1}{|\S_1^i|} \sum_{\z\in\S_1^i} \|\u^{r-1}_{j',t'}(\z) -  g(\wb^{r-1}_{t'}, \z, \wb^{r-1}_{t'}, \S_2))\|^2 \right], 
\end{split} 
\end{equation} 
\end{small}
which by setting of $\eta$ and $\beta$ leads to
\begin{small}
\begin{equation*}
\begin{split}
&\frac{1}{RK}\sum_{r,k} \E\|\bG^r_{k} - \nabla F(\wb^r_{k})\|^2 \leq \frac{2\Delta^0_0}{\beta RK}  
+ \frac{4\beta\sigma^2}{|P|} 
+ 10\beta \teta^2 C_\ell^2 C_g^2 + \frac{1}{4R}\sum_r \|\nabla F(\wb^{r-1})\|^2 \\
&+ 16\frac{1}{R}\sum_r\frac{1}{NK}\sum_{i,k} \frac{1}{|\S_1^i|} \sum_{\z\in\S_1^i} \E\|\u^r_{i,k}(\z) - g(\wb^r; \z, \S_2)\|^2 \\
&+ 32\frac{1}{R}\sum_r \frac{1}{NK} \sum_{j',t'}  \frac{1}{|\S_1^i|} \sum_{\z\in\S_1^i} \E\|\u^{r-1}_{j',t'}(\hat{\z}^{r-1}_{j',t',1}) -  g(\wb^{r-1};\hat{\z}^{r-1}_{j',t',1}, \S_2))\|^2 \\
&+ 32C_g^2 \frac{1}{R} \sum_r\frac{1}{K}\sum_{t'} \E\|\wb^{r-1} - \wb^{r-1}_{t'}\|^2. 
\end{split} 
\end{equation*} 
\end{small} 
Using Lemma \ref{lem:nonlinear_u_pc} yields
\begin{equation*}
\begin{split}
&\frac{1}{R}\sum_r\frac{1}{NK}\sum\limits_{i=1}^{N}\sum\limits_{k=1}^K
\frac{1}{|\S_1^i|}\sum\limits_{\z\in \S_1^i} \E \|\u^r_{i,k}(\z) - g(\wb^r_{k}, \z, \wb^r_{k}, \S_2)\|^2 \\
&\leq \frac{16M N}{\gamma |P^r|}  \frac{1}{R} \frac{1}{NK}\sum\limits_{i=1}^{N} 
\frac{1}{|\S_1^i|}\sum\limits_{\z\in \S_1^i}  \E \|\u^0_{i,0}(\z) - g(\wb^0_{0}, \z, \wb^0_{0}, \S_2)\|^2 \\
& + \frac{400M^2 N^2}{\gamma^2 |P^r|^2} \frac{1}{RK}\sum_{r,k} \tL^2\|\wb^r_{k-1} - \wb^r_k\|^2  + 150\gamma (\sigma^2+C_0^2)  + 256 \beta^2 K^2 C_0^2  \\ 
& + 128 \tL^2 \frac{|\S_1^i|}{\gamma} (\|\wb^r-\wb^{r-1}\|^2 + \|\wb^r-\wb^{r-1} \|^2) \\
& + 150 (\gamma |\S_1^i| + 1)\tL^2 \frac{1}{N}\sum_{i} \|\wb^{r} - \w^r_{i,k}\|^2 
%+ \frac{\gamma \tL^2}{|\S_1^i|} \|\wb^r - \wb^r_k\|^2  
+ 32 (\gamma |\S_1^i| + 1) \tL^2 \frac{1}{NK}\sum\limits_{i=1}^N \sum\limits_{k=1}^{K} \E\|\wb^{r-1} - \wb^{r-1}_{i,k}\|^2. 
%&~~~ + 8\gamma M \frac{1}{R}\sum_r \|\wb^r - \wb^{r-1}\|^2 
%+ 8\gamma |\S_1^i| \frac{1}{RNK}\sum_{r,i,k} \|\wb^{r} - \w^r_{i,k}\|^2 
%+ 8\frac{|\S_1^i|}{\gamma}\frac{1}{RK}\sum_{r,k}.  \|\wb^r - \wb^r_k\|^2. 
\end{split}
\end{equation*}
Combining this with previous five inequalities and noting the parameters settings, we obtain
\begin{equation*} 
\begin{split} 
&\frac{1}{R}\sum_r\frac{1}{NK}\sum\limits_{i=1}^{N}\sum\limits_{k=1}^K
\frac{1}{|\S_1^i|}\sum\limits_{\z\in \S_1^i} \E \|\u^r_{i,k}(\z) - g(\wb^r_{k}, \z, \wb^r_{k}, \S_2)\|^2 \\
&\leq O\bigg(\frac{M N}{\gamma R K |P|} + \eta^2 \frac{M^2 N^2}{\gamma^2 |P|^2}\frac{1}{RK}\sum_{r,k} \E\|\bG^r_{k} - \nabla F(\wb^r_{k})\|^2 +\gamma + \beta^2K^2  + \frac{M}{\gamma} \teta^2 (\frac{1}{\beta RK}+ \frac{\beta}{|P|}) \\
&~~~~~~~~~~~~ + \gamma M \eta^2 K^2  + \frac{1}{R} \sum_r \teta^2 \|\nabla F(\wb^{r-1})\|^2 \bigg) 
\end{split}
\label{app:lem1_corollary_pc}
\end{equation*}
and 
\begin{equation}
\begin{split}
&\frac{1}{RK}\sum_{r,k} \E\|\bG^r_{k} - \nabla F(\wb^r_{k})\|^2 \\
&\leq O\left(\frac{M N}{\gamma R K |P|} +\gamma + \beta^2K^2  + \frac{M}{\gamma} \teta^2 (\frac{1}{\beta RK}+ \frac{\beta}{|P|}) + \gamma M \eta^2 K^2  + \frac{1}{R} \sum_r \teta^2 \|\nabla F(\wb^{r-1})\|^2 \right). 
\end{split} 
\label{app:lem2_corollary_pc}
\end{equation} 
Then using the standard analysis of smooth function, we derive
\begin{equation} 
\begin{split}
&F(\wb^{r+1}) - F(\wb^r) \leq \nabla F(\wb^r)^\top (\wb^{r+1} - \wb^r) + \frac{\tL}{2} \|\wb^{r+1} - \wb^r\|^2 \\
% & = - \teta \nabla F(\wb^r)^\top 
% \left( \frac{1}{NK} \sum_i \sum_k G^r_{i,k}  \right) + \frac{\tL}{2} \|\wb^{r+1} - \wb^r\|^2 \\
& = - \teta \nabla F(\wb^r)^\top 
\left( \frac{1}{NK} \sum_i \sum_k G^r_{i,k} 
 - \nabla F(\wb^r) + \nabla F(\wb^r)
\right) + \frac{\tL}{2} \|\wb^{r+1} - \wb^r\|^2 \\
& = -\teta \|\nabla F(\wb^r)\|^2 + \frac{\teta}{2} \|\nabla F(\wb^r)\|^2 + \frac{\teta}{2} \|  \frac{1}{NK} \sum_i \sum_k G^r_{i,k} 
 - \nabla F(\wb^r) \|^2 \\
&~~~ + \frac{\tL}{2} \|\wb^{r+1} - \wb^r\|^2 \\
& \leq -\frac{\teta}{2} \|\nabla F(\wb^r)\|^2
+\teta \|\frac{1}{NK}\sum_i\sum_k (G^r_{i,k} - \nabla F(\wb^r_k))\|^2 \\
&~~~+ \teta \|\frac{1}{K} \sum_k (\nabla F(\wb^r_k) - \nabla F(\wb^r))\|^2  + \frac{\tL}{2} \|\wb^{r+1} - \wb^r\|^2 \\
&\leq -\frac{\teta}{2} \|\nabla F(\wb^r)\|^2
+\teta \frac{1}{K} \sum_k \|\frac{1}{N}\sum_i (G^r_{i,k} - \nabla F(\wb^r_k))\|^2 \\
&~~~ +\teta \frac{\tL^2}{K} \sum_k \|\wb^r_{k} - \wb^r\|^2 + \frac{\tL}{2} \|\wb^{r+1} - \wb^r\|^2.
\end{split} 
\end{equation}

Combining with (\ref{app:lem2_corollary_pc}), (\ref{app:thm3_eq_1_pc}), (\ref{app:thm3_eq_2_pc}), and (\ref{app:thm3_eq_3_pc}), we derive 
\begin{equation*}
\begin{split}
& \frac{1}{R} \sum_r \E\|\nabla F(\wb^{r})\|^2 \leq O\left(\frac{M N}{\gamma R K |P|} +\gamma + \beta^2K^2  + \frac{M}{\gamma} \teta^2 (\frac{1}{\beta RK}+ \frac{\beta}{|P|}) + \gamma M \eta^2 K^2 \right). 
% &\leq \frac{F(\wb^0) - F(\wb^{R+1})}{\teta R} + \frac{\beta\sigma^2}{N} + \eta^2K^2 P_i + \gamma \sigma^2  
\end{split}
\end{equation*}
By setting parameters as in the theorem, we can conclude the proof. 
Further, to get  $\frac{1}{R} \sum_r \E\|\nabla F(\wb^{r})\|^2\leq \epsilon^2$,
we just need to set $\gamma = O(\epsilon^2)$, $\beta = O(\frac{\epsilon^2}{\sqrt{M}})$, $K=O(\frac{N\sqrt{M}}{|P|\epsilon})$, $\eta=O(\frac{|P|\epsilon^2}{N M})$,
$R=O(\frac{\sqrt{M}}{\epsilon^3})$. 
\end{proof}

\end{document}